\documentclass{article}

\usepackage[preprint]{neurips_2026}

\usepackage[utf8]{inputenc} 
\usepackage[T1]{fontenc}    
\usepackage{hyperref}       
\usepackage{url}            
\usepackage{booktabs}       
\usepackage{amsfonts}       
\usepackage{nicefrac}       
\usepackage{microtype}      
\usepackage{xcolor}         

\usepackage{comment}
\usepackage{graphicx}
\usepackage[table]{xcolor}  
\usepackage{enumitem}
\usepackage{pgffor}         
\usepackage{ifthen}         
\usepackage{float}

\usepackage{xcolor}
\usepackage{fvextra}
\usepackage[most]{tcolorbox}
\tcbuselibrary{breakable}

\input{code_listings/default.pygstyle}

\definecolor{codebg}{RGB}{248,248,248}
\definecolor{codeframe}{RGB}{210,210,210}

\tcbset{
  jaxcode/.style={
    colback=codebg,
    colframe=codeframe,
    boxrule=1pt,
    arc=2pt,
    left=6pt,
    right=6pt,
    top=5pt,
    bottom=5pt,
    enhanced,
  }
}

\newtcolorbox{codebox}[1][]{%
  jaxcode,
  before upper={\fvset{fontsize=\footnotesize,breaklines,numbers=left,numbersep=2pt,tabsize=4}},
  #1}

\usepackage{wrapfig}
\usepackage{colortbl}   

\usepackage{amsmath}
\usepackage{amssymb}
\usepackage{mathtools}
\usepackage{amsthm}

\usepackage[capitalize,noabbrev]{cleveref}

\usepackage{xspace}
\usepackage{subcaption}

\usepackage{authblk}

\theoremstyle{plain}

\theoremstyle{definition}

\theoremstyle{remark}

\usepackage[textsize=tiny]{todonotes}

\definecolor{maroon}{cmyk}{0.2, 1, 0, 0.1}

\newcommand{\eg}{\emph{e.g.}~}

\newcommand{\cf}{\emph{cf.}~}

\newcommand{\jaxatari}{JAXtari\xspace}

\title{JAXtari: High-Throughput and Easy-to-Modify Arcade Learning Environment}

\author[1,2,*,\dag]{Quentin Delfosse}
\author[1,3,*]{Raban Emunds}
\author[1,*]{Paul Seitz}
\author[1,*]{Sebastian Wette}
\author[1,3,*]{Jannis Bl\"uml}
\author[1]{Daniel Kirn}
\author[1]{Dominik Mandok}
\author[1,3,4,5]{Kristian Kersting}

\affil[1]{Department of Computer Science, Technical University Darmstadt, Germany}
\affil[2]{Google Intrinsic AI Research}
\affil[3]{Hessian Center for Artificial Intelligence (hessian.AI)}
\affil[4]{Centre for Cognitive Science, Darmstadt}
\affil[5]{German Research Center for Artificial Intelligence (DFKI)}

\begin{document}
\maketitle

\begin{abstract}
The Arcade Learning Environment (ALE) has been a cornerstone of Deep Reinforcement Learning research, providing a standardized benchmark for evaluating general agents.
However, its lack of native support for object-extraction and environment modification restricts research into evaluating whether agents learn robust causal models or rely on brittle visual heuristics.
Furthermore, the ALE's dependence on CPU-based emulation creates a significant bottleneck in modern GPU-accelerated training pipelines, limiting both experimental scale and iteration frequency. 
In this work, we introduce \jaxatari, an open-source, high-performance reimplementation of Atari 2600 environments.
By leveraging massive GPU parallelization, \jaxatari allows PPO agents to finish training on 100M environment steps on a single A100-GPU in under an hour. 
Beyond speed, it provides object-centric observations and a flexible modification system, allowing researchers to, \eg develop neuro-symbolic methods or discover novel shortcuts and subsequently train and adapt their agents rapidly, sometimes in only minutes.
\end{abstract}

{
\renewcommand{\thefootnote}{*}
\footnotetext{These authors contributed equally to this work.}

\renewcommand{\thefootnote}{\dag}
\footnotetext{Work done while at TU Darmstadt; now at Google Intrinsic AI Research}
}

\section{Introduction}
\begin{figure}[tb]
    \centering
    \includegraphics[width=1.\linewidth]{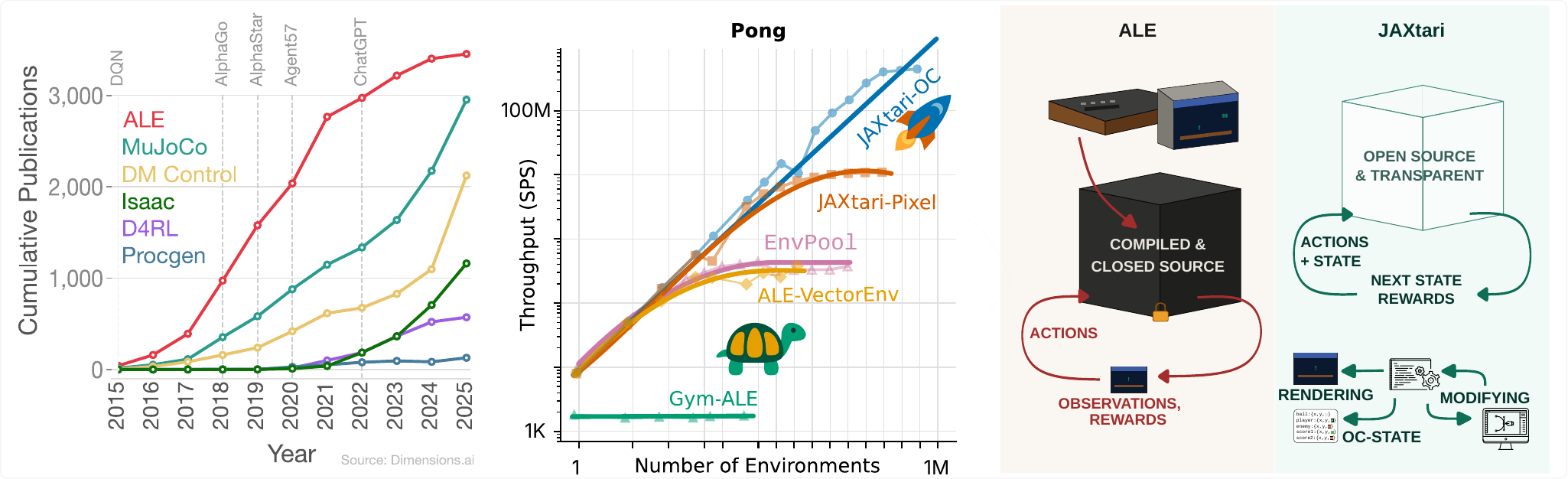}
    \caption{\textbf{\jaxatari modernizes ALE as a benchmark for scalable and robust evaluation.} ALE is the most-used RL benchmark (left), but its setup is constrained by CPU-bound emulation (center), pixel-only observations, and ROM-level barriers to semantic modification (right). \jaxatari preserves ALE's visual fidelity and dynamics, while reimplementing game logic in JAX, enabling accelerator-native throughput, direct source-level modifications, and native access to object-centric and symbolic state.}
    \label{fig:problem}
    \vspace{-3mm}
\end{figure}

The Arcade Learning Environment (ALE)~\citep{bellemare2013arcade} has shaped deep reinforcement learning (RL) for over a decade~\citep{mnih2015human, machado2018revisiting, Badia2020agent57}. 
As a suite of games originally designed by professional developers for human players, Atari possesses a unique advantage: it is inherently free of the "experimenter bias" that often plagues custom-built RL environments. 
Consequently, ALE remains the primary RL benchmark (\cf Figure~\ref{fig:problem}), particularly for studying tasks that require abstract, relational concepts and long-horizon planning, standing in contrast to continuous control benchmarks (\eg MuJoCo~\citep{todorov2012mujoco}, DM Control~\citep{tunyasuvunakool2020dm_control}, D4RL~\citep{fu2020d4rl}), that require low-level motor actuation control.

However, despite superhuman aggregate scores achieved on the diverse set of reasoning tasks, ALE is far from being solved. 
Recent analyses reveal that high performance often stems from \textit{shortcut learning} rather than robust human-like cognition; agents overfit to highly specific visual cues and frequently fail completely when a task is trivially simplified or visually perturbed~\citep{farebrother2018generalization, taiga2023investigating, delfosse2025deep}. 
Diagnosing and mitigating this brittleness requires controlled, counterfactual testing---such as systematically altering object locations, simplifying mechanics, or shifting domains, as introduced by HackAtari~\citep{delfosse2024hackatari, delfosse2025deep}. 
Unfortunately ALE prevents it, as it emulates closed-source binary ROMs that are semantically opaque. 
Both the games logic and their states are not natively interpretable, preventing straightforward implementations of neurosymbolic RL methods and of controlled semantic modifications. 
Furthermore, the CPU-bound nature of the emulation is completely out of step with modern accelerator-centric training pipelines. 
The continuous host--device synchronization required to transfer data between the CPU and GPU heavily penalizes wall-clock time, acting as a bottleneck for highly parallel algorithms, population-based training, and broad hyperparameter sweeps.

To resolve this dichotomy between the necessity of Atari as a reasoning benchmark and the limitations of its legacy architecture, we introduce and open source \jaxatari\footnote{Code provided at \url{https://github.com/k4ntz/JAXtari}}, a framework of faithful, fully JAX-based reimplementations of the ALE environments. 
By reconstructing the game logic from the ground up rather than wrapping legacy ROMs, we preserve the human-designed, bias-free nature of Atari while making the environments accelerator-native, semantically editable, and inherently object-centric. 
\jaxatari can expose the underlying game logic, rich object-centric (OC) states, and rendering pipeline directly to the user, bridging the gap between high-throughput scalability and the deep, counterfactual evaluation required to diagnose and mitigate goal misgeneralization.

Our main contributions are:
\begin{enumerate}[leftmargin=20pt,itemsep=-1pt,parsep=2pt,topsep=0pt,partopsep=2pt]
    \item[\textbf{(i)}] \textbf{Flexible Modification API:}
    We introduce a modular plugin system for editing game semantics directly at the source-code level, turning robustness evaluations and zero-shot shortcut-learning experiments into simple, few-line changes rather than complex ROM reverse-engineering projects.

    \item[\textbf{(ii)}] \textbf{Object-Centric by Design:}
    Unlike the constraints of ALE, \jaxatari exposes structured object states natively from every environment. This removes the need for post-hoc feature extractors and enables clean, rigorous studies of relational, compositional, and logic-augmented RL.

    \item[\textbf{(iii)}] \textbf{ALE Fidelity and Learning Task Parity:}
    Rather than constraining simulation to bit-level 1:1 emulation, \jaxatari targets behavioral and learning-task parity, preserving the human-designed mechanics and challenge of ALE. We validate this consistency by comparing learning dynamics and performance rankings across platforms using standard CleanRL and Gymnasium-compatible baselines.

    \item[\textbf{(iv)}] \textbf{Accelerator-Native Throughput:}
    We provide pure-JAX, JIT-compiled, \texttt{vmap}-able environments for high-throughput execution. By eliminating the CPU-GPU communication bottleneck, \jaxatari allows population-scale evaluation.
\end{enumerate}

The remainder of the paper introduces the design of \jaxatari, describes its object-centric and modification interfaces, evaluates fidelity against the ALE, and benchmarks accelerator-native throughput across games and training settings.

\section{JAXtari: The JAX-native Arcade Framework}
\label{sec:method}
 
This section introduces the \jaxatari framework. 
We first provide the technical details of the \jaxatari implementation, then describe the unique features it offers compared to the ALE.


\subsection{Architecture}
\begin{figure}[t]
    \centering    \includegraphics[width=\linewidth]{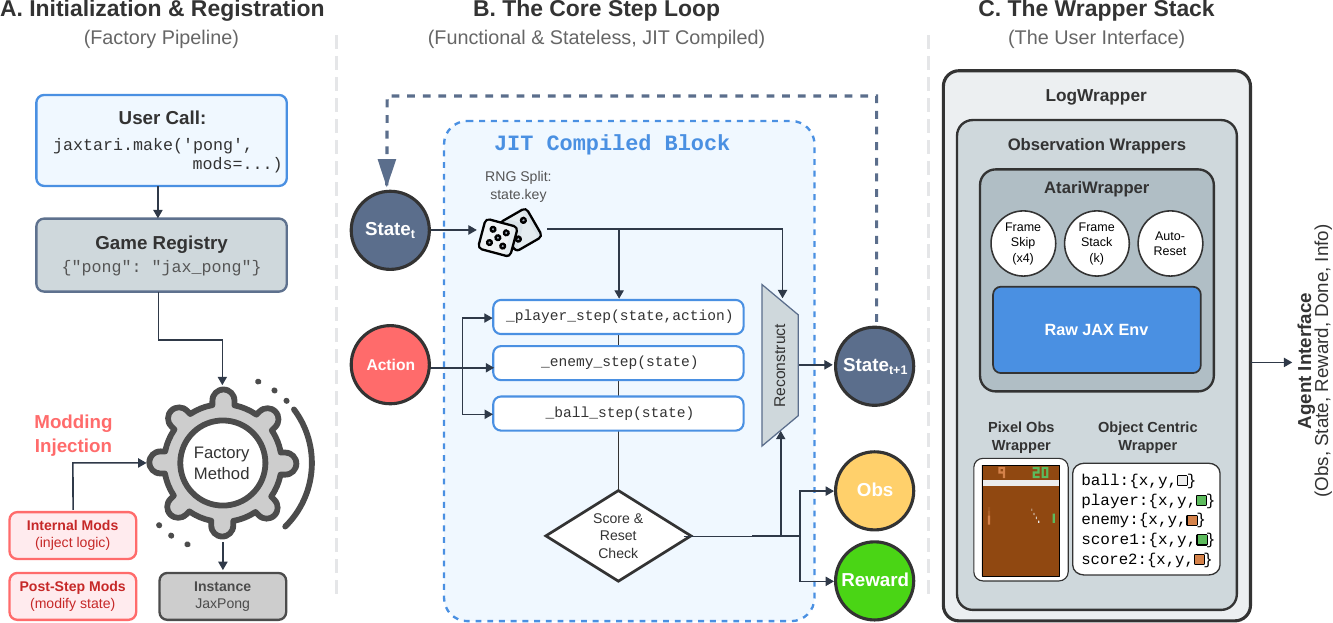}
    \caption{\textbf{JAXtari Architecture.} (A) Environment creation adheres to the gym API, but additionally enables injecting modifications. (B) The functional and fully JIT-compiled transition function produces the next state, observation and reward given the previous state and action. (C) The comprehensive wrapper suite implements best practices and simplifies training.}
    \label{fig:architecture}
    \vspace{-3mm}
\end{figure}

Implementing \jaxatari requires a fundamental shift from the stateful design of the ALE to a purely functional, stateless paradigm compatible with JAX's Just-In-Time (JIT) compilation.
Unlike the ALE, that relies on the Atari binary ROMs (\eg via hardware emulation or automated ROM-to-source transpilation), \jaxatari offers JAX-native, bottom-up reimplementations of Atari environments.\linebreak
Concretely, game physics, collision handling, object updates, scoring rules, and termination conditions are manually reimplemented using high-level JAX array operations.
The environment transitions are placed inside a single XLA (Accelerated Linear Algebra) compiled computation graph, removing the need for a CPU-bound emulator and enabling direct execution on accelerators.
We provide an overview of the \jaxatari architecture in \Cref{fig:architecture}. Let us now detail the different \jaxatari components.

\textbf{States.}
In \jaxatari, the transition function follows the signature \texttt{step(state, action) $\rightarrow$ (observation, next\_state, reward, done, info)}, ensuring full compatibility with the gymnax API~\citep{gymnax2022github}. 
The returned values are either primitive types (reward, done) or defined as immutable \texttt{PyTreeNode} objects (\eg \texttt{PongState}) that contain all necessary variables (\eg player positions, ball velocity, scores) and enable direct use within JAX transformations.
This allows future algorithms to perform interventions outside the game engine, \eg a (causal) algorithm may intervene on the symbolic state values between two subsequent calls of the step function, changing the game's behavior drastically from the outside.
Stochasticity is handled via \texttt{jax.random.PRNGKey}, which is defined as part of the game state and thus propagated through all steps. 

\textbf{Observations.}
\jaxatari natively supports two distinct observation modes, managed using wrappers:

\textbf{Object-Centric:} The core environment state naturally exposes structured data (\eg positions, velocities). The \texttt{ObjectCentricWrapper} flattens this information into a feature vector, enabling research into relational RL and agents that reason about object interactions directly, without the need for a separate vision backbone. It also only expose the observed objects (depicted on the screen), but hides internal variables (\eg respawning counters).\\
\textbf{Pixel renderings:} For compatibility with standard deep RL algorithms (\eg DQN, PPO), we provide a \texttt{JAXGameRenderer} for each environment. This differentiable renderer constructs the frames from the state variables using JAX operations, allowing fast convolutional agents training on GPUs.
Notably, both observation types can be retrieved simultaneously for neuro-symbolic approaches.
The object-centric observation is \jaxatari's native and thus most efficient instantiation, since it can directly be extracted from the state that is passed between transitions. 
The pixel-based observations require additional rendering functions that impact performance.
Despite this, \jaxatari is still significantly more efficient than ALE due to the missing necessity of emulating the hardware architecture, its ability to render game states only when necessary (typically every fourth frame is rendered) and the ability to natively downscale images (\cf~\Cref{appendix:native_down}).
The implications of the observation mode choice is analyzed in the experimental section.

%
\paragraph{Vectorized Accelerator Execution.}
One of the primary advantages of \jaxatari is its ability to leverage massive parallelization. Using \texttt{jax.vmap}, we vectorize the \texttt{step} function to simultaneously run thousands of environment instances on a single GPU.
This also avoids the data transfer overhead of CPU-based environment and GPU-hosted agent present in  ALE workflows.

\subsection{The Modification System}

A key goal of \jaxatari is its modular modification system, designed to test the agent alignment and generalization capabilities. 
Unlike HackAtari~\citep{delfosse2024hackatari}, which requires complex RAM manipulation, \jaxatari's codebase supports direct, programmatic modifications via a plugin system, making modifications authoring accessible. 
Furthermore, HackAtari can only modify RAM accessible parts of the state, mostly canceling environments' dynamics. 
For example, in Kangaroo, HackAtari \eg cancels the spawning of the monkeys. 
While \jaxatari also implements such simple modifications, it additionally allows switching the ladders positions (fixed in ALE and not stored in the game's RAM), changing the sprites, altering the reward system (\eg punishing the agent for violent actions such as punching), or modifying the environment dynamics (\eg turning bonuses into penalties).

To support this flexibility, \jaxatari includes two modification systems that intervene at different environment execution points.
Internal modifications are applied before the game environment is built, allowing researchers to efficiently alter core rules and assets—like gravity, enemy speeds, or reward structures—by deeply integrating them into the environment at construction time.
Conversely, post-step modifications adjust the environment's state immediately after each game step. This approach is ideal for making direct interventions that don't require rewriting the underlying game physics, such as instantly relocating an object, hiding specific variables, or forcing a game to end.
Modifications can be freely combined and grouped into reusable collections.

\subsection{Using \jaxatari for Benchmarking}
\label{sec:benchmark}

Having introduced \jaxatari as a flexible framework, we focus on its utility as a benchmark substrate to evaluate agent robustness under semantic distribution shifts and to diagnose reliance on spurious shortcuts.
Following the intervention-based protocol established by \citet{delfosse2025deep}, agents are trained exclusively on unmodified \jaxatari environments.
We subsequently evaluate their zero-shot transfer performance on controlled semantic variations that preserve the core task structure while systematically altering visual representations, object properties, game dynamics, or reward functions.

Discrepancies in performance under these distribution shifts serve as a proxy for the agents' generalization: identifying whether they have internalized robust, task-relevant causal structures or have instead converged on brittle heuristics tied to specific environment configurations.
For each game and variation, reports should contain the following metrics (defined formally in \Cref{appendix:metrics}).
\begin{enumerate}[leftmargin=15pt,itemsep=0pt,parsep=2pt,topsep=1pt,partopsep=2pt]

\item \textbf{Base Performance} ($\mathcal{P}_\text{base}$). The environmental return achieved by the agent trained on the unmodified environment.

\item \textbf{Modification Performance} ($\mathcal{P}_\text{mod}$). The return achieved by the base-trained agent when evaluated on the modified environment zero-shot.

\item \textbf{Performance Change} ($\mathcal{PC}$). The shift in performance across all evaluated games, quantifying the sensitivity to game modification.

\end{enumerate}

\section{Experimental Evaluation}
\label{sec:evaluation}

We evaluate \jaxatari along three axes: fidelity to ALE, accelerator-native scalability, and usefulness for controlled generalization studies.
\begin{enumerate}[leftmargin=31pt,itemsep=0pt,parsep=2pt,topsep=1pt,partopsep=2pt]
    \item[\textbf{(RQ1)}] {Does \jaxatari provide environments that are functionally equivalent to the original ALE?}
    \item[\textbf{(RQ2)}] {Can \jaxatari significantly improve training speeds?}
    \item[\textbf{(RQ3)}] {Does \jaxatari's modification system allow to probe agent alignment, robustness and shortcut learning behaviors?}
\end{enumerate}

\subsection*{Experimental Setup}
\label{sec:setup}

As standard deep reinforcement learning baseline, we employ PPO~\citep{Schulman2017ProximalPO}, a widely used policy-gradient method. 
For comparability with the ALE baselines, we employ the training and network configuration from the CleanRL PPO baseline~\citep{huang2022cleanrl}.
Unless stated otherwise, we follow standard Atari training conventions, including frame skipping, frame stacking, and preprocessing to $84 \times 84$ grayscale observations. 

We evaluate both raw environmental performance, reported as episodic returns, and performance change, which we analyze using the Interquartile Mean (IQM) with bootstrap confidence intervals. Throughput is measured in total steps per second (SPS).
\jaxatari experiments were run on a single NVIDIA A100 GPU with 40GB of memory, ALE ones on two AMD EPYC 7742 CPUs (128 cores, 256 threads). Unless specified, results are aggregated over $3$ random seeds, using $10$M steps per environment. See Appendix~\ref{appendix:training_setup} for more details regarding training and environment setup.

\paragraph{The \jaxatari-15 Benchmark.}
While \jaxatari covers many ALE games (45+,~\cf Appendix~\ref{appendix:environments}), we evaluate on a representative subset of 15 environments selected to span six core real-time control categories:
\textit{Shooters \& Space} (\textit{Asteroids}, \textit{Beam Rider}, \textit{Gravitar}, \textit{Phoenix}, \textit{Seaquest}),
\textit{Platformers \& Action} (\textit{Freeway}, \textit{Frostbite}, \textit{Kangaroo}),
\textit{Maze \& Pathfinding} (\textit{Ms.~Pac-Man}),
\textit{Paddle \& Ball} (\textit{Breakout}, \textit{Pong}),
\textit{Sports \& Racing} (\textit{Enduro}, \textit{Skiing}, \textit{Tennis}), and
\textit{Sparse Reward} exploration (\textit{Montezuma's Revenge}).
This selection isolates distinct RL challenges—from reactive motor control (\textit{Pong}, \textit{Breakout}) and delayed credit assignment (\textit{Skiing}) to sparse multi-room exploration (\textit{Montezuma's Revenge})—while intentionally omitting turn-based or static categories (board and gambling games) to focus on dynamic continuous-time interactions.
Each environment is validated against the corresponding ALE version to ensure behavioral consistency.
Next to task-fidelity to the original environments and scalability, we position \jaxatari as a benchmark for agent robustness. 
Thus, each game supports a set of pre-implemented modifications, which range from simple visual changes, such as altered colors or background elements, to more substantial gameplay variations. 
This allows evaluating not only base-task performance, but also the robustness of trained agents under controlled distribution shifts. 
Details of games and modifications are provided in \cref{appendix:game_mosaic}.

\subsection{Experimental Results}
\paragraph{\jaxatari games achieve learning task parity with ALE (RQ1).}
A central design principle of \jaxatari is prioritizing accelerator-native parallel throughput and transparent, source-level modification over bit-level, 1:1 ROM emulation.
Exact cycle-level emulation of legacy Atari 2600 hardware (such as TIA register timings and 6502 assembly constraints) imposes rigid execution bottlenecks on GPUs and precludes clean Pythonic interventions on game semantics.
Furthermore, deep RL policies are notoriously brittle; minor low-level implementation details or differing pseudo-random number generation prevent direct zero-shot policy transfer between independent simulators.
Consequently, \jaxatari is designed not as a drop-in emulator replacement, but to achieve \textit{behavioral and learning task parity} with the ALE.

\begin{figure}
    \centering
    \includegraphics[width=\linewidth]{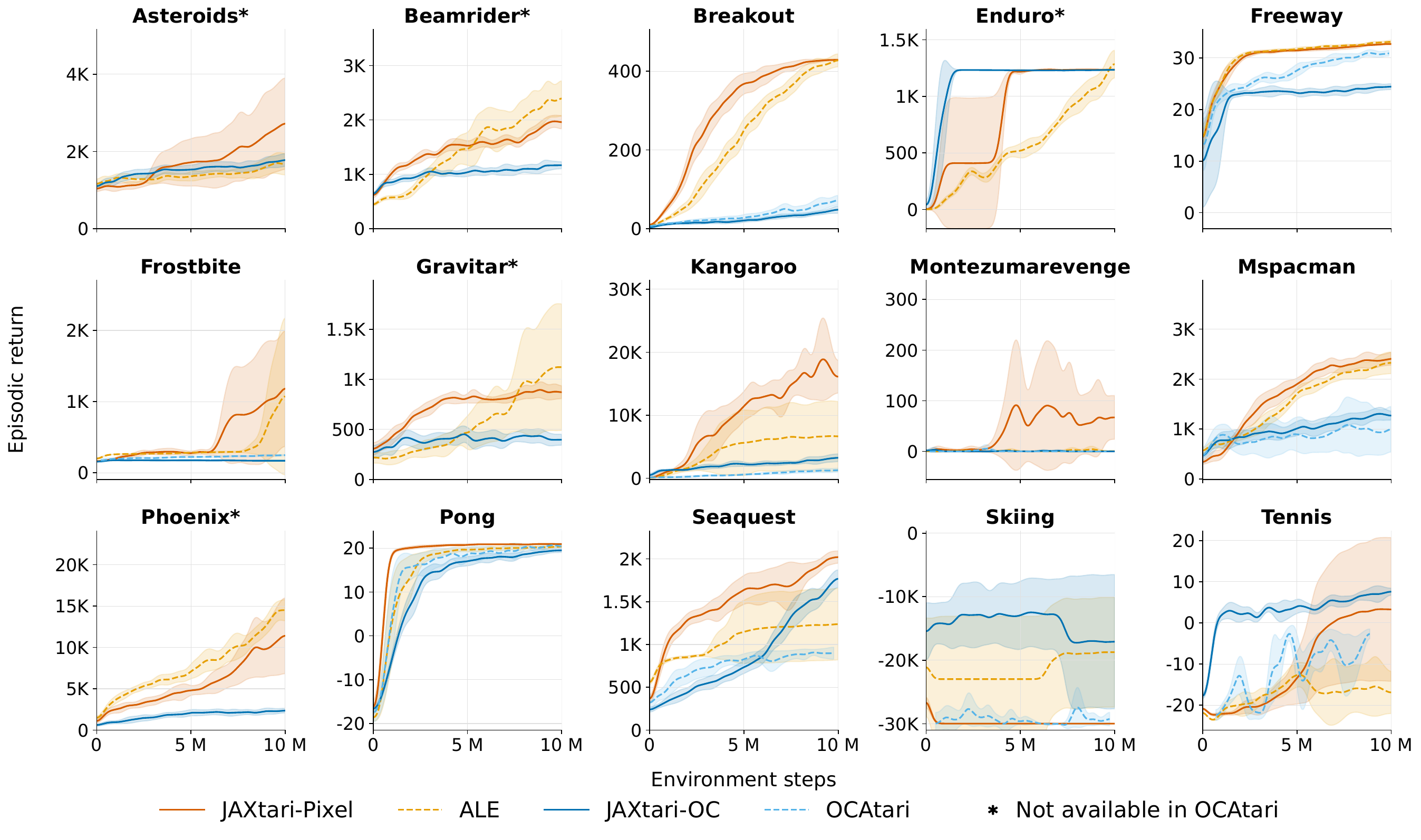}
    \caption{\textbf{\jaxatari enables similar learning behavior to ALE.} Learning curves of a PPO agent, learning in 15 \jaxatari and ALE games. Pixel-based \jaxatari (orange) behaves similar to ALE (green), object-centric observations (blue) differ more. Results are aggregated over $3$ seeds.}
    \label{fig:ppo_learning_curve_comparison}
\end{figure}

Under this criterion, environments must present equivalent learning dynamics and exploration challenges: tasks solved rapidly in ALE (\eg \textit{Pong}, \textit{Freeway}) are mastered similarly in \jaxatari, while sparse-reward exploration bottlenecks (\eg \textit{Montezuma's Revenge}, \textit{Skiing}) remain equally challenging under standard algorithms.
Moreover, agents discover and exploit identical behavioral shortcuts across both suites (\eg in \textit{Kangaroo} and \textit{Bankheist}).
As depicted in \Cref{fig:ppo_learning_curve_comparison}, pixel-based PPO runs (orange vs.\ green curves) confirm this alignment: learning curves follow comparable trajectories with asymptotic returns in the same order of magnitude.
In addition to pixel observations, \jaxatari natively provides object-centric (OC) states (\Cref{fig:ppo_learning_curve_comparison}, blue curves), allowing agents to train directly over structured entities without post-hoc extractors. 

To rigorously guarantee behavioral consistency across both suites despite independent JAX reimplementation, we employ a systematic \textit{Human + Agentic validation pipeline}.
First, an LLM agent extracts a ground-truth operational specification (active discrete action sets, entity dynamics, collision parameters, and reward formulas) by parsing official ALE documentation and original Atari 2600 manuals.
Second, an auditing agent cross-references the Python/JAX source code against this specification to identify implementation gaps (recall) and flag any extraneous logic (precision).
Third, human experts verify mapped code blocks and perform side-by-side gameplay validation against the original ALE ROMs to confirm intricate hardware edge cases.
Finally, policy training curves are evaluated to detect unintended shortcut exploitation.
A comprehensive walkthrough of this validation pipeline, including a case study on \textit{Frostbite}, is detailed in Appendix~\ref{appendix:validation_pipeline}. 

\begin{figure}[tb]
    \centering
    \begin{subfigure}[t]{0.48\linewidth}
        \centering
        \includegraphics[width=\linewidth]{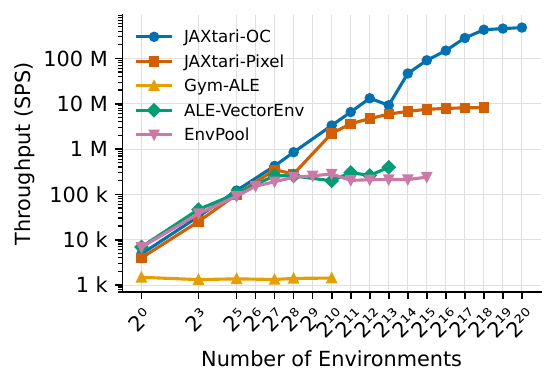}
        \caption{Scaling Pong to 1M Environments.}
        \label{fig:pong1M}
    \end{subfigure}
    \hfill
    \hfill
    \begin{subfigure}[t]{0.48\linewidth}
        \centering
        \includegraphics[width=\linewidth]{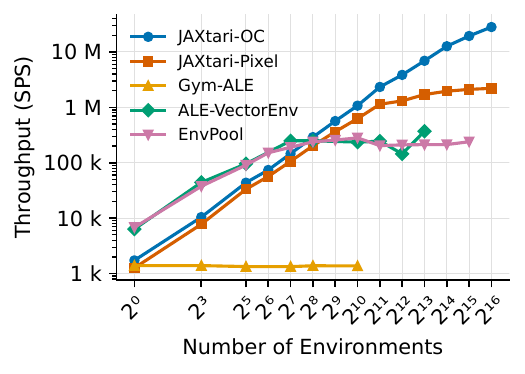}
        \caption{Average throughput comparison.}
        \vfill
        \label{fig:avg_throughput}
    \end{subfigure}

    \caption{\textbf{\jaxatari increases the throughput by enabling more parallel environments.} (a) The JAX-native \jaxatari implementation scales up to 1M parallel environments in Pong before running out of memory. (b) On average, across all games, \jaxatari increases throughput by 5$\times$ for pixel-based runs compared to ALE-VectorEnv and EnvPool. An in-depth analysis is available in Appendix~\ref{app:extended_results}.}
    \label{fig:problem2}
\end{figure}

\paragraph{\jaxatari achieves significantly higher throughput, reducing the needed training time (RQ2).}
The scalability properties of the \jaxatari-15 suite are validated against standard Gym-ALE, EnvPool~\citep{weng2022envpool} and ALE-VectorEnv, a recent EnvPool variation within the ALE.
\jaxatari-15 environments reach at least a throughput of $1$M steps-per-second compared to a maximum of $360$k for ALE-VectorEnv and a maximum of $280k$ for EnvPool (see \Cref{fig:problem2}~b). The \jaxatari environments significantly outscale CPU-based ALE with a single GPU. 
Some lower-complexity games, \eg, pong or freeway, reach numbers exceeding $100$M steps per second when using OC observations (\cf Figure~\ref{fig:pong1M}). 
The observed variations in throughput across the \jaxatari suite are attributable to heterogeneous visual and gameplay complexities, alongside implementation-specific nuances arising from our ground-up reimplementation of each game.
In general, OC observations exhibit significantly higher scalability and throughput, which we attribute to the ability to execute the environment logic exclusively without rendering, significantly reducing memory usage and computational complexity. This is an advantage over ALE, whose emulator does not allow to skip rendering.
Additionally, the smaller per-step memory footprint fits roughly $4\times$ as many parallel environments on GPU as in the pixel-based setting.

We also validate the utility of scaled-up \jaxatari runs by performing PPO experiments with a large number of parallel environments (8192 OC, 2048 RGB, \cf~\Cref{app:throughput}).
When translated to wall-clock time, PPO agents complete training on $100$M environment steps in under one hour when learning from pixels, and in under seventeen minutes using object-centric observations on a single A100 GPU (\cf Figures~\ref{fig:big_pixel}, \ref{fig:big_oc}).
This stands in stark contrast to the multi-hour runtimes required by equivalent CleanRL+ALE configurations.
This acceleration effectively collapses the experiment-iteration loop, \eg for uncovering novel shortcuts, reducing the experimental cycle from several days to a single working session.
Consequently, large-scale sweeps over game variants, which would typically induce long runtimes, have become computationally tractable on modest hardware.

\paragraph{JAXtari modification system allows shortcut detection (RQ3).} 
\begin{figure}[tb]
    \centering
    \begin{subfigure}[b]{0.48\textwidth}
        \centering
        \includegraphics[width=.85\linewidth]{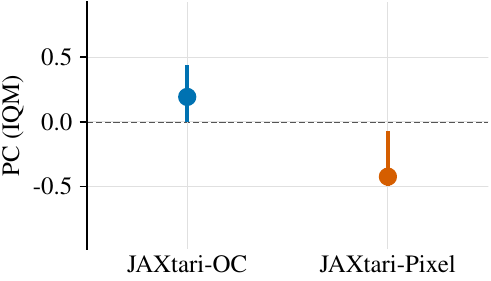}
        \caption{Signed performance change.}
        \label{fig:pc_signed}
    \end{subfigure}
    \hfill
    \begin{subfigure}[b]{0.48\textwidth}
        \centering
        \includegraphics[width=.85\linewidth]{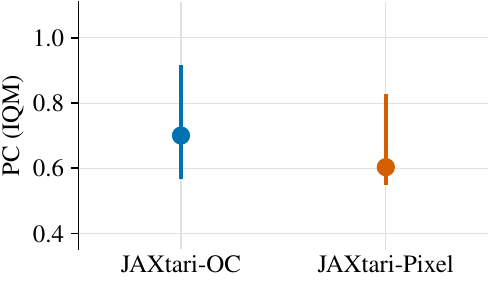}
        \caption{Absolute performance change.}
        \label{fig:pc_abs}
    \end{subfigure}
    \caption{\textbf{\jaxatari allows the detection of shortcuts by easily modifying the environment.} Performance change ($\mathcal{PC}$, defined in Appendix~\ref{appendix:metrics}) of PPO agents trained on the non-modified \jaxatari-15 environments and evaluated zero-shot on their modified counterparts, IQM-aggregated across $3$ seeds with $95\%$ stratified bootstrap confidence intervals.
    \textbf{(a)} Signed $\mathcal{PC}$: values below zero indicate that the agent performs worse on the modified environment than on the base version; values above zero indicate (rare) improvement.
    \textbf{(b)} Absolute $\mathcal{PC}$: magnitude of deviation regardless of direction.
    These experiments extend the systematic performance drops of standard, model-free RL agents on task simplifications, initially exposed by~\cite{delfosse2025deep} to a more diverse set of modifications.
    Per game and mod results are depicted in Appendix~\ref{app:robustness}.
    }
    \label{fig:performance_change}
\end{figure}

The pre-implemented game modifications (\cf~Appendix~\ref{appendix:environments}) allow us to analyze an agents robustness to various modifications (e.g., visual changes, dynamics simplifications). For this, we follow the proposed benchmark detailed in Section~\ref{sec:benchmark} and perform zero-shot adaptation evaluation on each modification.
The results, depicted in Figure~\ref{fig:performance_change}, suggest that agents converged on the original games systematically lose performance under modifications, even on simplifications.
The signed performance change ($\mathcal{PC}$) is negative across the majority of the game-modifications and the absolute $\mathcal{PC}$ stays well above zero, indicating that the drops are consistent collapses. 
Game simplifications and sprite swaps that preserve gameplay are sufficient to break pixel-based agents on several games, mirroring the brittleness reported in~\citet{delfosse2025deep}, confirming \jaxatari's utility in evaluating robustness and uncovering shortcuts.

Beyond exposing these vulnerabilities, \jaxatari provides the essential substrate to actively mitigate shortcut learning. 
Because environment parameters and visual assets are exposed directly in Python and natively vectorizable under \texttt{jax.vmap}, researchers can perform extensive domain randomization across game dynamics and visual features during training with zero simulation overhead. 
Furthermore, \jaxatari's native object-centric state representations offer the structural scaffolding required to develop and evaluate neuro-symbolic methods—such as NUDGE~\citep{Delfosse2023InterpretableAE}, BlendRL~\citep{shindo2024blendrl}, and Kintsugi~\citep{cao2026kintsugi}—which explicitly ground policy learning in relational concepts and allow repairing brittle behavioral shortcuts. 
Finally, exposing structured semantic states facilitates interfacing multimodal foundation models and code-as-policy architectures (\eg Atari-GPT~\citep{waytowich2024atarigpt}, RL-GPT~\citep{liu2024rlgpt}) to test whether external commonsense priors yield true zero-shot generalization across environmental perturbations.

\section{Related Work}

\jaxatari builds on three lines of RL research: Atari benchmarks, accelerator-native simulators, and generalization benchmarks.
Prior work has improved Atari along individual axes such as speed, stochasticity, editability, or state access, but these capabilities have typically remained separate.

\begin{wrapfigure}[24]{r}{.48\linewidth}
    \centering
    \vspace{-5mm}
    \includegraphics[width=0.97\linewidth]{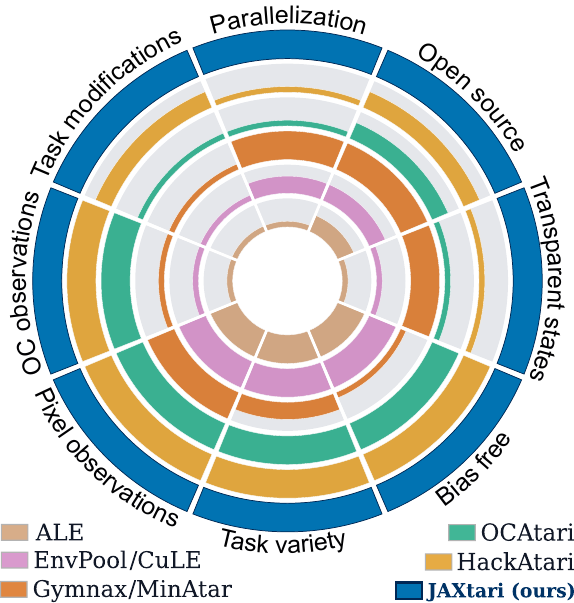}
    \caption{\textbf{Feature coverage across Atari-based reinforcement learning frameworks.} JAXtari (ours) is the only library to support the full spectrum of features, including GPU parallelization, diverse observation modes (Pixel and OC), and extensive task modification capabilities.}
    \label{fig:sun_burst}
\end{wrapfigure}

\paragraph{The Arcade Learning Environment and its extensions.}
The ALE~\citep{bellemare2013arcade} remains the canonical Atari benchmark, and many extensions address specific limitations of the original setup.
Sticky actions and action noise reduce determinism~\citep{machado2018revisiting, koch2021objective}, while \textit{MinAtar}~\citep{young2019minatar} provides simplified Atari-like games for faster experimentation.
Other extensions preserve the Atari setting but target different bottlenecks: \textit{HackAtari}~\citep{delfosse2024hackatari} enables semantic interventions through RAM manipulation, \textit{CuLE}~\citep{dalton2020accelerating} and \textit{EnvPool}~\citep{weng2022envpool} accelerate environment execution, and \textit{AtariARI}~\citep{anand2019unsupervised} and \textit{OCAtari}~\citep{delfosse2024ocatari} expose structured state on top of ALE.
These systems expand Atari's utility, but typically improve one axis at a time.
\jaxatari instead reimplements Atari-style games directly in JAX, making the environments simultaneously accelerator-native, semantically editable, pixel-compatible, and object-centric by design. We compare \jaxatari to ALE and the above-mentioned extensions in Figure~\ref{fig:sun_burst}.

\paragraph{Accelerator-native RL environments.}
Recent RL benchmarks increasingly move environment execution into JAX or GPU-native backends to reduce simulation bottlenecks and support large-scale vectorized training.
Examples include \textit{Brax}~\citep{freeman2021brax} for continuous control, \textit{Gymnax}~\citep{gymnax2022github} for classic control and lightweight environments, \textit{Jumanji}~\citep{bonnet2024jumanji} for structured decision-making, \textit{Pgx}~\citep{koyamada2023pgx} for board games, and \textit{JAX-MARL}~\citep{flair2024jaxmarl} for multi-agent RL.
Game-based suites such as \textit{Craftax}~\citep{matthews2024craftax} and \textit{Octax}~\citep{radji2026octax} further show the value of scalable JAX-native simulation.
These frameworks share \jaxatari's systems motivation, but target domains other than ALE-style Atari gameplay or simplified arcade settings.
Atari remains a milestone benchmark because it combines visual input, long-horizon control, diverse game mechanics, and a long history of comparable baselines.
Bringing this benchmark class into an accelerator-native, editable, and object-centric setting is therefore a necessary step for testing modern RL agents at Atari-like complexity.

\paragraph{Generalization and Robustness in RL.}
Generalization remains a central challenge in RL: agents can achieve high returns in fixed environments while failing under small distribution shifts~\citep{moos2022robust}.
Procedural benchmarks such as \textit{CoinRun}~\citep{cobbe2019quantifying} and \textit{Procgen}~\citep{cobbe2020procgen} expose agents to diverse generated levels, while \textit{XLand-MiniGrid}~\citep{nikulin2023xlandminigrid} and \textit{Craftax}~\citep{matthews2024craftax} provide scalable task diversity.
Another line improves robustness through visual augmentation or representation learning, including RAD~\citep{laskin2020rad}, DrQ~\citep{yarats2021image}, and the combination of pixel-based and object-centered representations~\citep{bluml2025deep} or environment perturbations~\citep{robustrl2024}.
These approaches broadly improve or evaluate robustness, but do not isolate which semantic assumptions an agent has learned.
Targeted semantic interventions address this diagnostic question more directly.
HackAtari~\citep{delfosse2024hackatari} introduced controlled modifications to ALE games through RAM manipulation, and subsequent work showed that deep RL agents can fail even when task variants become easier for humans~\citep{delfosse2025deep}.
\jaxatari builds on this intervention-based view, but moves it from opaque RAM manipulation in a CPU-bound emulator to source-level modifications of JAX game logic.
This makes counterfactual evaluation easier to implement, compositional across modifications, and scalable to large experimental sweeps.

\paragraph{Object-centric representations and neuro-symbolic RL.}
Learning structured, object-level representations is an established paradigm for compositional generalization and sample efficiency in RL~\citep{diuk2008object,garnelo2016towards}.
While prior Atari tools such as AtariARI~\citep{anand2019unsupervised} and OCAtari~\citep{delfosse2024ocatari} extracted object states post-hoc from emulator RAM, they remained constrained by CPU-bound stepping and lacked native integration with GPU training loops.
Natively exposing object states unlocks methods that incorporate relational inductive biases and neuro-symbolic reasoning.
Frameworks such as NUDGE~\citep{Delfosse2023InterpretableAE} induce differentiable first-order logic rules over extracted entities, BlendRL~\citep{shindo2024blendrl} unifies neural and symbolic policy learning, and Kintsugi~\citep{cao2026kintsugi} repairs misaligned or brittle policies by directly editing executable knowledge bases.
Furthermore, structured object representations bridge the semantic gap to foundation models, enabling multimodal large language models and code-as-policy systems (\eg Atari-GPT~\citep{waytowich2024atarigpt}, RL-GPT~\citep{liu2024rlgpt}) to interpret Atari scenes with human-like commonsense priors.
\jaxatari provides the first accelerator-native environment suite where such object-centric and neuro-symbolic architectures can be trained, verified, and stress-tested at modern GPU throughput.

\paragraph{Positioning of \jaxatari.}
While prior research has independently established the necessity of Atari as a reasoning benchmark, the scalability of accelerator-native simulation, and the diagnostic power of object-centric states and controlled interventions, these capabilities have largely remained fragmented across distinct frameworks. 
\jaxatari resolves this fragmentation by fundamentally reconstructing Atari-style game logic directly in JAX, bypassing the severe bottlenecks of legacy CPU-bound emulation. Through this purely functional paradigm, agents can be trained from standard pixel renderings or evaluated using natively exposed object-centric features, effectively bridging the gap between deep and relational reinforcement learning. 
Furthermore, the framework leverages massive GPU parallelization for highly efficient scaling, and its modular API allows agents to be tested under direct, source-level semantic modifications—such as shifting dynamics, rules, or visual properties—without relying on complex and opaque RAM manipulation. 
Ultimately, this unification makes \jaxatari a comprehensive benchmark substrate for studying not only how well agents perform, but precisely what causal structures or brittle heuristics they rely on when they succeed.

\section{Limitations and Future Work}
\label{sec:limitation}
\jaxatari is intended as a scalable and editable benchmark substrate for Atari-style RL, but it has limitations.
First, reimplementing Atari games from scratch is substantially more labor-intensive than wrapping an existing emulator.
As a result, the current release covers a subset of the full ALE suite rather than the typically used 57 games.
We prioritize commonly used games and provide validation for the environments included in the benchmark, but the suite should not yet be interpreted as a complete replacement for ALE.
Furthermore, \jaxatari does not cover the ALE mode and difficulty switches of the original ROMs which alter core game parameters but instead relies on an independent modification system to introduce similar task-variance.
Expanding coverage across the full Atari benchmark remains an important direction for future work.

Second, \jaxatari aims to reproduce ALE-style gameplay, but it is not an emulator and therefore cannot guarantee bit-level equivalence with the original Atari ROMs.
Small divergences in timing, collision handling, object behavior, or rare edge cases may remain, in part due to differences in random number generation.
We mitigate this by using visual comparisons, per-game divergence measurements, and downstream learning-curve comparisons, but users should interpret \jaxatari as a high-fidelity reimplementation designed for scalable experimentation and controlled intervention, rather than a drop-in substitute for exact emulator-based evaluation.
Third, the flexibility of the modification system introduces a benchmark-design challenge.
Not every modification defines a meaningful generalization test: some preserve the task structure, while others may change the objective, difficulty, or optimal policy in ways that make comparisons ambiguous.
For this reason, benchmark results should distinguish between visual shifts, dynamics shifts, reward shifts, simplifications, and more difficult out-of-distribution variants.
Each existing modification is categorized by its intended semantic category and the expected effect.
Next to increasing the number of available environments, future releases will include a larger curated modification library.

Future work will focus on three directions.
First, we plan to expand the number of supported games and broaden coverage of the ALE benchmark.
Second, we will extend the library of validated semantic variations and integrate them with benchmark protocols for robustness, shortcut learning, controlled generalization and continual RL.
Third, \jaxatari's explicit JAX state and source-level transition functions open opportunities for model-based RL, planning, world modeling, and curriculum generation.
Although Atari's discrete actions and collision-heavy dynamics limit straightforward gradient-based policy optimization, the JAX-native implementation makes it easier to inspect, transform, batch, and potentially differentiate through selected components of the environment.

\section{Conclusion}
\label{sec:conclusion}
We introduced \jaxatari, a JAX-native framework and benchmark substrate for scalable, editable Atari-style reinforcement learning.
By reimplementing game logic directly in JAX, \jaxatari removes the CPU-bound emulator bottleneck of standard ALE workflows, enables accelerator-native vectorized execution, and exposes native object-centric state alongside pixel observations.
Its source-level modification system further enables controlled semantic variations, allowing agents trained on standard Atari-style tasks to be evaluated under targeted visual, dynamical, reward-level, and object-level shifts.
Our results show that \jaxatari preserves ALE-style learning behavior while providing substantially higher throughput and making counterfactual evaluation practical at scale.
Together, these properties make \jaxatari a benchmark substrate for studying not only how well RL agents perform, but also whether their behavior reflects robust task understanding or brittle shortcut reliance.
By reducing experiments that previously required many hours of emulator-bound execution to substantially shorter accelerator-native runs, \jaxatari makes systematic robustness evaluation over many agents, seeds, and environment variants substantially more accessible.

\paragraph{Broader Impact.}
\jaxatari is an open-source benchmark substrate for academic reinforcement learning research.
Its primary positive impact is democratizing large-scale rl experimentation: by reducing the compute required to run meaningful experiments from many CPU-hours to short GPU-native runs, it lowers the barrier for researchers. 
The modification system additionally makes systematic robustness and shortcut-learning evaluation accessible, which we expect to benefit the broader community studying agent generalization. 
We do not foresee significant negative societal impacts; \jaxatari releases no pretrained agents, collects no user data, and involves no human subjects.

\paragraph{Acknowledgements.}
This research work has been funded by the Deutsche Forschungsgemeinschaft (DFG, German Research Foundation) under Germany’s Excellence Strategy (EXC-3057/1 ”Reasonable Artificial Intelligence“, Project No. 533677015). The authors also thank Lars Teubner and the students at TU Darmstadt who participated in the practical lab implementing \jaxatari environments.


\newpage
\appendix
\onecolumn


\section*{Appendix Overview}
\vspace{-0.5em}
\begin{itemize}
    \item[\textbullet] \textbf{Appendix~\ref{appendix:code_data}:} \textit{Code and Data}\\
    Describes access to JAXtari environments, modifications, data and experimental pipeline.

    \item[\textbullet] \textbf{Appendix~\ref{appendix:validation_pipeline}:} \textit{Human + Agentic Validation Pipeline and Parity Case Study}\\
    Protocol for guaranteeing behavioral fidelity to ALE, featuring the Frostbite verification checklist and documented hardware divergences.

    \item[\textbullet] \textbf{Appendix~\ref{appendix:metrics}:} \textit{Metrics}\\
    Definitions of Interquartile Mean (IQM), and performance change metrics.

    \item[\textbullet] \textbf{Appendix~\ref{appendix:training_setup}:} \textit{Agent Architectures and Training Setup}\\
    Details hardware setup, GPU usage, training time, and total inference budget required for evaluation.

    \item[\textbullet] \textbf{Appendix~\ref{app:extended_results}:} \textit{Extended Results}\\
    Provides extended results and tables for deep agents, performance comparisons, and modification evaluations.

    \item[\textbullet] \textbf{Appendix~\ref{appendix:environments}:} \textit{JAXtari Game Descriptions and Modifications}\\
    Full documentation of available game environments and all implemented task variances.

\end{itemize}
\vspace{1em}


\newpage

\section{Code and Data Availability}
\label{appendix:code_data}

\subsection{Artifact Overview}

We release \jaxatari as an open-source software benchmark for scalable and editable Atari-style reinforcement learning.
The artifact includes JAX-native environment implementations, pixel-based and object-centric observation interfaces, Gymnasium-compatible wrappers, a source-level modification API, predefined modification plugins, and scripts for reproducing the main experiments.
Detailed evaluation settings are provided in Appendix~\ref{appendix:training_setup}, and the full list of supported games and modifications is provided in Appendix~\ref{appendix:environments}.

The full source code, benchmark suite, and documentation are publicly available at:
\begin{center}
\url{https://github.com/k4ntz/JAXtari}
\end{center}

\subsection{License and Intellectual Property}

\jaxatari is a bottom-up reimplementation of Atari-style environments in JAX.
It does not distribute or execute proprietary Atari ROMs.
The released code, wrappers, benchmark configurations, and modification plugins are provided under the \texttt{MIT} license. 

\jaxatari is not affiliated with, endorsed by, or sponsored by Atari or the original game rights holders.
We handle visual assets analogously to ALE's handling of ROMs: \jaxatari does not distribute proprietary Atari assets by default, but provides the infrastructure to load user-supplied or license-compatible assets when available.
Users who compare against ALE or original Atari ROMs are responsible for ensuring that their use of third-party ROMs, assets, or emulators complies with the relevant licenses and local regulations.

\subsection{Reproducibility Statement}
To facilitate reproducibility, we make the code publicly available.
All scripts used to train and evaluate agents are included in the repository; We also provide a complete description of the directory structure and instructions for running each experiment. The exact configuration files used for each experiment are part of the codebase, allowing exact reproduction of all reported runs. Experiments were averaged over 3 seeds; all scripts enforce fixed seeding to ensure deterministic results. Compute resources are disclosed in Appendix~\ref{appendix:training_setup}, which also details the exact hyperparameters, hardware, and training protocols.

All experiment scripts log the environment name, observation mode, modification name, agent type, random seed, episode return, episode length, wall-clock time, software version, and hardware configuration. Because \jaxatari is a software benchmark, reproducibility also depends on stable releases and regression testing; we therefore provide versioned benchmark releases, fixed configurations for reported baselines, and tests covering environment transitions, rewards, rendering, wrappers, and modification behavior. Substantive changes to game dynamics or evaluation protocols will be documented in the changelog and released under a new version number.

\clearpage
\subsection{Usage}

\paragraph{Installation.}
The package can be installed from source as follows:
\begin{codebox}\input{code_listings/listing1.pygtex}\end{codebox}
A complete list of dependencies is provided in \texttt{pyproject.toml}.

\paragraph{Basic usage.}
A minimal example for creating and stepping an environment is:
\begin{codebox}\input{code_listings/listing2.pygtex}\end{codebox}
%
%
%

Modifications can be injected at environment creation:
\begin{codebox}\input{code_listings/listing3.pygtex}\end{codebox}

The same interface can be used with vectorized JAX execution, allowing users to batch environment instances with \texttt{jax.vmap} and compile rollouts with \texttt{jax.jit}.

\newpage
\section{Human + Agentic Validation Pipeline and Implementation Verification}
\label{appendix:validation_pipeline}

Reimplementing Atari 2600 game mechanics from scratch in a pure-JAX functional paradigm requires rigorous quality control to ensure learning task parity with the original Arcade Learning Environment (ALE)~\citep{bellemare2013arcade}.
Because Atari ROMs were originally written in 6502 assembly with tight hardware constraints (Television Interface Adapter / TIA clock timings), subtle gameplay mechanics and edge cases can easily be missed in high-level reimplementations.
To systematically prevent behavioral discrepancies, we established a four-stage \textbf{Human + Agentic validation pipeline} combining autonomous specification extraction, automated code auditing, human-in-the-loop verification, and empirical policy stress-testing.

\subsection{The Four-Stage Validation Protocol}

\begin{enumerate}[leftmargin=*]
    \item \textbf{Stage 1: Autonomous Specification Extraction (Agentic).}
    For every environment, an LLM agent extracts an exhaustive ground-truth operational specification by reading official documentation from the Farama Foundation (\texttt{ale.farama.org}) and canonical AtariAge manuals (\texttt{atariage.com}). This structured specification document formalizes:
    \begin{itemize}
        \item \textit{Action Space:} The active subset of the 18 discrete Atari actions, identifying which inputs induce positional movement, fire projectiles, or trigger special mechanics.
        \item \textit{Game Mechanics:} Complete entity lifecycle, including spawn conditions, directional velocity, screen boundaries, multi-level progression, and collision consequences.
        \item \textit{Reward Structure and Termination:} Explicit scoring criteria (points per entity hit or collectable picked up), milestone bonuses, life counters, and game-over triggers.
    \end{itemize}

    \item \textbf{Stage 2: Code Mapping and Gap Auditing (Agentic).}
    A secondary auditing LLM agent parses the JAX implementation against the generated specification:
    \begin{itemize}
        \item \textit{Recall Audit:} Maps each specified game mechanic and scoring rule to exact line numbers and function definitions in the Python/JAX source code, ensuring that no specified rule was omitted.
        \item \textit{Precision Audit:} Examines all unmapped code paths to identify undocumented logic, hardcoded constants, or custom mechanics not found in official manuals.
    \end{itemize}

    \item \textbf{Stage 3: Dual Ground-Truth Verification (Human-in-the-Loop).}
    Human expert verifiers evaluate the audit outputs along two complementary dimensions:
    \begin{itemize}
        \item \textit{Recall Verification:} Experts review the mapped code blocks to confirm that the mathematical dynamics and coordinate transforms accurately match Atari mechanics.
        \item \textit{Precision Verification:} For any undocumented code behavior flagged in the precision audit, human verifiers play the original ALE ROM side-by-side with the JAXtari environment. This step confirms whether the logic reflects an authentic assembly-level edge case of the original game or represents an unintended implementation artifact.
    \end{itemize}

    \item \textbf{Stage 4: Empirical Policy Stress-Testing.}
    Finally, standard RL agents (CleanRL PPO) are trained on both JAXtari and ALE under identical hyperparameters. Learning curve trajectories, asymptotic returns, and exploration rates are compared to guarantee that the JAX environment presents an equivalent learning challenge and does not permit unintended shortcut strategies.
\end{enumerate}

\subsection{Case Study: Validation Checklist for \textit{Frostbite}}

To demonstrate the efficacy of this validation protocol, \Cref{tab:frostbite_action_mechanics,tab:frostbite_rewards} present the verified specification mapping and human audit for \textit{Frostbite}.

\begin{table}[htbp]
\centering
\small
\caption{\textbf{\textit{Frostbite} Action Space and Core Dynamics Validation Checklist.} Mapped against the JAXtari implementation and verified against ALE.}
\label{tab:frostbite_action_mechanics}
\begin{tabular}{p{3.2cm} p{8.0cm} p{2.8cm}}
\toprule
\textbf{Mechanic} & \textbf{Operational Specification} & \textbf{Source Implementation} \\
\midrule
\multicolumn{3}{l}{\textit{Action Space Coverage}} \\
Full Discrete Space & Utilizes full \texttt{Discrete(18)} action space; \texttt{NOOP}, \texttt{FIRE}, directional jumps, and combined inputs are supported. & \texttt{jax\_frostbite.py} \\
Directional Controls & Moving joystick triggers Bailey's jumps across moving ice floes and between shores. & \texttt{jax\_frostbite.py} \\
\texttt{FIRE} Button & Pressing \texttt{FIRE} reverses the movement direction of the ice row currently occupied. & \texttt{jax\_frostbite.py} \\
\midrule
\multicolumn{3}{l}{\textit{Core Game Dynamics}} \\
Igloo Construction & Landing on a white ice floe turns it blue and adds one block to the igloo (15 blocks required for completion). & \texttt{jax\_frostbite.py} \\
Floe Resetting & Once all ice blocks in a single row are turned blue, the entire row reverts back to white. & \texttt{jax\_frostbite.py} \\
Reversal Penalty & Using \texttt{FIRE} to reverse row direction deducts one uncompleted block from the igloo. & \texttt{jax\_frostbite.py} \\
\midrule
\multicolumn{3}{l}{\textit{Precision Audit Finding (Authentic Atari Edge Case)}} \\
Free Reversals & \textit{Auditor note:} When the igloo is complete (15 blocks), subsequent \texttt{FIRE} reversals incur no block deduction penalty. \textit{Human verification:} Verified in original ALE ROM; a completed igloo cannot be diminished. & \texttt{jax\_frostbite.py} \\
\bottomrule
\end{tabular}
\end{table}

\begin{table}[htbp]
\centering
\small
\caption{\textbf{\textit{Frostbite} Reward System Validation Checklist.}}
\label{tab:frostbite_rewards}
\begin{tabular}{p{3.5cm} p{8.0cm} p{2.5cm}}
\toprule
\textbf{Scoring Event} & \textbf{Reward Formula} & \textbf{Source Implementation} \\
\midrule
Coloring an Ice Block & $10 \times \text{current level}$ points awarded upon stepping on a white floe. & \texttt{jax\_frostbite.py} \\
Entering Completed Igloo & $160 \times \text{current level}$ bonus points awarded per block upon entering the igloo to clear the level. & \texttt{jax\_frostbite.py} \\
Level Multiplier Cap & Level multiplier increases up to level 8; from level 9 onward, scoring caps at 90 points per ice block. & \texttt{jax\_frostbite.py} \\
Catching a Fish & Flat bonus of $+200$ points upon collision with a swimming fish. & \texttt{jax\_frostbite.py} \\
\bottomrule
\end{tabular}
\end{table}

\subsection{Documented Implementation Differences Between JAXtari and ALE}
\label{appendix:implementation_differences}
While the validation pipeline guarantees learning task parity, several fundamental differences arise from modernizing Atari for accelerator-native execution:
\begin{itemize}[leftmargin=*]
    \item \textbf{Explicit PRNG Keying vs.\ Implicit Hardware Registers:} ALE generates pseudo-randomness from uninitialized memory and Stella hardware counter side-effects. \jaxatari uses functional JAX PRNG splitting (\texttt{jax.random.split}), ensuring strict mathematical determinism across massive parallel threads.
    \item \textbf{Pure Functional State vs.\ Emulated RAM:} In ALE, object coordinates must be inferred by decoding opaque RAM bytes. In \jaxatari, state is maintained as typed PyTrees of JAX arrays, guaranteeing exact floating/fixed-point entity bounds.
    \item \textbf{Floating-Point Coordinate Accumulation:} To prevent GPU branching bottlenecks, continuous entity velocities are vectorized across arrays rather than relying on discrete sub-pixel scanline counters.
\end{itemize}

\newpage
\section{Evaluation Metrics and Computational Benchmarking}
\label{appendix:metrics}

We assess agent performance and framework efficiency using common metrics designed to provide statistically rigorous insights into learning dynamics and hardware utilization.



\paragraph{Reliable Performance Aggregation.} 
To mitigate the impact of outliers and account for the high variance inherent in RL training, we adopt the statistical framework proposed by \texttt{rliable}~\citep{agarwal2021deep}. We aggregate final performance using the Interquartile Mean (IQM), which computes the mean over the middle 50\% of the data distribution (i.e., the 25th to 75th percentiles). This provides a more robust estimate of central tendency than the grand mean or median. Furthermore, we report 95\% stratified bootstrap confidence intervals to quantify uncertainty. For longitudinal analysis, learning curves represent the mean in-game rewards accompanied by the standard deviation across 3 independent seeds.

\paragraph{Performance Change ($\mathcal{PC}$).} 
To quantify the impact of our environmental perturbations on agent proficiency, we define the Performance Change metric \cite{delfosse2025deep}. This metric assesses the relative shift in performance when an agent trained on a standard environment is evaluated on a modified variant. It is formulated as:
\begin{equation}
\mathcal{PC} = \frac{(S_{\text{mod}} - R_{\text{mod}})-(S_{\text{org}} - R_{\text{org}})}{|S_{\text{org}} - R_{\text{org}}|}
\end{equation}
where $S_{\text{mod}}$ is the score in the modified environment, $S_{\text{org}}$ is the score in the original environment, and $R$ the random baselines. A value of $\mathcal{PC} = 0$ indicates perfect performance invariance, while $\text{PC} < 0$ signifies a collapse in the perturbation.

\paragraph{Throughput and Hardware Efficiency.} 
The computational utility of JAXtari is quantified via environmental throughput, defined as the total number of frames processed per second across all parallel environments. A primary architectural objective of our JAX-native implementation is to maximize hardware efficiency, targeting 100\% GPU utilization by eliminating CPU-GPU communication bottlenecks that typically constrain traditional emulators. To provide a transparent account of the computational costs associated with our benchmarks, we track and report total GPU hours. Also, all experiments were done on a single A100 GPU, while scaling with multiple GPUs is possible.


\newpage
\let\oldfootnotesize\footnotesize
\renewcommand*{\footnotesize}{\oldfootnotesize\scriptsize}
\section{Agent Architectures and Training Setup}
\label{appendix:training_setup}
We evaluate the \jaxatari-15 environments on PPO (Proximal Policy Optimization) with both pixel-based and object-centric observation types. Whenever possible the implementations are JAX-based for maximum performance (jitted and scanned stepping through the environment). Note that \jaxatari can also be employed with non JAX implementations, e.g., pytorch or tensorflow (using DLPack~\footnote{\url{https://dmlc.github.io/dlpack/latest/}} for zero-copy tensor transfer). We also provide a gym-wrapper for maximum compatibility.
The network architecture and hyperparameters are adapted from the same sources as the baselines we compare {\jaxatari}'s environments against to ensure that discrepancies in agent performance can, with high certainty, be attributed to environment differences and not the training setup. Therefore PPO is adapted from CleanRL~\citep{huang2022cleanrl}.

\subsection{Baseline environments: ALE, EnvPool and ALE-VectorEnv}
\label{appendix:baselines}
We evaluate our environments against the ALE~\citep{bellemare2013arcade} which serves as the standard CPU-based framework for running Atari 2600 games as RL environments. However, the ALE executes environments sequentially, incurring throughput bottlenecks. Since our environments are highly parallelizable, we also evaluate throughput against EnvPool~\citep{weng2022envpool} and ALE-VectorEnv. Both are high-performance, native C++ multi-threaded implementations for running Atari2600 games.

Unfortunately, CuLE~\citep{dalton2020accelerating} is currently unmaintained. We made extensive efforts to compile and evaluate the framework locally; however, following the official installation instructions did not result in usable states. Reproducing the environment on modern hardware proved unfeasible due to three well-documented technical blockers: (1) strict dependencies on deeply deprecated software stacks, specifically CUDA 10.0 and PyTorch 1.2\footnote{\url{https://github.com/NVlabs/cule/issues/4} and \url{https://github.com/NVlabs/cule/issues/20}}, (2) fatal compilation failures when building the C++ source with modern GCC compilers or the provided Docker containers\footnote{\url{https://github.com/NVlabs/cule/issues/34} and \url{https://github.com/NVlabs/cule/issues/28}}, and (3) unresolved assertion crashes during initialization on multi-GPU systems for users who manage to bypass the build errors\footnote{\url{https://github.com/NVlabs/cule/issues/5}}. Because we could not establish a reliable, non-deprecated baseline for this hardware-accelerated framework, we focus our throughput evaluations on the ALE, EnvPool and ALE-VectorEnv.

\subsection{Preprocessing}
Both agents share the same preprocessing parameters, which directly match 
the standard ALE preprocessing used in the CleanRL PPO baseline\footnote{\url{https://docs.cleanrl.dev/rl-algorithms/ppo/}}, 
ensuring fair comparison.

For \textbf{pixel-based} observations:
\begin{itemize}
    \item \textbf{Grayscale} pixel-wise
    \item \textbf{Resize} via native downscaling to $84 \times 84$
    \item \textbf{Frame skip} 4 frames
    \item \textbf{Max pooling} over last two frames
    \item \textbf{Frame stack} of 4
    \item \textbf{Normalization} of pixels through division by 255 
    \item \textbf{Reward clipping} to $\{-1, 0, +1\}$
\end{itemize}

For \textbf{object-centric} observations:
\begin{itemize}
    \item \textbf{Frame skip} 4 frames
    \item \textbf{Frame stack} of 4
    \item \textbf{Normalization} to $[0, 1]$ using feature specific bounds
    \item \textbf{Flatten} features into 1D vector
    \item \textbf{Reward clipping:} to $\{-1, 0, +1\}$
\end{itemize}

In both modalities:
\begin{itemize}
    \item \textbf{Noop reset} executes up to 30 random no-ops on episode reset
    \item \textbf{Episodic life} terminates current episode on life loss 
    \item \textbf{Sticky actions} currently disabled ($p = 0.0$)
\end{itemize}

\subsubsection{Native Downscaling}
\label{appendix:native_down}
The aforementioned native downscaling is a feature of \jaxatari's rendering approach. Since the rendering is not directly tied to the game logic, but a separate computational step that constructs the image from a given state some preprocessing operations can be performed directly inside the renderer instead of afterwards on the finished image.
Concretely, rather than rendering the full 210x160 pixel image, the renderer scales down all sprites at initialization and construct the image at the target resolution.
This reduces memory consumption and increases throughput significantly, as operating on smaller images lowers the computational cost.

\subsection{Network Architecture}
\subsubsection{Pixel-Based Architectures}
For pixel-based observations, PPO uses a standard DQN-like convolutional neural network~\citep{mnih2015human} consisting of three convolutional layers followed by a fully connected layer:
\begin{itemize}
    \item \textbf{Conv1:} 32 filters, $8 \times 8$ kernel, stride 4, ReLU
    \item \textbf{Conv2:} 64 filters, $4 \times 4$ kernel, stride 2, ReLU
    \item \textbf{Conv3:} 64 filters, $3 \times 3$ kernel, stride 1, ReLU
    \item \textbf{FC:} 512 units, ReLU
\end{itemize}

\subsubsection{Object-Centric Architecture}

For object-centric observations the agents use a fully-connected architecture instead of convolutional layers used for pixel-based observations.
For PPO, we use 461 nodes in the first layer and 512 in the second. The first layer size is specifically chosen to match the convolutional layer's parameter count, enabling a fair comparison between the two input variants. Separate linear actor and critic heads are applied to the output.

\subsection{Algorithm Details}
\subsubsection{PPO}

\begin{table}[H]
\centering
\caption{\textbf{PPO hyperparameters.} Identical for pixel-based and object-centric training runs}
\begin{tabular}{ll}
\toprule
\textbf{Hyperparameter} & \textbf{Value} \\
\midrule
Total timesteps         & $10^7$ \\
Parallel environments   & 8 \\
Rollout length          & 128 steps \\
Update epochs           & 4 \\
Minibatches             & 4 \\
Minibatch size          & $8 \times 128 / 4 = 256$ \\
Learning rate           & $2.5 \times 10^{-4}$ (linear decay) \\
Discount factor $\gamma$& 0.99 \\
GAE $\lambda$           & 0.95 \\
Clip $\varepsilon$      & 0.1 \\
Value function coeff.   & 0.5 \\
Entropy coeff.          & 0.01 \\
Max gradient norm       & 0.5 \\
Advantage normalization & Yes \\
Value loss clipping     & No \\
Optimizer               & Adam ($\varepsilon = 10^{-5}$) \\
\bottomrule
\end{tabular}
\label{tab:ppo_hyperparams}
\end{table}


\subsection{Training Protocol}
All \jaxatari agents reported in this paper were trained from scratch and without using pretrained weights. Furthermore, each agent is trained on environments initialized with three independent random seeds and evaluated on greedy ($\varepsilon=0$) policies. We report IQM and confidence intervals across seeds.

\subsection{Computational Resources}
We perform all our experiments on NVIDIA DGX hardware with A100 GPUs (40GB memory) and two AMD EPYC 7742 CPUs (128 cores, 256 threads).
Each experiments uses the maximum of one GPU per experiment.
The scaling experiment took a total of 32.5 GPU hours (including 5 ALE/vec-env runs for an average).
The learning curve computation (PPO) took 180h GPU hours for the main results and 20h for appendix results.
The evaluation on mods took a total of 2 GPU hours.
With additional non-reported development runs, we estimate the total number of GPU hours to be roughly 500.

\clearpage
\section{Extended Results}
\label{app:extended_results}

This section provides an extension to the evaluation of the JAXtari framework (\cf \cref{sec:evaluation}, specifically detailing the empirical performance across our benchmark suite. We first present an extension of the throughput analysis in \Cref{app:throughput} to quantify the computational advantages of our JAX-native implementation, followed by performance comparisons across 15 environments (\cf \cref{app:robustness}).

\subsection{Throughput and Scaling Analysis}
\label{app:throughput}

We extend our work by reporting three complementary views of scaling. First, we present \emph{per-game scaling curves}, quantifying how throughput, measured in environment steps per second, evolves as the number of parallel environments $N$ increases, and revealing the point at which each title becomes bandwidth- or compute-bound. Second, we report \emph{wall-clock time to $100 \times 10^6$ environment steps}, which serves as a practical, hardware-grounded measure of how quickly researchers can complete long-horizon training runs on a single accelerator. Third, we report the corresponding \emph{agent performance}, i.e., final and intermediate returns achieved within these extended training budgets, allowing the reader to relate raw throughput to learning outcomes rather than treating speed in isolation. Together, these three axes, scaling, wall-clock cost, and resulting performance, are intended to give a faithful picture of what \jaxatari enables in practice and to support reproducibility for follow-up work. Note that the reported ALE-VenctorEnv and EnvPool curves are the same over 5 games averaged throughput values for all runs.

\begin{figure}[ht]
    \centering
    \includegraphics[width=\linewidth]{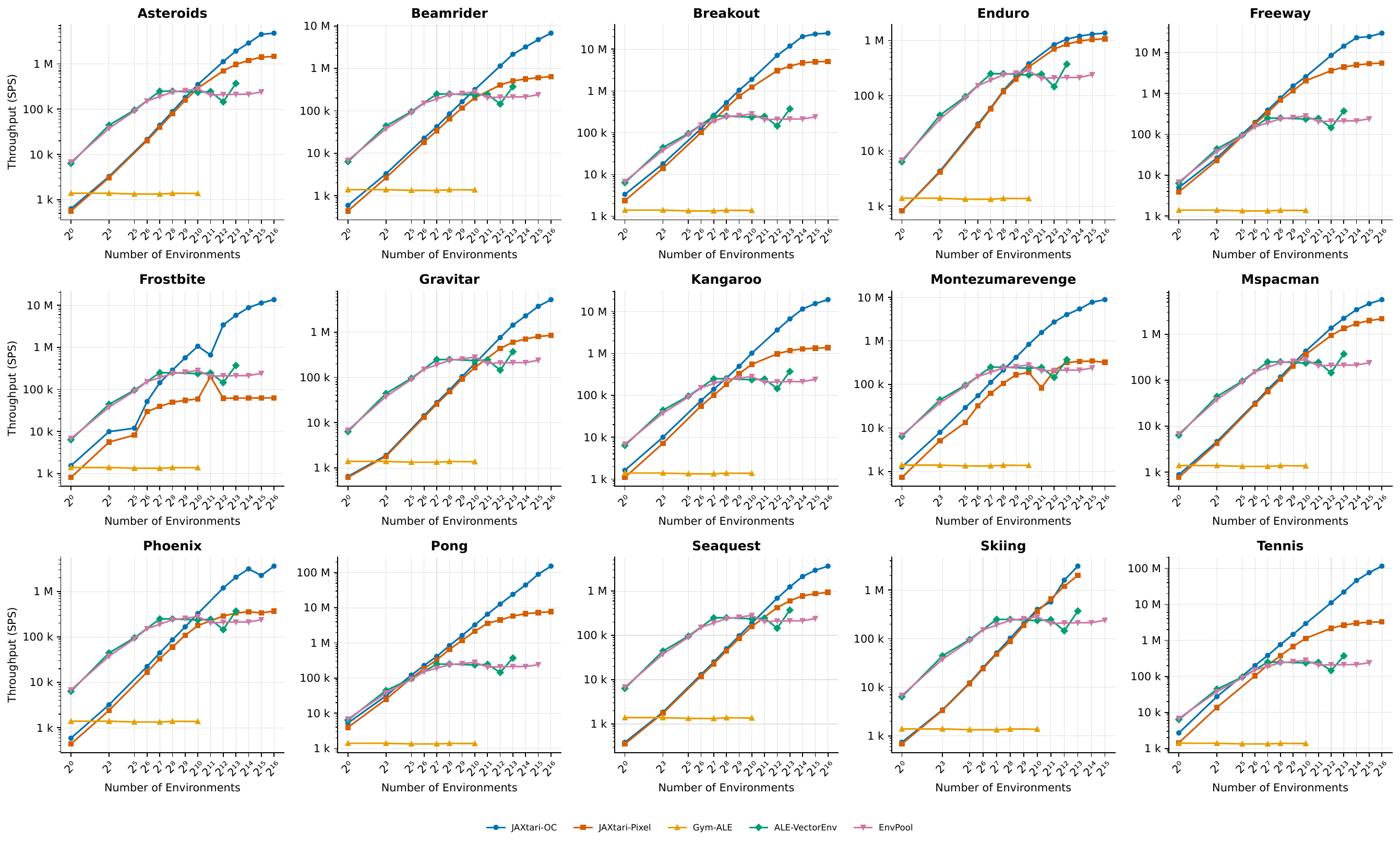}
    \caption{\textbf{\jaxatari increased the number of possible environments and consequently the steps per second.} While EnvPool and ALE-VectorEnv scale with the number of CPU threads (in our experiments, 256), our approach is not limited by it. Agents reported until reaching out-of-memory errors or 65K environments. ALE-VectorEnv and EnvPool curves are the same for all runs, averaged over 5 games.}
    \label{app:grid_throughput_scaling}
\end{figure}

\clearpage

\begin{figure}[ht]
    \centering
    \includegraphics[width=\linewidth]{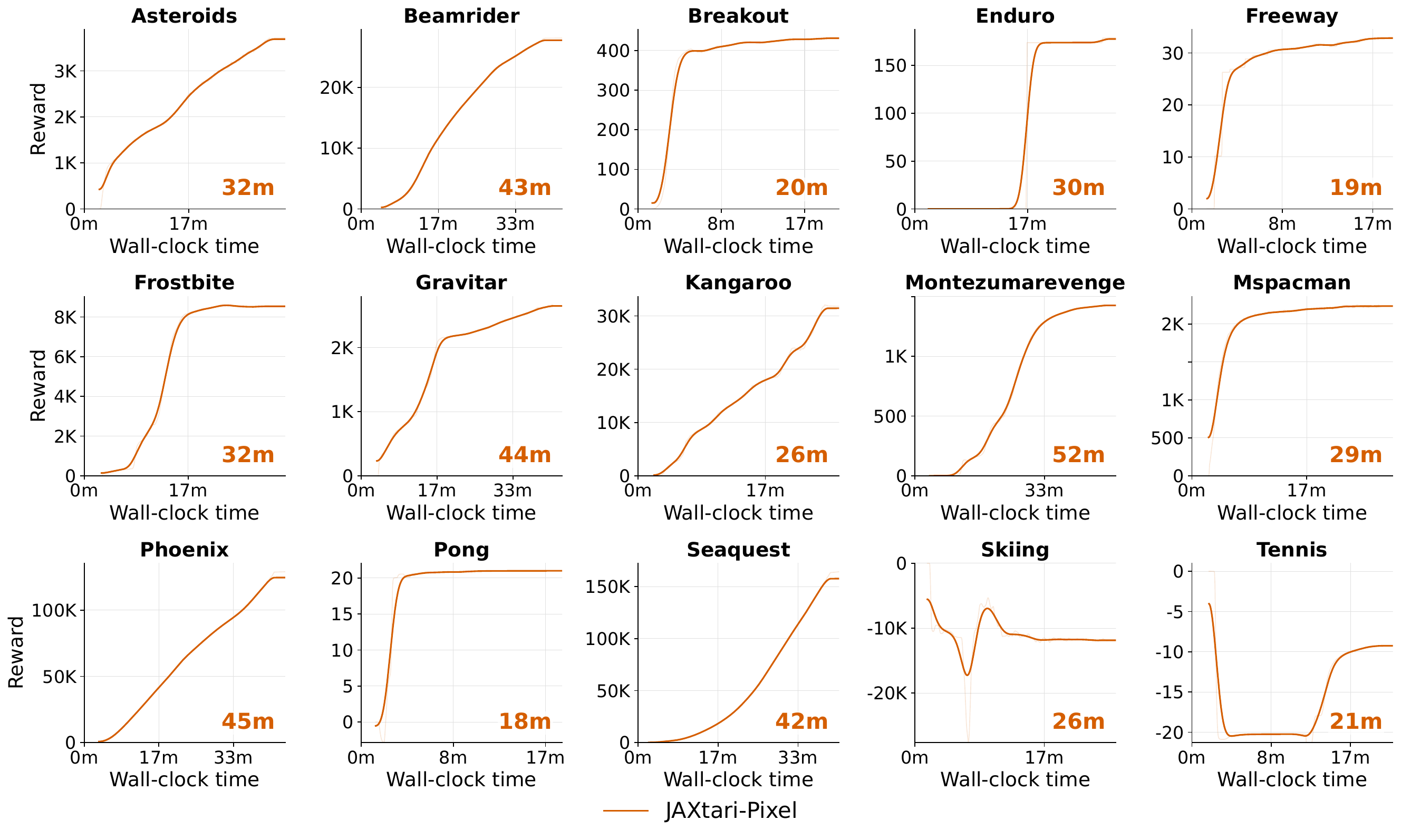}
    \caption{\textbf{On average, pixel-based approaches need 32 minutes to reach 100M steps on a single GPU.} Training curves (reward vs. wall-clock time) for PPO with pixel-based observations across 15 Atari games, using 2048 parallel environments. Orange labels indicate total training time per game, including JAX compilation time.}
    \label{fig:big_pixel}
\end{figure}

\begin{figure}[ht]
    \centering
    \includegraphics[width=\linewidth]{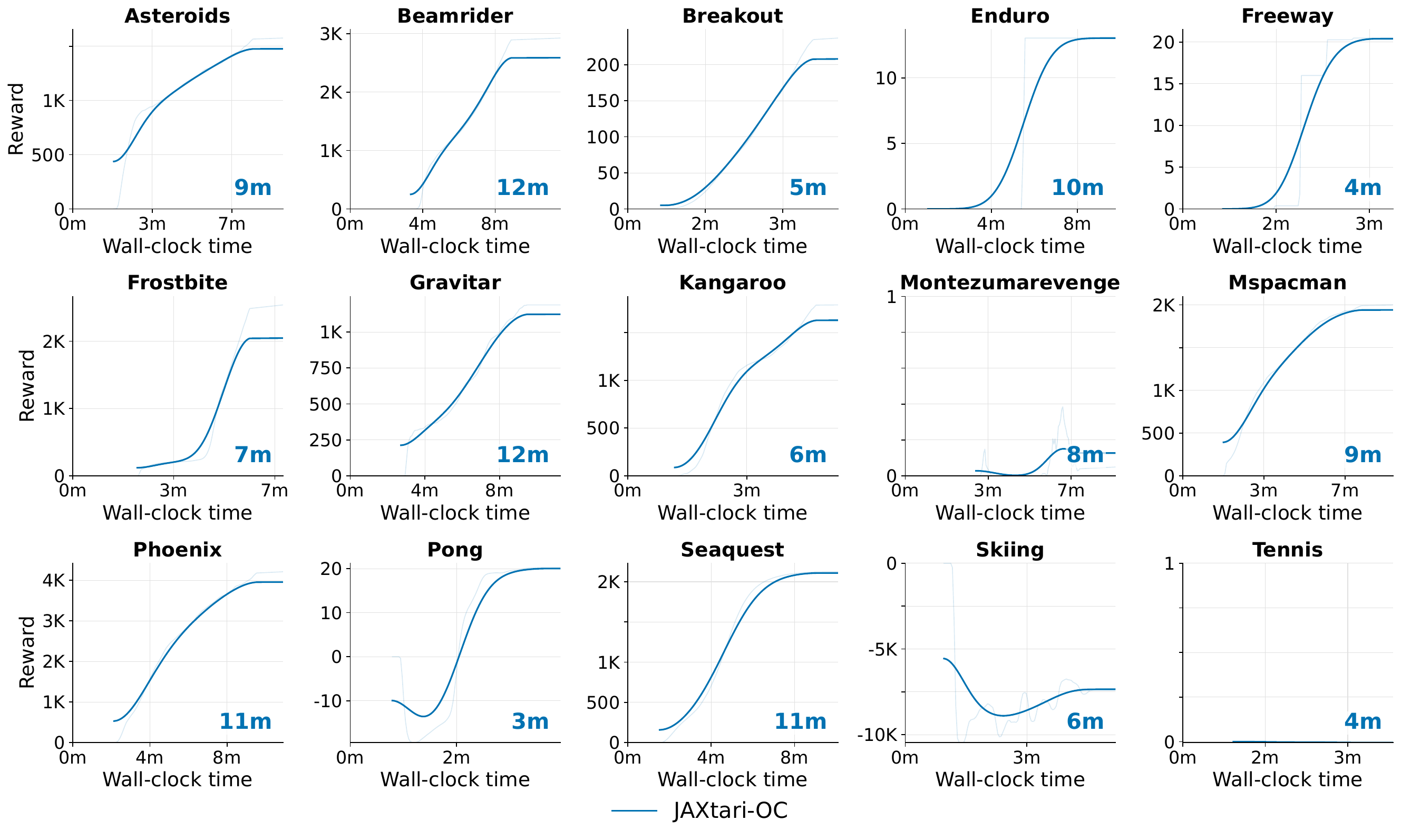}
    \caption{\textbf{On average, OC-based approaches need 8 minutes to reach 100M on a single GPU.} Training curves (reward vs. wall-clock time) for PPO with object-centric observations across 15 Atari games, using 8192 parallel environments. Blue labels indicate total training time per game, including JAX compilation time.}
    \label{fig:big_oc}
\end{figure}

\paragraph{Wall-Clock Throughput to 100M Steps.}
To probe the high-throughput regime of \jaxatari, we measure the wall-clock time required to reach $100 \times 10^6$ environment steps ($400 \times 10^6$ frames at frame-skip $4$), well beyond the $10$M--$50$M step horizon typically reported in the ALE. All runs execute on a single GPU and exploit JAX's \texttt{vmap} and JIT compilation, keeping the rollout fully on-device. To maximize GPU utilization, we set the number of parallel environments $N$ to $2048$ for pixel-based PPO and $8192$ for OC-based PPO, which is commonly the maximum for most games. All pixel-based agents reached $100M$ steps in under 1h, and OC even faster at a maximum of 12min (\cf \cref{fig:big_pixel} and \cref{fig:big_oc}).
For scaling up PPO, we follow the recommendation from \citet{beukman2026preventing} and simply increase the the number of minibatches to match the original size, as well as the total number of steps.

\clearpage
\paragraph{Does More Training Time Yield Better Performance?}

\begin{wrapfigure}[23]{r}{0.5\linewidth}
    \centering
    \includegraphics[width=\linewidth]{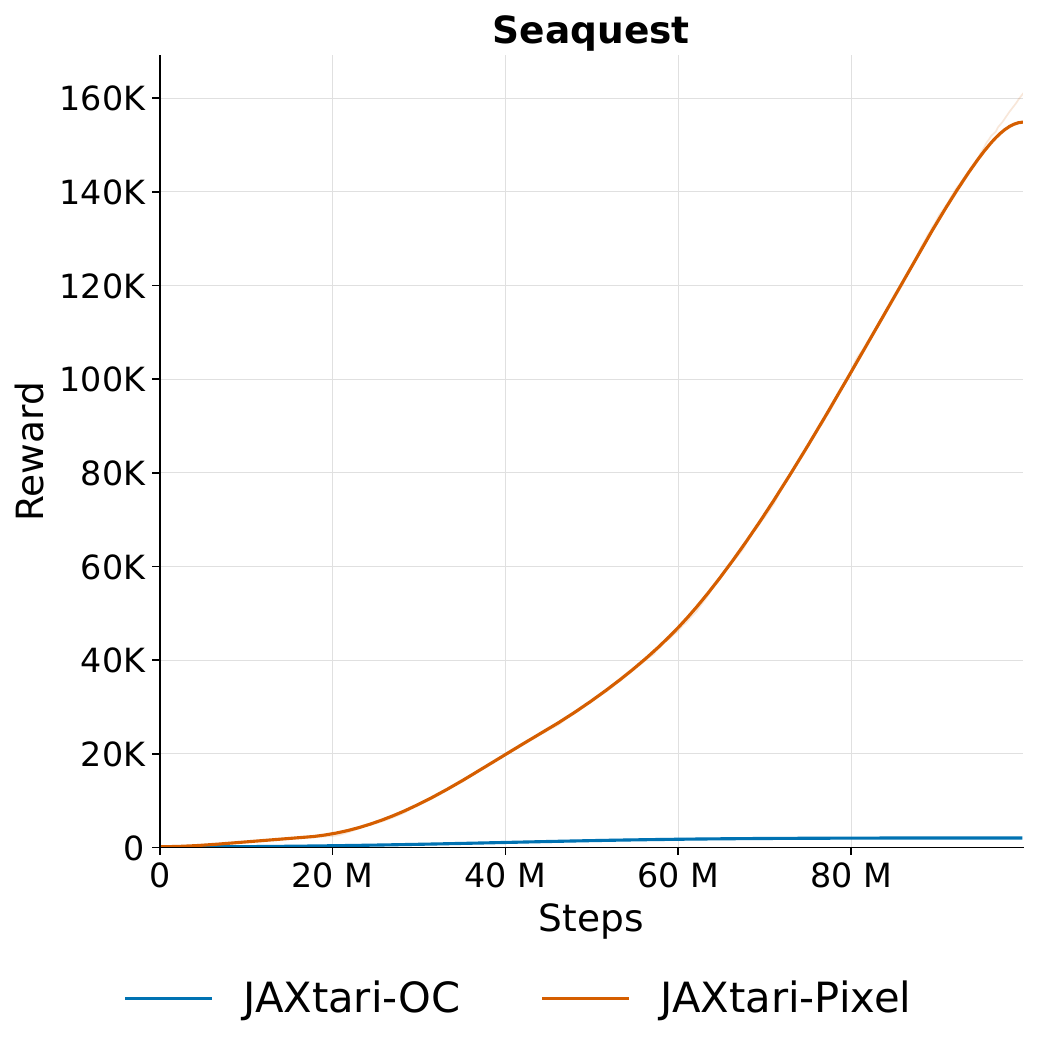}
    \caption{\textbf{Performance after 100M steps.} PPO reaches over 150k points in Seaquest before truncation by reaching 108000 frames.}
    \label{fig:seaquest100M}
\end{wrapfigure}

Whether extended training budgets translate into improved agent performance is not a question with a universal answer. On a substantial fraction of ALE titles, learning curves saturate well before the $100$M-step mark, and additional environment interactions yield diminishing, or even negligible, returns. However, this picture changes markedly for the subset of games that are notoriously difficult to solve or converge upon, most prominently \textit{Seaquest} and \textit{Montezuma's Revenge}. These environments are characterized by sparse, delayed, or hierarchically structured rewards, which require the agent to execute long sequences of coordinated actions before any learning signal is produced. In such regimes, a longer training horizon directly enlarges the agent's effective exploration budget: it increases the probability of stumbling upon rare reward-bearing trajectories, allows replay buffers to accumulate a more diverse set of transitions, and gives policy- and value-estimates additional time to escape local optima. Empirically, this manifests as higher final scores and qualitatively richer behaviors that simply do not emerge under shorter budgets. Our results in \cref{fig:seaquest100M} show that, with a training budget of 100M (or just 43min), we can achieve very high scores.

The throughput characterized in the preceding section opens a concrete avenue for future investigation. Because \jaxatari renders $100$M-step (and beyond) training runs tractable on a single GPU, it becomes feasible to systematically revisit the hard-exploration subset of the ALE under substantially extended compute budgets, $200$M, $500$M, or even $1$B steps, without resorting to large-scale distributed infrastructure. We see several promising directions enabled by this regime, which can be found in our future work section.

\subsection{Can JAXtari variations be used to evaluate robustness?}
\label{app:robustness}

\begin{figure}[t]
    \centering
    \includegraphics[width=\linewidth]{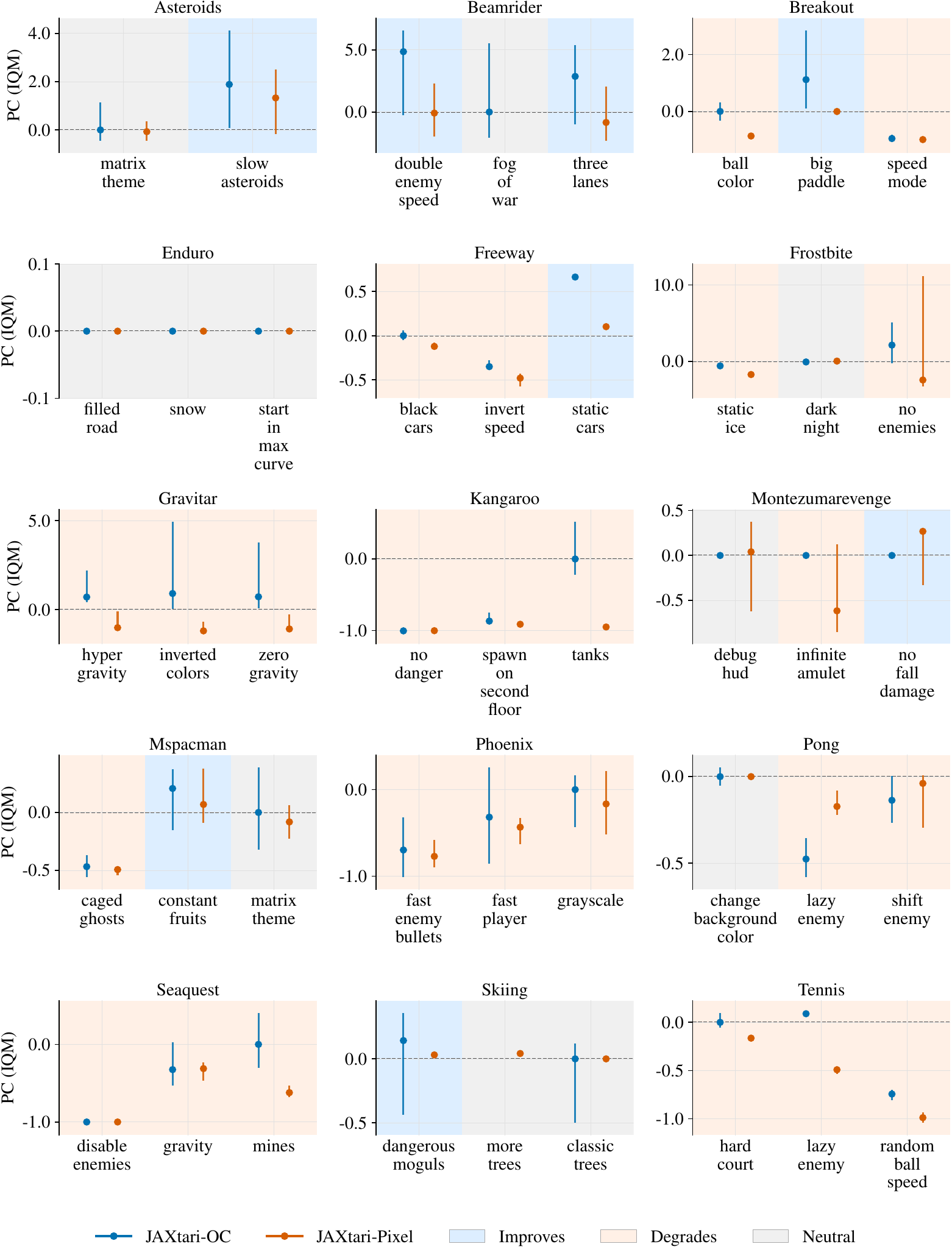}
    \caption{\textbf{\jaxatari allows users to evaluate performance changes when the environment changes, making it a valuable tool in researching shortcuts.} Per game evaluation measuring the drop or improvements of the agents. Evaluations were done over 10 runs per seed. Values show IQM with a 95\% CI. }
    \label{fig:performance_change_pergame}
\end{figure}

Regarding RQ3 we evaluated the performance of our PPO agents in variations of their trained environments. \cref{fig:performance_change} aggregates these over games and mods. \cref{fig:performance_change_pergame} shows the performance change per game and per mod. 

\clearpage

\section{\jaxatari Environment Descriptions and Modifications}
\label{appendix:environments}

\subsection{\jaxatari with open-source sprite-replacements}
\label{appendix:sprite_replacement}
\begin{figure}[htbp]
     \centering
     \begin{subfigure}[b]{0.18\textwidth}
         \centering
         \includegraphics[width=\textwidth]{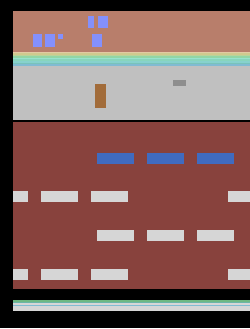}
         \caption{Frostbite}
     \end{subfigure}
     \hfill 
     \begin{subfigure}[b]{0.18\textwidth}
         \centering
         \includegraphics[width=\textwidth]{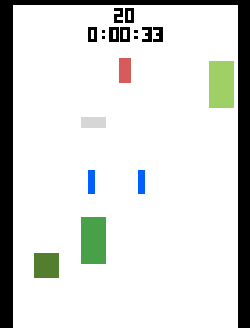}
         \caption{Skiing}
     \end{subfigure}
     \hfill
     \begin{subfigure}[b]{0.18\textwidth}
         \centering
         \includegraphics[width=\textwidth]{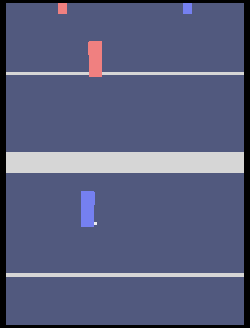}
         \caption{Tennis}
     \end{subfigure}
     \hfill
     \begin{subfigure}[b]{0.18\textwidth}
         \centering
         \includegraphics[width=\textwidth]{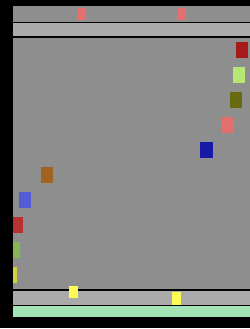}
         \caption{Freeway}
     \end{subfigure}
     \hfill
     \begin{subfigure}[b]{0.18\textwidth}
         \centering
         \includegraphics[width=\textwidth]{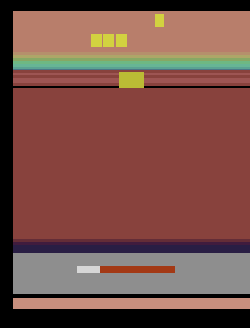}
         \caption{Seaquest}
     \end{subfigure}
     
     \caption{The replacement sprites only change the games visually but preserve game mechanics.}
    \label{fig:replace_grid}
\end{figure}

Users that do not own a license to the original Atari ROMs can instead utilize our open-source sprite replacement package, available for all games.
The replacement sprites are inspired by MinAtar~\citep{young2019minatar} and are single-color rectangles instead of the actual sprites (\cf~\Cref{fig:replace_grid}).

\subsection{Creating new Modifications}

JAXtari features a modular modification system that allows users to edit game semantics directly at the source-code level (as described in \cref{sec:method}). 

Modifications are built around the \texttt{JaxtariInternalModPlugin} and \texttt{JaxtariPostStepModPlugin}. When creating a new mod, you can intervene at two distinct points in the environment's lifecycle:

\begin{itemize}
    \item \textbf{Internal Modifications:} These interventions are applied \emph{before} Just-In-Time (JIT) compilation. Because they become part of the compiled transition function, they execute highly efficiently. Use this method to override core game constants, collision rules, environment methods, enemy speeds, or object spawning behaviors. You can also use the declarative asset system here to swap visual sprites by overriding values in the environment's original \texttt{ASSET\_CONFIG}.
    \item \textbf{Post-Step Modifications:} These interventions operate on the environment state \emph{after} the original transition function has evaluated. They are ideal for changes that do not require rewriting the core game dynamics, such as relocating an object, masking a state variable, changing a reward, or triggering an alternative episode termination.
\end{itemize}

Once your modification is defined, you can inject it directly during environment initialization by passing it to the \texttt{mods} argument: \texttt{env = jaxtari.make("game\_name", mods=["custom\_mod"])}.

\vspace{1em}
\noindent\rule{\textwidth}{0.4pt}
\vspace{1em}

\subsection*{Example: Modifying Kangaroo}

In the original Kangaroo environment, the player controls Mother Kangaroo on a mission to rescue her baby. This involves navigating obstacles, climbing ladders, collecting bonus fruit, and punching interfering monkeys. 

To evaluate whether an agent has learned robust causal mechanics (like climbing) rather than overfitting to punching the monkeys on the first floor, we can introduce a modification. Let's create a conceptual implementation of the \texttt{no\_monkey} mod of \textit{HackAtari}, which removes the monkeys from the game. 

Depending on your design, you could implement this as either an internal (\cf \cref{fig:internal-NoMonkey})  or post-step modification (\cf \cref{fig:post_step-NoMonkey}). This mod intercepts the state after the transition function evaluates and simply forces the monkeys' active status to be disabled.

\begin{figure}[t]
    \centering
    \begin{codebox}\input{code_listings/listing4.pygtex}\end{codebox}
    %
    %
    %
    %
    %
    \caption{Implementation of an \texttt{InternalModPlugin} in JAXtari. This example demonstrates a structural intervention where the environment's internal \texttt{\_monkey\_controller} is patched to a no-op, effectively removing a specific game mechanic while maintaining JIT-compatibility and end-to-end execution.}
    \label{fig:internal-NoMonkey}
\end{figure}

\begin{figure}[t]
    \centering
    \begin{codebox}\input{code_listings/listing5.pygtex}\end{codebox}
    %
    %
    %
    %
    %
    \caption{State-level modification using the \texttt{PostStepModPlugin}. Unlike internal logic patching, this plugin performs post-hoc state manipulation after each transition, allowing researchers to pin specific environmental variables or perform causal interventions on the observation and state space. This however have to be done in every step.}
    \label{fig:post_step-NoMonkey}
\end{figure}

Further, we can use the mod system to swap assets or constants within the game implementation. Lets say the agent learned to dodge coconuts. Now we swap the coconuts with a slightly larger fireball and see what is happening (\cf \cref{fig:fireball-mod}).

\begin{figure}
    \centering
    \begin{codebox}\input{code_listings/listing6.pygtex}\end{codebox}
    \caption{Example of an asset and constant injection for robustness evaluation. By overriding sprite assets and physical constants (e.g., object dimensions or speed), \jaxatari enables the rapid generation of Out-of-Distribution (OOD) scenarios to test the generalization capabilities of RL agents.}
    \label{fig:fireball-mod}
\end{figure}

While both \jaxatari and HackAtari allow for environment modifications, they differ fundamentally in their implementation and performance goals. HackAtari is built on top of the standard ALE and Stella emulators. In HackAtari, modifications involve intercepting the emulator's RAM. In \jaxatari, modifications are applied directly to the high-level JAX state objects or by overriding Python methods before JIT compilation.

By keeping the environment functional and stateless, these modifications remain compatible with JAX transformations like \texttt{jax.vmap} and \texttt{jax.jit}, allowing you to run thousands of modified environment instances simultaneously on a single GPU.

\clearpage
\subsection{List of Games and Modifications}
\label{appendix:game_mosaic}
\begin{figure}[H]
  \centering
  \resizebox{\textwidth}{!}{%
    \setlength{\tabcolsep}{3pt}%
    \renewcommand{\arraystretch}{1.1}%
    \begin{tabular}{p{3.2cm} p{3.2cm} p{3.2cm} p{3.2cm} p{3.2cm}}

      \begin{minipage}[t]{\linewidth}\centering\includegraphics[width=\linewidth]{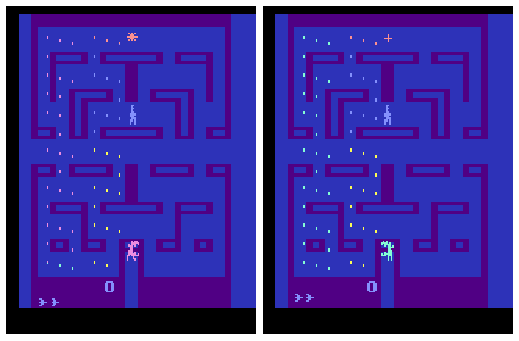}\\[-3pt]{\scriptsize\textit{Alien}}\end{minipage} &
      \begin{minipage}[t]{\linewidth}\centering\includegraphics[width=\linewidth]{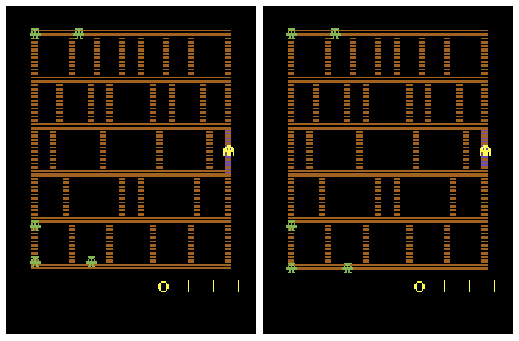}\\[-3pt]{\scriptsize\textit{Amidar}}\end{minipage} &
      \begin{minipage}[t]{\linewidth}\centering\includegraphics[width=\linewidth]{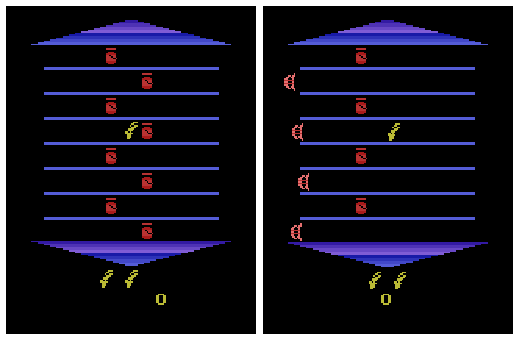}\\[-3pt]{\scriptsize\textit{Asterix}}\end{minipage} &
      \begin{minipage}[t]{\linewidth}\centering\includegraphics[width=\linewidth]{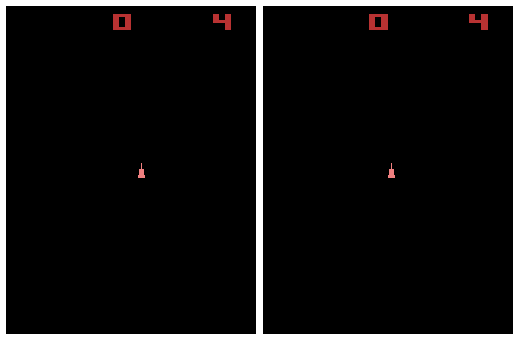}\\[-3pt]{\scriptsize\textit{Asteroids}}\end{minipage} &
      \begin{minipage}[t]{\linewidth}\centering\includegraphics[width=\linewidth]{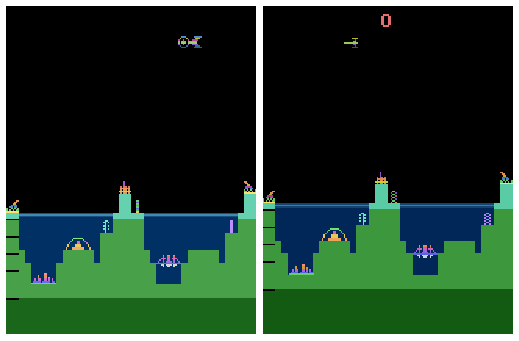}\\[-3pt]{\scriptsize\textit{Atlantis}}\end{minipage} \\

      \begin{minipage}[t]{\linewidth}\centering\includegraphics[width=\linewidth]{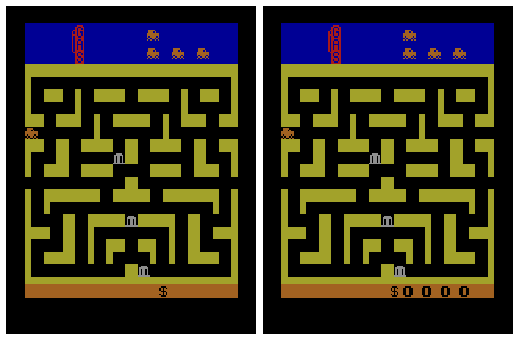}\\[-3pt]{\scriptsize\textit{Bank Heist}}\end{minipage} &
      \begin{minipage}[t]{\linewidth}\centering\includegraphics[width=\linewidth]{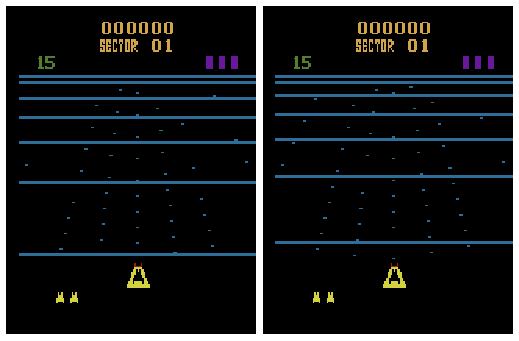}\\[-3pt]{\scriptsize\textit{Beam Rider}}\end{minipage} &
      \begin{minipage}[t]{\linewidth}\centering\includegraphics[width=\linewidth]{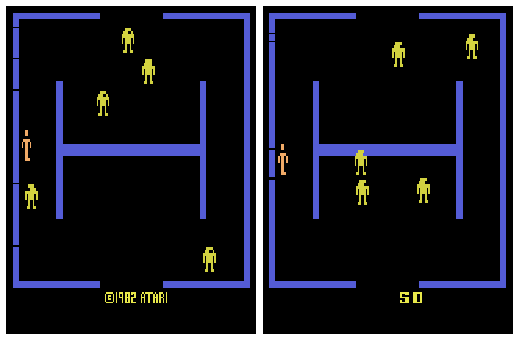}\\[-3pt]{\scriptsize\textit{Berzerk}}\end{minipage} &
      \begin{minipage}[t]{\linewidth}\centering\includegraphics[width=\linewidth]{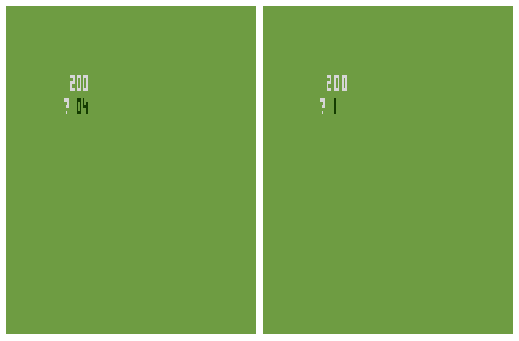}\\[-3pt]{\scriptsize\textit{Blackjack}}\end{minipage} &
      \begin{minipage}[t]{\linewidth}\centering\includegraphics[width=\linewidth]{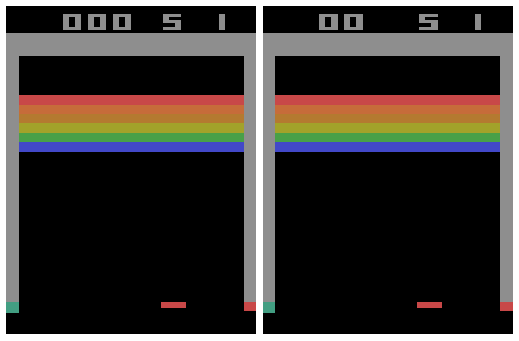}\\[-3pt]{\scriptsize\textit{Breakout}}\end{minipage} \\

      \begin{minipage}[t]{\linewidth}\centering\includegraphics[width=\linewidth]{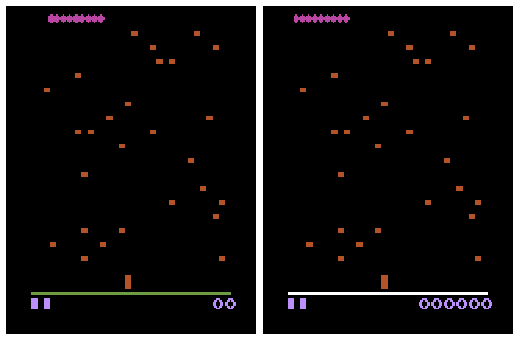}\\[-3pt]{\scriptsize\textit{Centipede}}\end{minipage} &
      \begin{minipage}[t]{\linewidth}\centering\includegraphics[width=\linewidth]{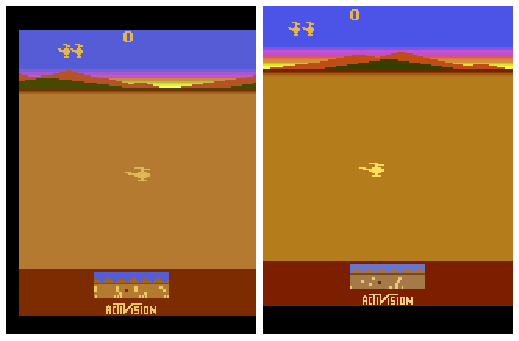}\\[-3pt]{\scriptsize\textit{Chopper Cmd.}}\end{minipage} &
      \begin{minipage}[t]{\linewidth}\centering\includegraphics[width=\linewidth]{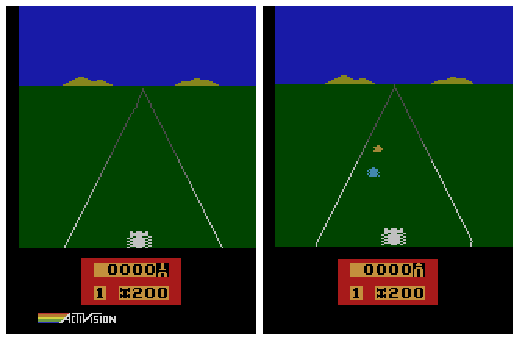}\\[-3pt]{\scriptsize\textit{Enduro}}\end{minipage} &
      \begin{minipage}[t]{\linewidth}\centering\includegraphics[width=\linewidth]{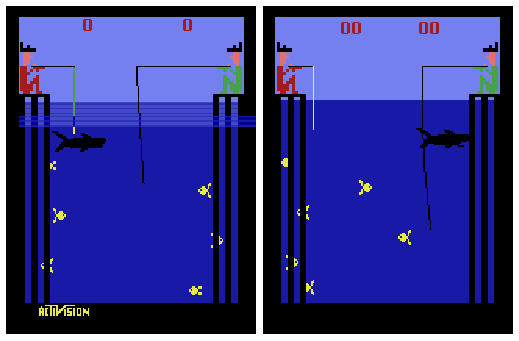}\\[-3pt]{\scriptsize\textit{Fishing Derby}}\end{minipage} &
      \begin{minipage}[t]{\linewidth}\centering\includegraphics[width=\linewidth]{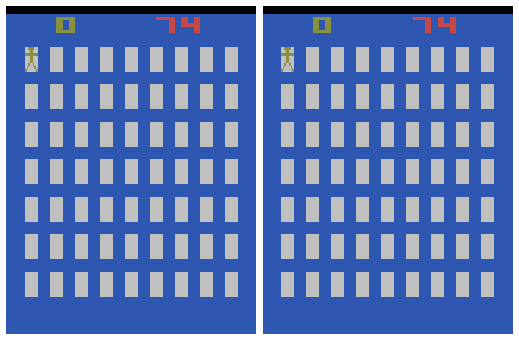}\\[-3pt]{\scriptsize\textit{Flag Capture}}\end{minipage} \\

      \begin{minipage}[t]{\linewidth}\centering\includegraphics[width=\linewidth]{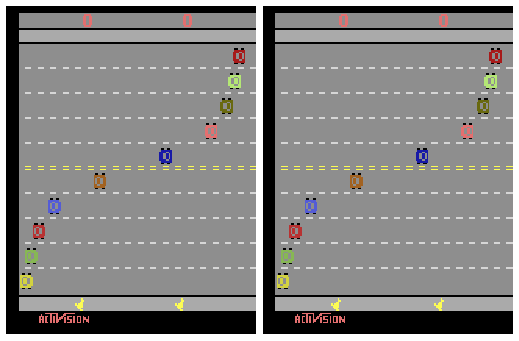}\\[-3pt]{\scriptsize\textit{Freeway}}\end{minipage} &
      \begin{minipage}[t]{\linewidth}\centering\includegraphics[width=\linewidth]{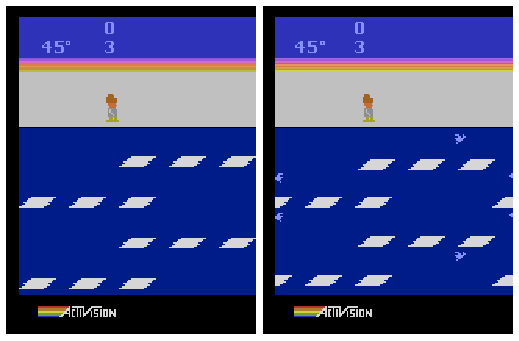}\\[-3pt]{\scriptsize\textit{Frostbite}}\end{minipage} &
      \begin{minipage}[t]{\linewidth}\centering\includegraphics[width=\linewidth]{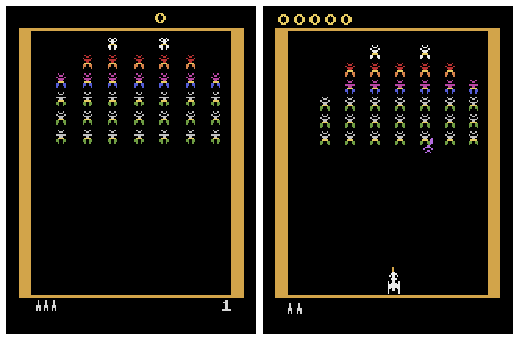}\\[-3pt]{\scriptsize\textit{Galaxian}}\end{minipage} &
      \begin{minipage}[t]{\linewidth}\centering\includegraphics[width=\linewidth]{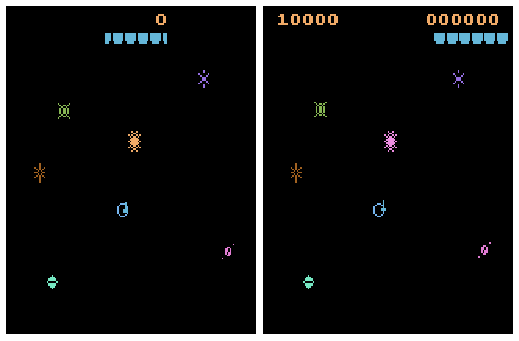}\\[-3pt]{\scriptsize\textit{Gravitar}}\end{minipage} &
      \begin{minipage}[t]{\linewidth}\centering\includegraphics[width=\linewidth]{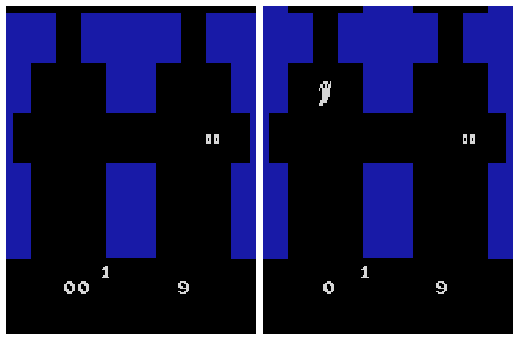}\\[-3pt]{\scriptsize\textit{Haunted House}}\end{minipage} \\

      \begin{minipage}[t]{\linewidth}\centering\includegraphics[width=\linewidth]{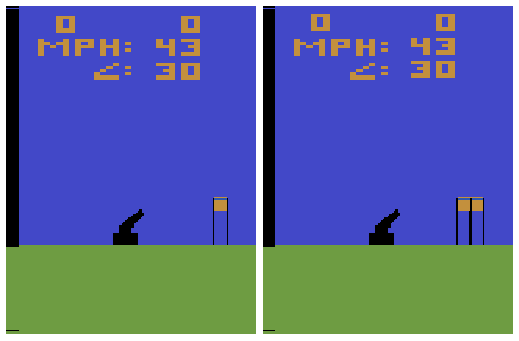}\\[-3pt]{\scriptsize\textit{H.\ Cannonball}}\end{minipage} &
      \begin{minipage}[t]{\linewidth}\centering\includegraphics[width=\linewidth]{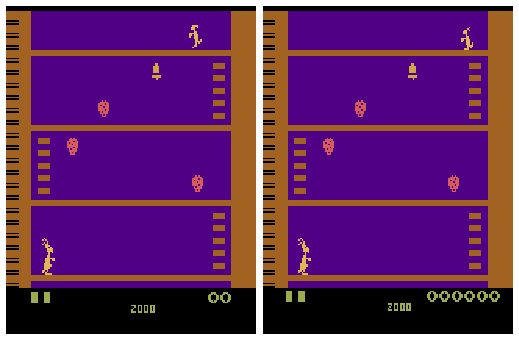}\\[-3pt]{\scriptsize\textit{Kangaroo}}\end{minipage} &
      \begin{minipage}[t]{\linewidth}\centering\includegraphics[width=\linewidth]{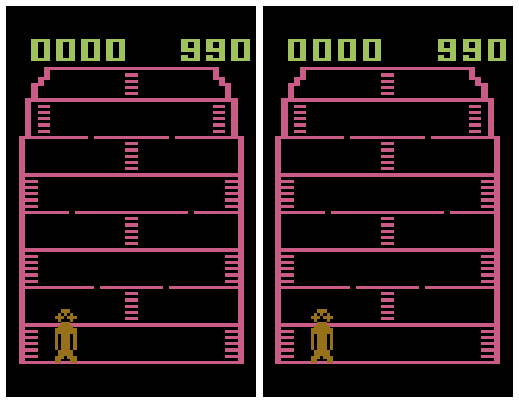}\\[-3pt]{\scriptsize\textit{King Kong}}\end{minipage} &
      \begin{minipage}[t]{\linewidth}\centering\includegraphics[width=\linewidth]{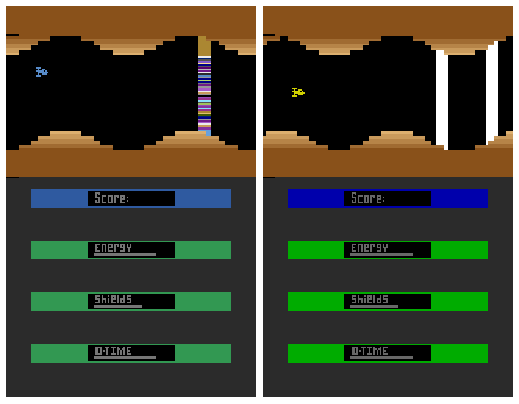}\\[-3pt]{\scriptsize\textit{Laser Gates}}\end{minipage} &
      \begin{minipage}[t]{\linewidth}\centering\includegraphics[width=\linewidth]{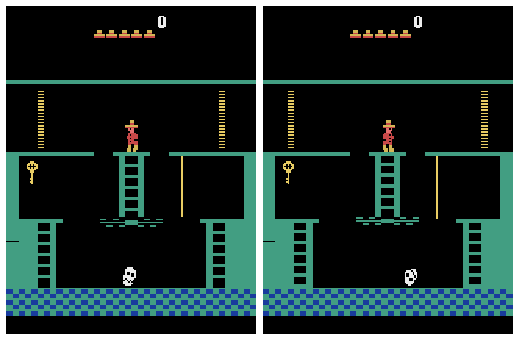}\\[-3pt]{\scriptsize\textit{Montezuma}}\end{minipage} \\

      \begin{minipage}[t]{\linewidth}\centering\includegraphics[width=\linewidth]{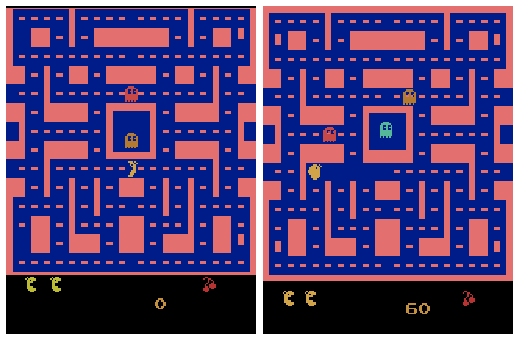}\\[-3pt]{\scriptsize\textit{Ms.\ Pac-Man}}\end{minipage} &
      \begin{minipage}[t]{\linewidth}\centering\includegraphics[width=\linewidth]{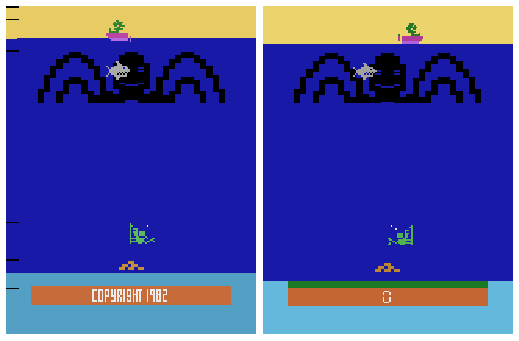}\\[-3pt]{\scriptsize\textit{Name This Game}}\end{minipage} &
      \begin{minipage}[t]{\linewidth}\centering\includegraphics[width=\linewidth]{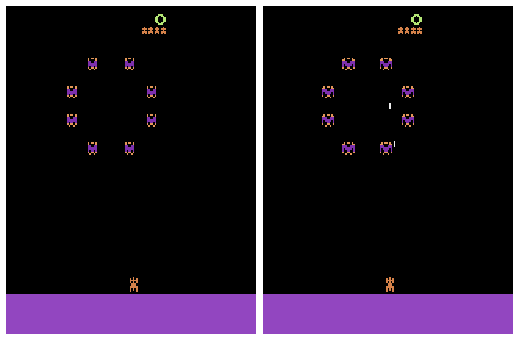}\\[-3pt]{\scriptsize\textit{Phoenix}}\end{minipage} &
      \begin{minipage}[t]{\linewidth}\centering\includegraphics[width=\linewidth]{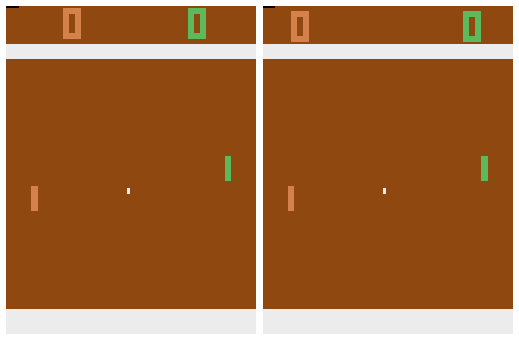}\\[-3pt]{\scriptsize\textit{Pong}}\end{minipage} &
      \begin{minipage}[t]{\linewidth}\centering\includegraphics[width=\linewidth]{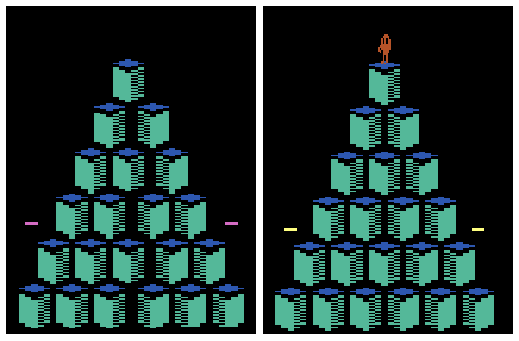}\\[-3pt]{\scriptsize\textit{Q*bert}}\end{minipage} \\

      \begin{minipage}[t]{\linewidth}\centering\includegraphics[width=\linewidth]{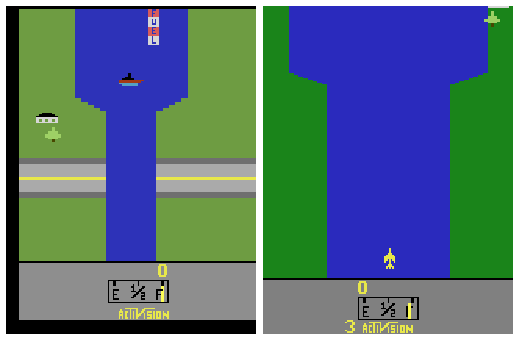}\\[-3pt]{\scriptsize\textit{River Raid}}\end{minipage} &
      \begin{minipage}[t]{\linewidth}\centering\includegraphics[width=\linewidth]{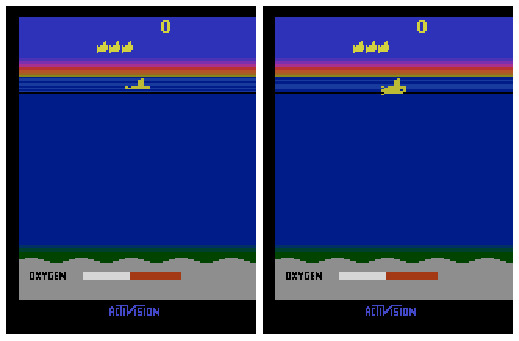}\\[-3pt]{\scriptsize\textit{Seaquest}}\end{minipage} &
      \begin{minipage}[t]{\linewidth}\centering\includegraphics[width=\linewidth]{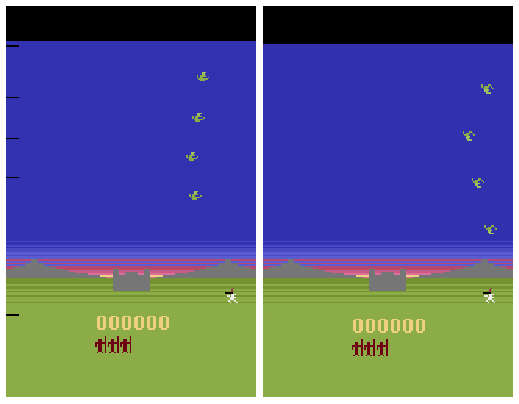}\\[-3pt]{\scriptsize\textit{Sir Lancelot}}\end{minipage} &
      \begin{minipage}[t]{\linewidth}\centering\includegraphics[width=\linewidth]{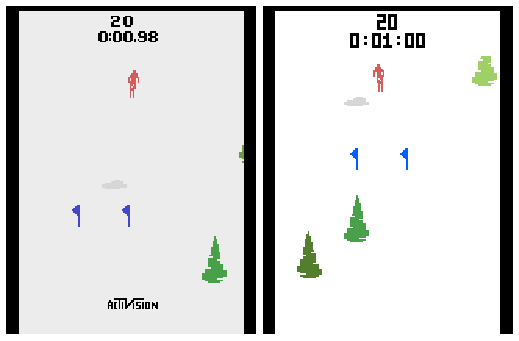}\\[-3pt]{\scriptsize\textit{Skiing}}\end{minipage} &
      \begin{minipage}[t]{\linewidth}\centering\includegraphics[width=\linewidth]{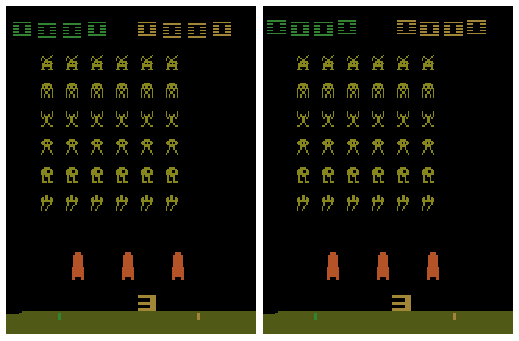}\\[-3pt]{\scriptsize\textit{Space Invaders}}\end{minipage} \\

      \begin{minipage}[t]{\linewidth}\centering\includegraphics[width=\linewidth]{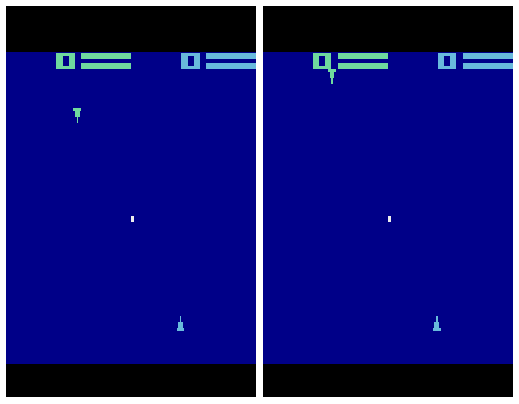}\\[-3pt]{\scriptsize\textit{Spacewar}}\end{minipage} &
      \begin{minipage}[t]{\linewidth}\centering\includegraphics[width=\linewidth]{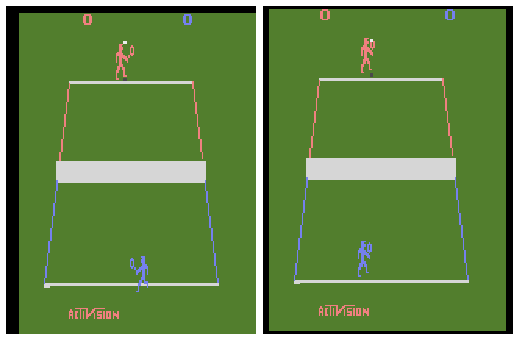}\\[-3pt]{\scriptsize\textit{Tennis}}\end{minipage} &
      \begin{minipage}[t]{\linewidth}\centering\includegraphics[width=\linewidth]{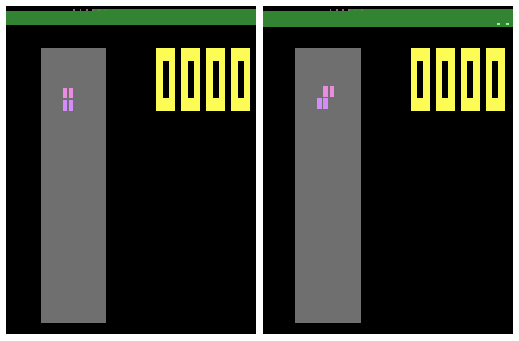}\\[-3pt]{\scriptsize\textit{Tetris}}\end{minipage} &
      \begin{minipage}[t]{\linewidth}\centering\includegraphics[width=\linewidth]{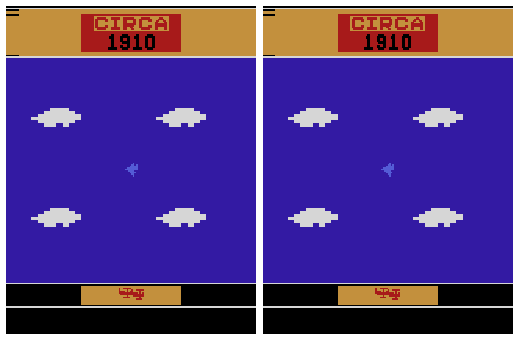}\\[-3pt]{\scriptsize\textit{Time Pilot}}\end{minipage} &
      \begin{minipage}[t]{\linewidth}\centering\includegraphics[width=\linewidth]{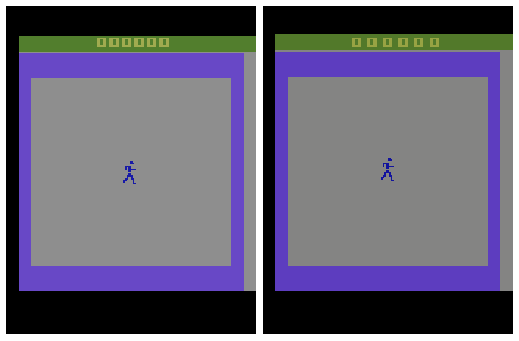}\\[-3pt]{\scriptsize\textit{Tron}}\end{minipage} \\

      \begin{minipage}[t]{\linewidth}\centering\includegraphics[width=\linewidth]{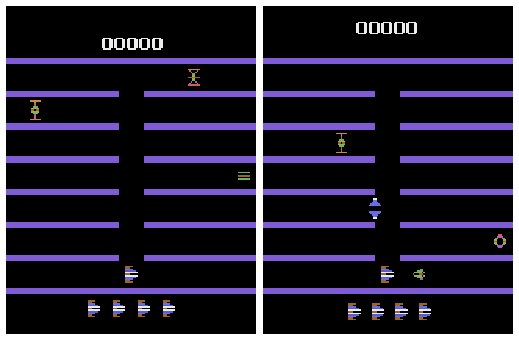}\\[-3pt]{\scriptsize\textit{Turmoil}}\end{minipage} &
      \begin{minipage}[t]{\linewidth}\centering\includegraphics[width=\linewidth]{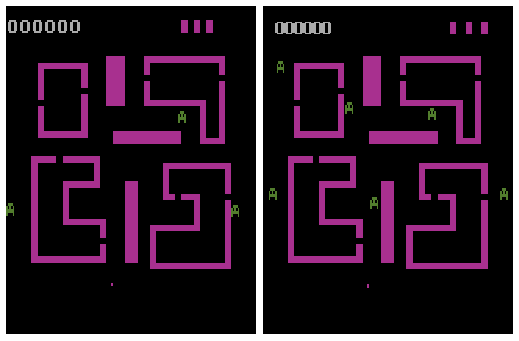}\\[-3pt]{\scriptsize\textit{Venture}}\end{minipage} &
      \begin{minipage}[t]{\linewidth}\centering\includegraphics[width=\linewidth]{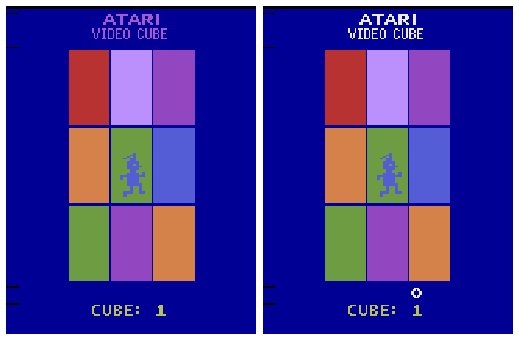}\\[-3pt]{\scriptsize\textit{Video Cube}}\end{minipage} &
      \begin{minipage}[t]{\linewidth}\centering\includegraphics[width=\linewidth]{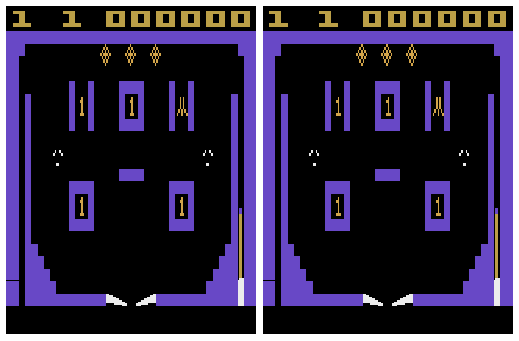}\\[-3pt]{\scriptsize\textit{Video Pinball}}\end{minipage} &
      \begin{minipage}[t]{\linewidth}\centering\includegraphics[width=\linewidth]{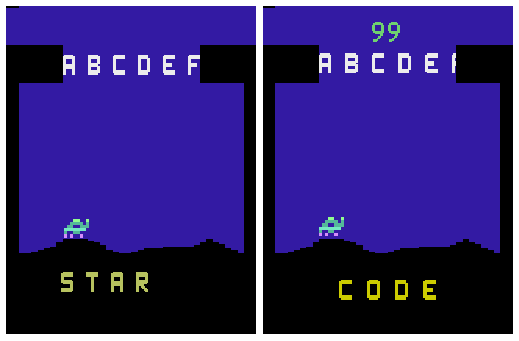}\\[-3pt]{\scriptsize\textit{Word Zapper}}\end{minipage} \\

    \end{tabular}%
  }
  \caption{%
    ALE (left) vs. JAXtari (right) renderings for all 45 supported games.
  }
  \label{fig:env_grid_all}
\end{figure}

\noindent Figure~\ref{fig:env_grid_all} provides a visual overview of all 45 environments currently supported by \jaxatari. Full environment descriptions and available modifications follow below.

\subsection{Alien}
\subsubsection{Alien \ details}

\vspace{0.5cm}

\noindent
\begin{minipage}[c]{0.52\textwidth}
\raggedright
The player is trapped aboard a maze-like spaceship alongside three hostile aliens, whose eggs are scattered throughout the vessel and must be destroyed. The aliens actively pursue and attack the player, but can be temporarily repelled using a flamethrower. Occasionally a pulsar power-up can be collected, granting the ability to kill aliens outright for a limited time.
\end{minipage}%
\hfill%
\begin{minipage}[c]{0.38\textwidth}
\centering
\setlength{\fboxsep}{0pt}%
\setlength{\fboxrule}{0.4pt}%
\fcolorbox{gray}{white}{%
  \includegraphics[width=\linewidth,height=6.0cm,keepaspectratio]{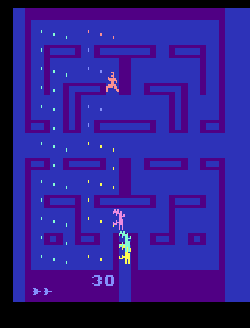}%
}
\end{minipage}

\subsubsection{Available Modifications for Alien}
\begin{table}[h]
    \centering
    \begin{tabular}{p{3cm} | p{10cm}} \toprule
        \textbf{Modification} & \textbf{Effect} \\ \midrule
       aggressive\_swarm & Aliens spend almost all their time actively chasing the player. \\
       dont\_kill & Punishes for killing and shooting enemies. \\
       end\_game & Most eggs are already eaten. \\
       extra\_lives & Start with extra lives. \\
       last\_egg & Only one egg is remaining. \\
       matrix\_theme & A Matrix-themed mod: black background, green walls/enemies/eggs. \\
       pacifist\_mode & Aliens are never frightened, forcing a pure evasion playstyle. \\
       short\_circuit & Vulnerability window is drastically reduced. \\
        \bottomrule
    \end{tabular}
\end{table}

\subsubsection{Selection of Available Visual Modifications}
\begin{figure}[H]
  \centering
  \setlength{\fboxsep}{0pt}%
  \setlength{\fboxrule}{0.4pt}%
  \fcolorbox{gray}{white}{%
    \includegraphics[height=6cm,keepaspectratio]{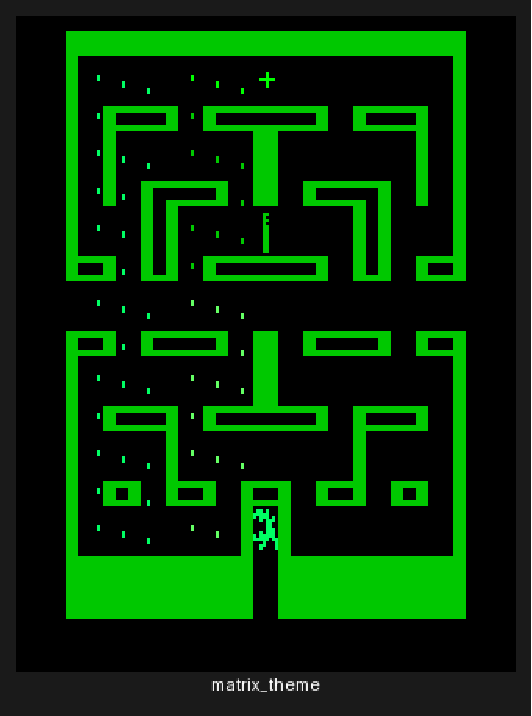}%
  }
  \caption{%
    A selection of visual modifications available for Alien,
    with each frame showing a different mod applied to the base environment.%
  }
  \label{fig:mods-Alien}
\end{figure}

\newpage

\subsection{Amidar}
\subsubsection{Amidar \ details}

\vspace{0.5cm}

\noindent
\begin{minipage}[c]{0.52\textwidth}
\raggedright
The player traverses a 2-dimensional grid, aiming to visit every cell while evading pursuing enemies. At certain moments the dynamic reverses: enemies transform into chickens that the player can actively chase and catch.
\end{minipage}%
\hfill%
\begin{minipage}[c]{0.38\textwidth}
\centering
\setlength{\fboxsep}{0pt}%
\setlength{\fboxrule}{0.4pt}%
\fcolorbox{gray}{white}{%
  \includegraphics[width=\linewidth,height=6.0cm,keepaspectratio]{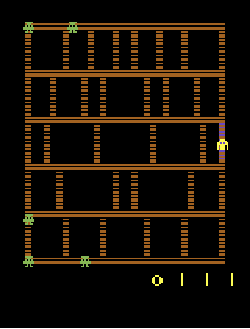}%
}
\end{minipage}

\subsubsection{Available Modifications for Amidar}
\begin{table}[h]
    \centering
    \begin{tabular}{p{3cm} | p{10cm}} \toprule
        \textbf{Modification} & \textbf{Effect} \\ \midrule
       \multicolumn{2}{l}{\textit{No modifications implemented yet.}} \\
        \bottomrule
    \end{tabular}
\end{table}

\newpage

\subsection{Asterix}
\subsubsection{Asterix \ details}

\vspace{0.5cm}

\noindent
\begin{minipage}[c]{0.52\textwidth}
\raggedright
The player controls Asterix, who moves continuously along the horizontal axis and discretely along the vertical. A stream of objects scrolls across the screen, including harmful lyres and various collectibles. The goal is to collect as many useful objects as possible while avoiding lyres, which cost a life on contact. The game starts with three lives, and bonus points are awarded upon reaching certain score thresholds.
\end{minipage}%
\hfill%
\begin{minipage}[c]{0.38\textwidth}
\centering
\setlength{\fboxsep}{0pt}%
\setlength{\fboxrule}{0.4pt}%
\fcolorbox{gray}{white}{%
  \includegraphics[width=\linewidth,height=6.0cm,keepaspectratio]{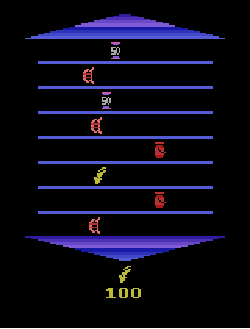}%
}
\end{minipage}

\subsubsection{Available Modifications for Asterix}
\begin{table}[h]
    \centering
    \begin{tabular}{p{3cm} | p{10cm}} \toprule
        \textbf{Modification} & \textbf{Effect} \\ \midrule
       \multicolumn{2}{l}{\textit{No modifications implemented yet.}} \\
        \bottomrule
    \end{tabular}
\end{table}

\newpage

\subsection{Asteroids}
\subsubsection{Asteroids \ details}

\vspace{0.5cm}

\noindent
\begin{minipage}[c]{0.52\textwidth}
\raggedright
The player pilots a spaceship through an asteroid field, destroying asteroids by shooting them. Large asteroids break into smaller fragments upon being hit, all of which must be cleared to advance to the next level. Flying saucers occasionally appear as an additional threat.
\end{minipage}%
\hfill%
\begin{minipage}[c]{0.38\textwidth}
\centering
\setlength{\fboxsep}{0pt}%
\setlength{\fboxrule}{0.4pt}%
\fcolorbox{gray}{white}{%
  \includegraphics[width=\linewidth,height=6.0cm,keepaspectratio]{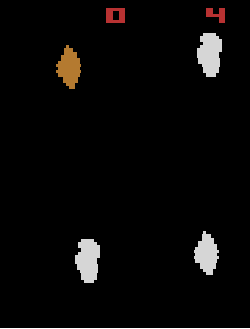}%
}
\end{minipage}

\subsubsection{Available Modifications for Asteroids}
\begin{table}[h]
    \centering
    \begin{tabular}{p{3cm} | p{10cm}} \toprule
        \textbf{Modification} & \textbf{Effect} \\ \midrule
       dont\_shoot & Internal mod that punishes shooting \\
       matrix\_theme & A Matrix-themed mod: black background, green asteroids \\
       slow\_asteroids & The asteroids are slower \\
       instant\_turn & Player instantly turns in the pressed direction \\
        \bottomrule
    \end{tabular}
\end{table}

\subsubsection{Selection of Available Visual Modifications}
\begin{figure}[H]
  \centering
  \setlength{\fboxsep}{0pt}%
  \setlength{\fboxrule}{0.4pt}%
  \fcolorbox{gray}{white}{%
    \includegraphics[height=6cm,keepaspectratio]{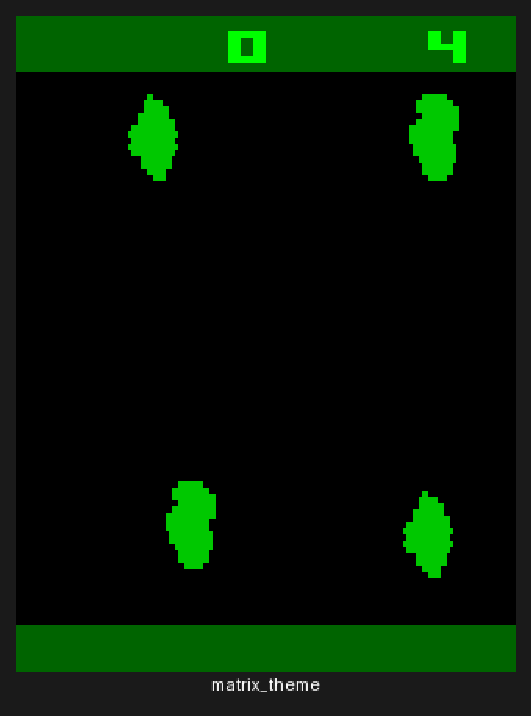}%
  }
  \caption{%
    A selection of visual modifications available for Asteroids,
    with each frame showing a different mod applied to the base environment.%
  }
  \label{fig:mods-Asteroids}
\end{figure}

\newpage

\subsection{Atlantis}
\subsubsection{Atlantis \ details}

\vspace{0.5cm}

\noindent
\begin{minipage}[c]{0.52\textwidth}
\raggedright
The player defends the submerged city of Atlantis against waves of enemies that slowly descend toward it. Three defense posts are at the player's disposal to shoot down attackers before they reach striking distance. The game is lost if all seven of Atlantis' installations are destroyed. After successfully repelling a wave and accumulating enough points, damaged installations can be rebuilt.
\end{minipage}%
\hfill%
\begin{minipage}[c]{0.38\textwidth}
\centering
\setlength{\fboxsep}{0pt}%
\setlength{\fboxrule}{0.4pt}%
\fcolorbox{gray}{white}{%
  \includegraphics[width=\linewidth,height=6.0cm,keepaspectratio]{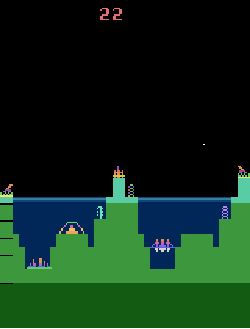}%
}
\end{minipage}

\subsubsection{Available Modifications for Atlantis}
\begin{table}[h]
    \centering
    \begin{tabular}{p{3cm} | p{10cm}} \toprule
        \textbf{Modification} & \textbf{Effect} \\ \midrule
        jets\_only & Overrides enemy spawn probabilities to only spawn Jet-type enmies. \\
        no\_last\_line & Disable the lowest lane so no enemies can spawn there. \\
        random\_enemies & Overrides enemy spawn probabilities to make spawning random. \\
        speed\_mode\_fast & Increases overall speed of the game to 6. \\
        speed\_mode\_medium & Increases overall speed of the game to 4. \\
        speed\_mode\_slow & Increases overall speed of the game to 2. \\
        \bottomrule
    \end{tabular}
\end{table}

\newpage

\subsection{Bankheist}
\subsubsection{Bankheist \ details}

\vspace{0.5cm}

\noindent
\begin{minipage}[c]{0.52\textwidth}
\raggedright
The player takes the role of a bank robber navigating a getaway car through maze-like cities. Banks can be robbed along the way, though each robbery triggers a police pursuit. Police cars can be neutralized by dropping dynamite, though driving over one's own explosives is equally fatal. Entering a new city refills the gas tank. The game starts with four lives, which are lost by running out of fuel, getting caught, or falling victim to one's own dynamite.
\end{minipage}%
\hfill%
\begin{minipage}[c]{0.38\textwidth}
\centering
\setlength{\fboxsep}{0pt}%
\setlength{\fboxrule}{0.4pt}%
\fcolorbox{gray}{white}{%
  \includegraphics[width=\linewidth,height=6.0cm,keepaspectratio]{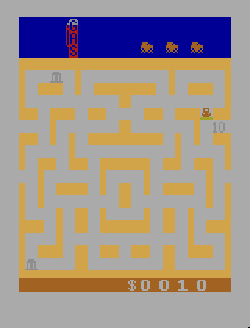}%
}
\end{minipage}

\subsubsection{Available Modifications for Bankheist}
\begin{table}[h]
    \centering
    \begin{tabular}{p{3cm} | p{10cm}} \toprule
        \textbf{Modification} & \textbf{Effect} \\ \midrule
        2\_police\_cars & Replaces 2 banks with police cars. Each bank robbed gives 50 points.\\
        blue\_player & Changes the color of the player to light blue.\\
        double\_speed & Doubles the player speed. \\
        dynamite\_penalty & Changes reward for killing a police car with dynamite (-500).\\
        fuel\_for\_banks & Augments the player's fuel when 3 banks have been robbed.\\
        golden\_banks & Changes the color of banks to golden.\\
        grey\_road & Changes the road color to grey.\\
        moving\_banks & Makes banks move like police cars.\\
        no\_police & Removes police cars from the game and automatically respawns banks.\\
        random\_city & Randomizes which city is entered next.\\
        random\_spawns & Restores the procedural, fully random bank spawns.\\
        red\_police\_cars & Changes the color of police cars to red.\\
        revisit\_city & Allows player to go to the previous city by going through the left portal.\\
        unlimited\_gas & Prevents gas consumption. Only dropping dynamite consumes gas.\\
        \bottomrule
    \end{tabular}
\end{table}


\subsubsection{Selection of Available Visual Modifications}
\begin{figure}[H]
  \centering
  \setlength{\fboxsep}{0pt}%
  \setlength{\fboxrule}{0.4pt}%
  \fcolorbox{gray}{white}{%
    \includegraphics[width=\linewidth,keepaspectratio]{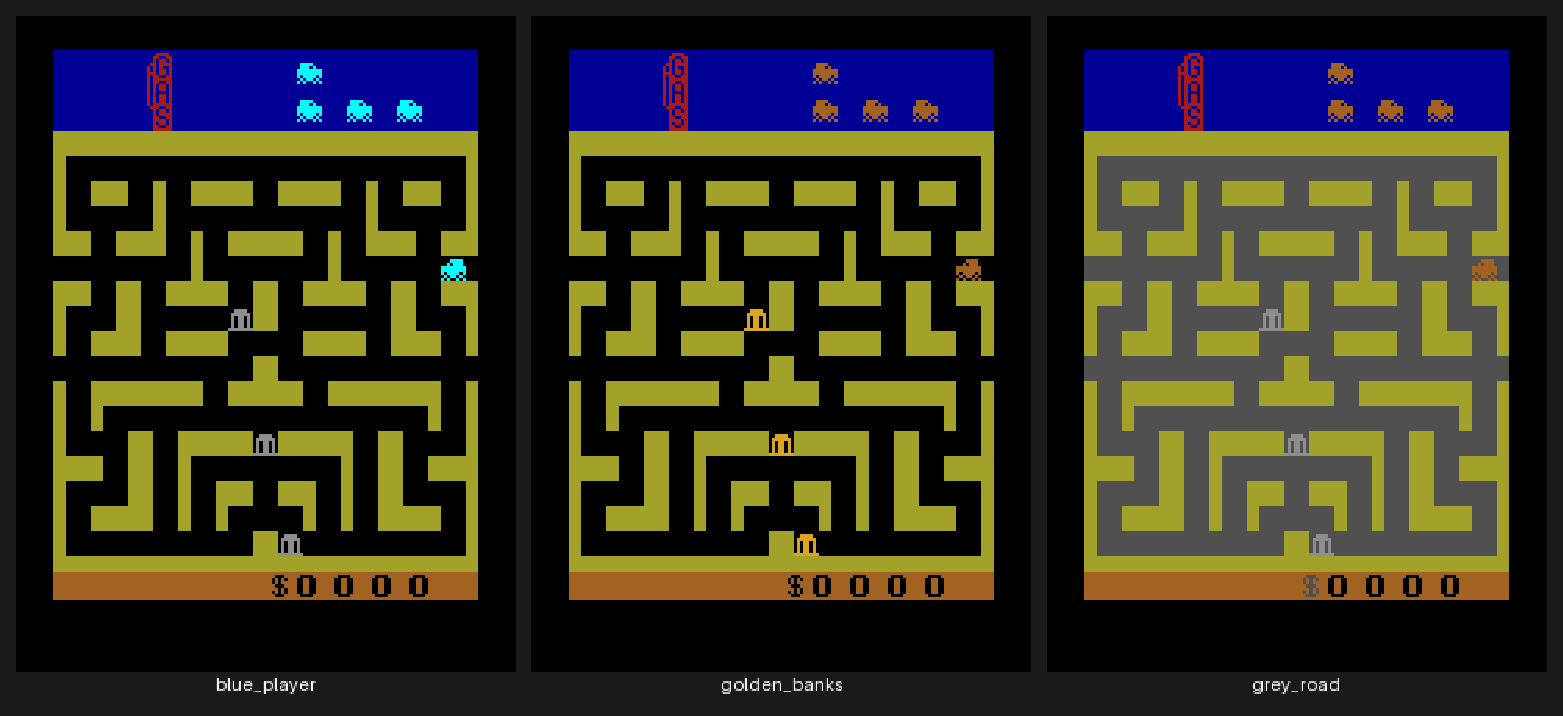}%
  }
  \caption{%
    A selection of visual modifications available for Bankheist,
    with each frame showing a different mod applied to the base environment.%
  }
  \label{fig:mods-Bankheist}
\end{figure}

\newpage

\subsection{Beamrider}
\subsubsection{Beamrider \ details}

\vspace{0.5cm}

\noindent
\begin{minipage}[c]{0.52\textwidth}
\raggedright
The player pilots a spaceship moving forward at constant speed, steerable only laterally across a set of discrete positions. The objective is to shoot down enemy vessels while evading their projectiles and navigating around space debris.
\end{minipage}%
\hfill%
\begin{minipage}[c]{0.38\textwidth}
\centering
\setlength{\fboxsep}{0pt}%
\setlength{\fboxrule}{0.4pt}%
\fcolorbox{gray}{white}{%
  \includegraphics[width=\linewidth,height=6.0cm,keepaspectratio]{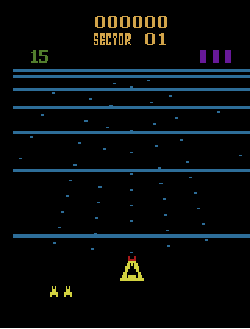}%
}
\end{minipage}

\subsubsection{Available Modifications for Beamrider}
\begin{table}[h!]
    \centering
    \begin{tabular}{p{3cm} | p{10cm}} \toprule
        \textbf{Modification} & \textbf{Effect} \\ \midrule
        dont\_kill & Internal mod that punishes killing and shooting.\\
        double\_enemy\_speed & Doubles enemy movement speed.\\
        fog\_of\_war & Hide the upper two thirds of the screen behind opaque fog.\\
        hardcore & Start with one life and never allow extra lives.\\
        mothership\_laser & Turn the mothership sequence into a stationary lane laser attack.\\
        same\_enemies & Render all enemy types using the white UFO visuals.\\
        teleport\_ufos & Adds a rare attack pattern for the UFOs.\\
        three\_lanes & Collapses strack to the center three lanes.\\
        toaster & Disable heavy visual effects while keeping gameplay unchanged.\\
        \bottomrule
    \end{tabular}
\end{table}

\subsubsection{Selection of Available Visual Modifications}
\begin{figure}[H]
  \centering
  \setlength{\fboxsep}{0pt}%
  \setlength{\fboxrule}{0.4pt}%
  \fcolorbox{gray}{white}{%
    \includegraphics[height=6cm,keepaspectratio]{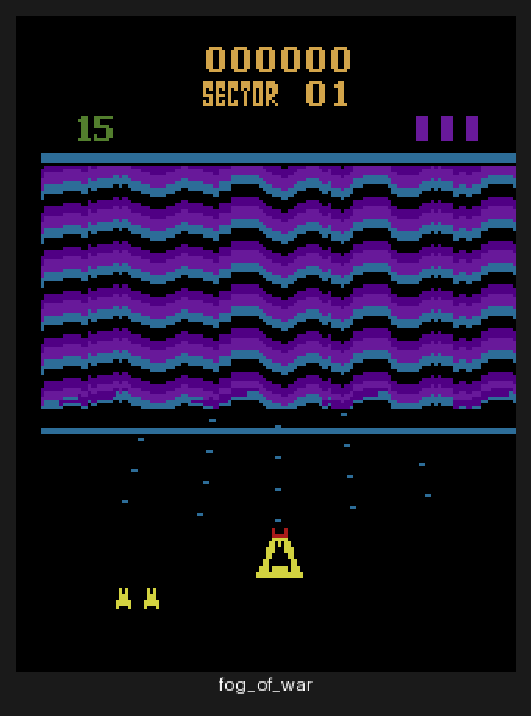}%
  }
  \caption{%
    A selection of visual modifications available for Beamrider,
    with each frame showing a different mod applied to the base environment.%
  }
  \label{fig:mods-Beamrider}
\end{figure}

\newpage

\subsection{Berzerk}
\subsubsection{Berzerk \ details}

\vspace{0.5cm}

\noindent
\begin{minipage}[c]{0.52\textwidth}
\raggedright
The player navigates a maze populated by evil robots that must be destroyed, while also avoiding contact with the maze walls, which is instantly fatal. Enough points can earn extra lives depending on the selected game mode. In certain modes, the indestructible Evil Otto appears and pursues the player, adding an additional threat that cannot be eliminated and must simply be outrun.
\end{minipage}%
\hfill%
\begin{minipage}[c]{0.38\textwidth}
\centering
\setlength{\fboxsep}{0pt}%
\setlength{\fboxrule}{0.4pt}%
\fcolorbox{gray}{white}{%
  \includegraphics[width=\linewidth,height=6.0cm,keepaspectratio]{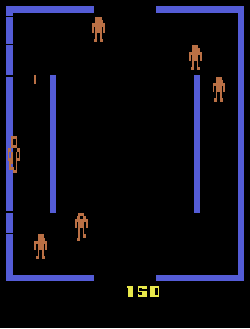}%
}
\end{minipage}

\subsubsection{Available Modifications for Berzerk}
\begin{table}[h]
    \centering
    \begin{tabular}{p{3cm} | p{10cm}} \toprule
        \textbf{Modification} & \textbf{Effect} \\ \midrule
       \multicolumn{2}{l}{\textit{No modifications implemented yet.}} \\
        \bottomrule
    \end{tabular}
\end{table}

\newpage

\subsection{Blackjack}
\subsubsection{Blackjack \ details}

\vspace{0.5cm}

\noindent
\begin{minipage}[c]{0.52\textwidth}
\raggedright
The player competes against a dealer in a game of Blackjack, drawing cards with the aim of reaching a total as close to 21 as possible without exceeding it and going bust.
\end{minipage}%
\hfill%
\begin{minipage}[c]{0.38\textwidth}
\centering
\setlength{\fboxsep}{0pt}%
\setlength{\fboxrule}{0.4pt}%
\fcolorbox{gray}{white}{%
  \includegraphics[width=\linewidth,height=6.0cm,keepaspectratio]{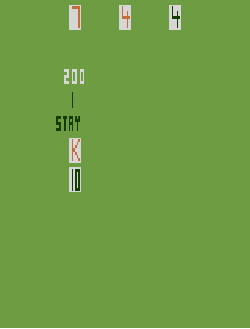}%
}
\end{minipage}

\subsubsection{Available Modifications for Blackjack}
\begin{table}[h]
    \centering
    \begin{tabular}{p{3cm} | p{10cm}} \toprule
        \textbf{Modification} & \textbf{Effect} \\ \midrule
       \multicolumn{2}{l}{\textit{No modifications implemented yet.}} \\
        \bottomrule
    \end{tabular}
\end{table}

\newpage

\subsection{Breakout}
\subsubsection{Breakout \ details}

\vspace{0.5cm}

\noindent
\begin{minipage}[c]{0.52\textwidth}
\raggedright
The player controls a paddle at the bottom of the screen, bouncing a ball against a wall of bricks above. Each hit destroys a brick, and breaking through to the other side allows the ball to clear bricks freely on its own. The game provides five lives before it ends.
\end{minipage}%
\hfill%
\begin{minipage}[c]{0.38\textwidth}
\centering
\setlength{\fboxsep}{0pt}%
\setlength{\fboxrule}{0.4pt}%
\fcolorbox{gray}{white}{%
  \includegraphics[width=\linewidth,height=6.0cm,keepaspectratio]{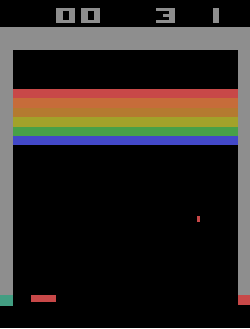}%
}
\end{minipage}

\subsubsection{Available Modifications for Breakout}
\begin{table}[h]
    \centering
    \begin{tabular}{p{3cm} | p{10cm}} \toprule
        \textbf{Modification} & \textbf{Effect} \\ \midrule
        ball\_color & Changes the balls color to a set color. \\
        ball\_drift & Consistently drift the ball to the right. \\
        ball\_gravity & Pulls the ball down by 1 every step of the gravity\_buffer. \\
        big\_paddle & Always use a bigger paddle (Size 40). \\
        block\_color & Changes the blocks color to a set color. \\
        player\_color & Changes the player color to a set color. \\
        small\_paddle & Always use a small paddle (Size 4). \\
        speed\_mode & Increase speed to maximum at all time steps. \\
        \bottomrule
    \end{tabular}
\end{table}

\subsubsection{Selection of Available Visual Modifications}
\begin{figure}[H]
  \centering
  \setlength{\fboxsep}{0pt}%
  \setlength{\fboxrule}{0.4pt}%
  \fcolorbox{gray}{white}{%
    \includegraphics[width=\linewidth,keepaspectratio]{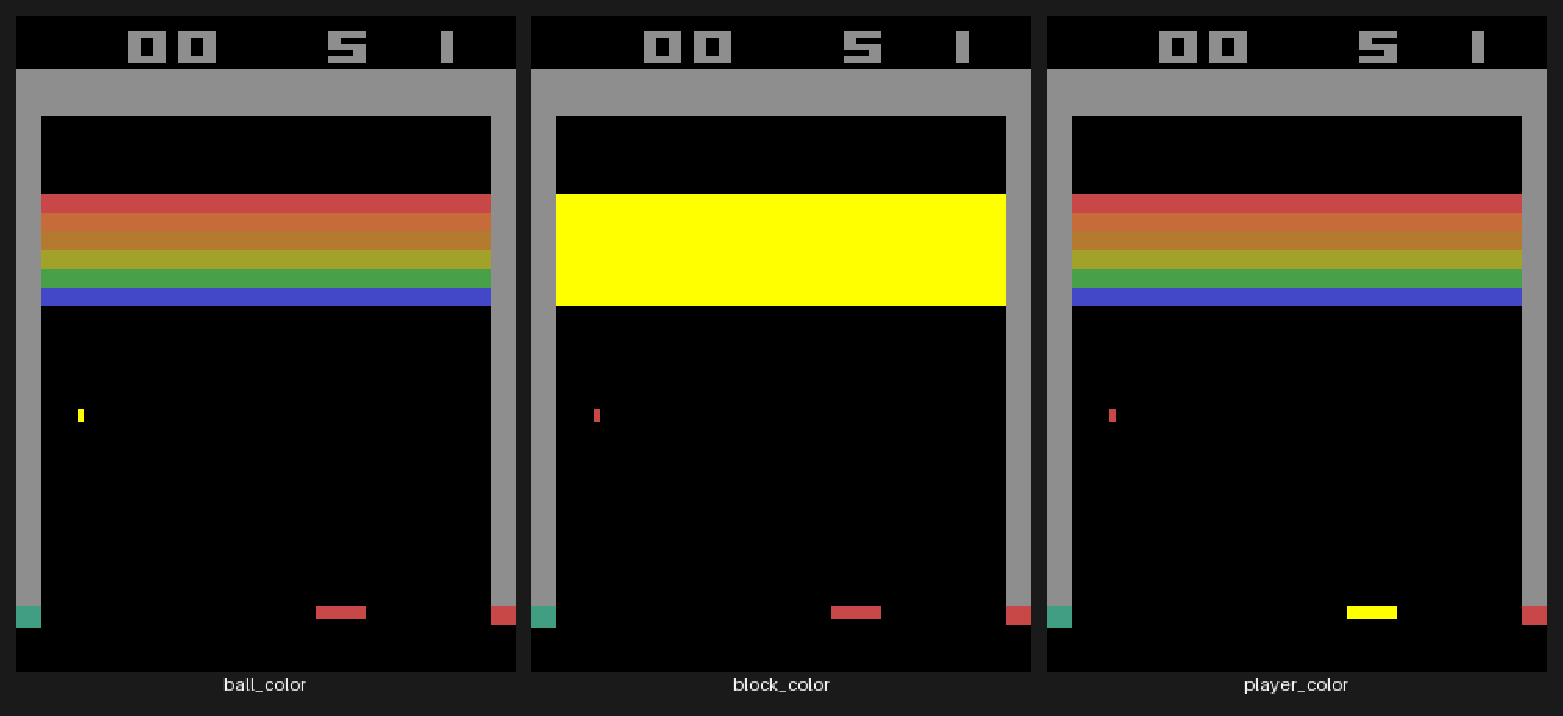}%
  }
  \caption{%
    A selection of visual modifications available for Breakout,
    with each frame showing a different mod applied to the base environment.%
  }
  \label{fig:mods-Breakout}
\end{figure}

\newpage

\subsection{Centipede}
\subsubsection{Centipede \ details}

\vspace{0.5cm}

\noindent
\begin{minipage}[c]{0.52\textwidth}
\raggedright
The player takes the role of an elf defending an enchanted forest, using magic wands to fend off waves of centipedes, spiders, and fleas that threaten the mushrooms. Being bitten by any creature causes temporary paralysis and costs one wand. Additional wands can be earned by reaching certain score thresholds, but the game ends once all wands are lost.
\end{minipage}%
\hfill%
\begin{minipage}[c]{0.38\textwidth}
\centering
\setlength{\fboxsep}{0pt}%
\setlength{\fboxrule}{0.4pt}%
\fcolorbox{gray}{white}{%
  \includegraphics[width=\linewidth,height=6.0cm,keepaspectratio]{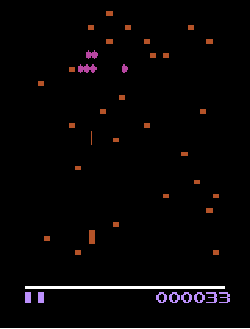}%
}
\end{minipage}

\subsubsection{Available Modifications for Centipede}
\begin{table}[h]
    \centering
    \begin{tabular}{p{3cm} | p{10cm}} \toprule
        \textbf{Modification} & \textbf{Effect} \\ \midrule
       \multicolumn{2}{l}{\textit{No modifications implemented yet.}} \\
        \bottomrule
    \end{tabular}
\end{table}

\newpage

\subsection{ChopperCommand}
\subsubsection{ChopperCommand \ details}

\vspace{0.5cm}

\noindent
\begin{minipage}[c]{0.52\textwidth}
\raggedright
The player pilots a helicopter tasked with protecting ground-based truck convoys by shooting down enemy aircraft. A mini-map at the bottom of the screen provides an overview of the surrounding area.
\end{minipage}%
\hfill%
\begin{minipage}[c]{0.38\textwidth}
\centering
\setlength{\fboxsep}{0pt}%
\setlength{\fboxrule}{0.4pt}%
\fcolorbox{gray}{white}{%
  \includegraphics[width=\linewidth,height=6.0cm,keepaspectratio]{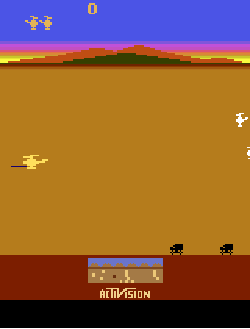}%
}
\end{minipage}

\subsubsection{Available Modifications for ChopperCommand}
\begin{table}[h]
    \centering
    \begin{tabular}{p{3cm} | p{10cm}} \toprule
        \textbf{Modification} & \textbf{Effect} \\ \midrule
       \multicolumn{2}{l}{\textit{No modifications implemented yet.}} \\
        \bottomrule
    \end{tabular}
\end{table}

\newpage

\subsection{Enduro}
\subsubsection{Enduro \ details}

\vspace{0.5cm}

\noindent
\begin{minipage}[c]{0.52\textwidth}
\raggedright
The player competes in the National Enduro, a long-distance endurance race. Each day a minimum number of cars must be overtaken to remain in the race: 200 on the first day and 300 on each subsequent day. Failing to meet the daily quota ends the game.
\end{minipage}%
\hfill%
\begin{minipage}[c]{0.38\textwidth}
\centering
\setlength{\fboxsep}{0pt}%
\setlength{\fboxrule}{0.4pt}%
\fcolorbox{gray}{white}{%
  \includegraphics[width=\linewidth,height=6.0cm,keepaspectratio]{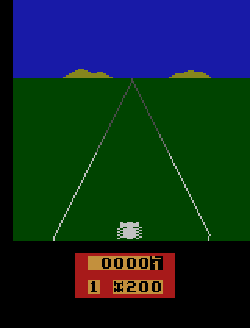}%
}
\end{minipage}

\subsubsection{Available Modifications for Enduro}
\begin{table}[h]
    \centering
    \begin{tabular}{p{3cm} | p{10cm}} \toprule
        \textbf{Modification} & \textbf{Effect} \\ \midrule
        dawn & Forces the weather to dawn. \\
        day & Forces the weather to day. \\
        dont\_max\_speed & Does not reward the player if at max. speed. \\
        filled\_road & Displays the full filled road instead of only the side. \\
        fog & Forces the weather to fog. \\
        hud & Renders a HUD that displays sped and x-position.\\
        max\_speed\_reward & Provides +2 every 20th frame when at maximum speed. \\
        night & Forces the weather to night. \\
        no\_opponents & Disables all opponent cars. \\
        short\_days & Iterates weather state every 16km driven instead of based on time. \\
        snow & Forces the weather to snow. \\
        start\_in\_curve & Starts the game directly in the curve. \\
        start\_in\_max\_curve & Starts the game in a full curve. \\
        sunset & Forces the weather to sunset. \\
        \bottomrule
    \end{tabular}
\end{table}

\subsubsection{Selection of Available Visual Modifications}
\begin{figure}[H]
  \centering
  \setlength{\fboxsep}{0pt}%
  \setlength{\fboxrule}{0.4pt}%
  \fcolorbox{gray}{white}{%
    \includegraphics[width=\linewidth,keepaspectratio]{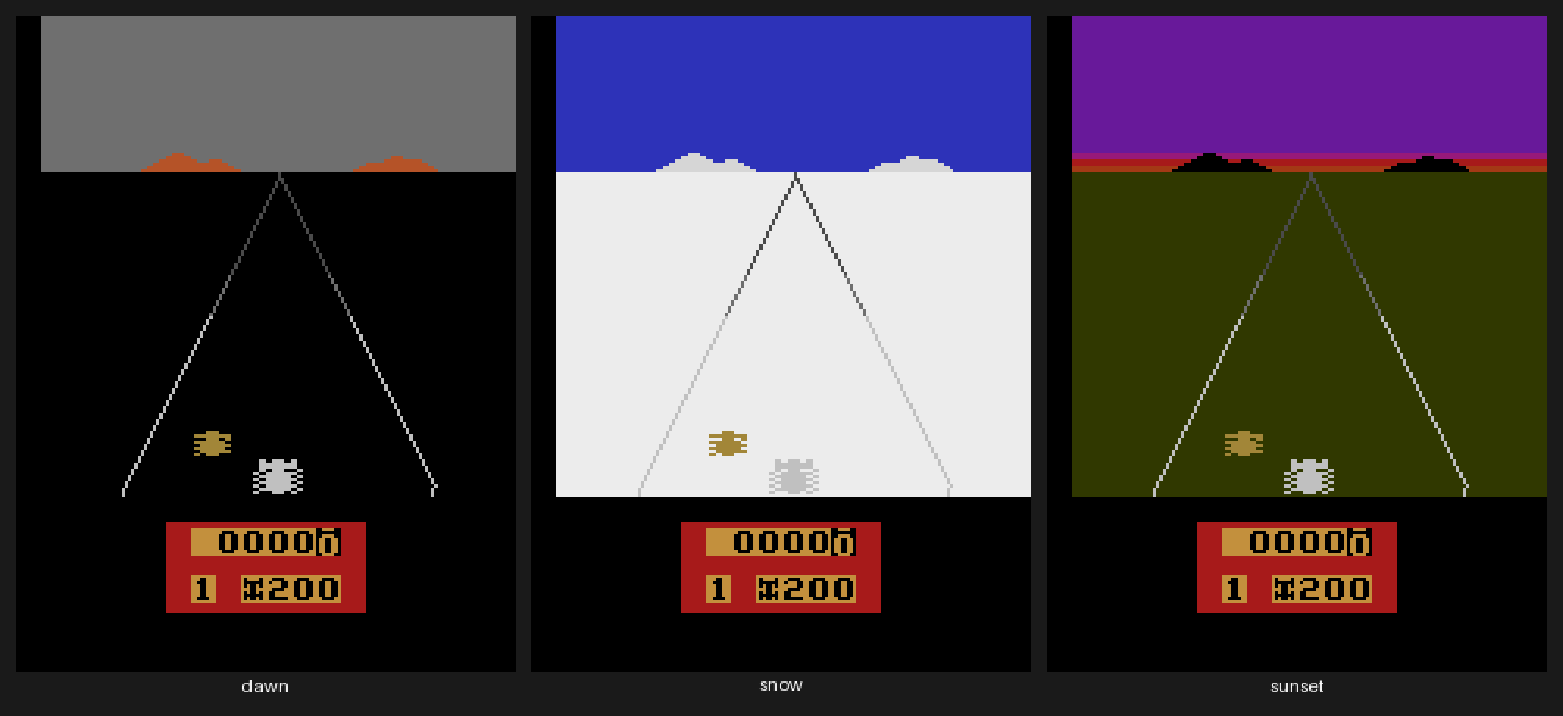}%
  }
  \caption{%
    A selection of visual modifications available for Enduro,
    with each frame showing a different mod applied to the base environment.%
  }
  \label{fig:mods-Enduro}
\end{figure}

\newpage

\subsection{Fishingderby}
\subsubsection{Fishingderby \ details}

\vspace{0.5cm}

\noindent
\begin{minipage}[c]{0.52\textwidth}
\raggedright
The player competes against an opponent to catch more sunfish, each fishing from opposite sides of a lake. Lines are cast downward through several layers of water, where fish of varying sizes swim by at different depths. The catch must be reeled in quickly, as sharks patrol the water and will steal the fish right off the line if given the chance. The first player to reach 99 points wins.
\end{minipage}%
\hfill%
\begin{minipage}[c]{0.38\textwidth}
\centering
\setlength{\fboxsep}{0pt}%
\setlength{\fboxrule}{0.4pt}%
\fcolorbox{gray}{white}{%
  \includegraphics[width=\linewidth,height=6.0cm,keepaspectratio]{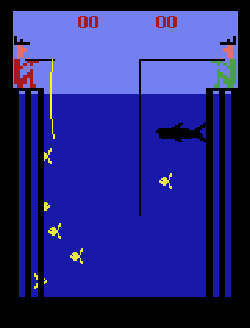}%
}
\end{minipage}

\subsubsection{Available Modifications for Fishingderby}
\begin{table}[h]
    \centering
    \begin{tabular}{p{4cm} | p{9cm}} \toprule
        \textbf{Modification} & \textbf{Effect} \\ \midrule
        fish\_in\_middle & All fish are bound to a small area in the middle (x-position).\\
        fish\_on\_different\_sides & Half the fish are bound to an area on the right, half on the left. \\
        fish\_on\_player\_side & All fish are bound to an area on the players side. \\
        shark\_no\_movement\_easy & Shark is stuck to the right pier. \\
        shark\_no\_movement\_middle & Shark is stuck to the center (x-position). \\
        \bottomrule
    \end{tabular}
\end{table}

\newpage

\subsection{Flagcapture}
\subsubsection{Flagcapture \ details}

\vspace{0.5cm}

\noindent
\begin{minipage}[c]{0.52\textwidth}
\raggedright
The player explores a grid-based world, moving from square to square while collecting clues scattered throughout. The ultimate objective is to locate and reach a hidden flag.
\end{minipage}%
\hfill%
\begin{minipage}[c]{0.38\textwidth}
\centering
\setlength{\fboxsep}{0pt}%
\setlength{\fboxrule}{0.4pt}%
\fcolorbox{gray}{white}{%
  \includegraphics[width=\linewidth,height=6.0cm,keepaspectratio]{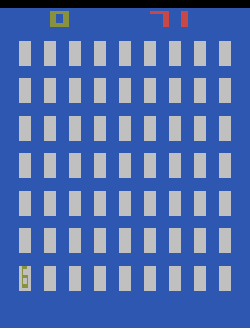}%
}
\end{minipage}

\subsubsection{Available Modifications for Flagcapture}
\begin{table}[h]
    \centering
    \begin{tabular}{p{3cm} | p{10cm}} \toprule
        \textbf{Modification} & \textbf{Effect} \\ \midrule
       \multicolumn{2}{l}{\textit{No modifications implemented yet.}} \\
        \bottomrule
    \end{tabular}
\end{table}

\newpage

\subsection{Freeway}
\subsubsection{Freeway \ details}

\vspace{0.5cm}

\noindent
\begin{minipage}[c]{0.52\textwidth}
\raggedright
The player guides a chicken across multiple lanes of dense rush hour traffic, aiming to reach the top of the screen. Each successful crossing earns a point, while collisions with vehicles send the chicken back toward the starting side.
\end{minipage}%
\hfill%
\begin{minipage}[c]{0.38\textwidth}
\centering
\setlength{\fboxsep}{0pt}%
\setlength{\fboxrule}{0.4pt}%
\fcolorbox{gray}{white}{%
  \includegraphics[width=\linewidth,height=6.0cm,keepaspectratio]{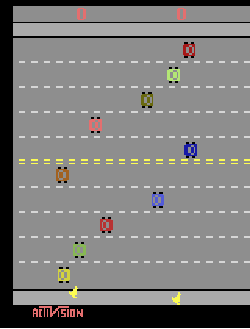}%
}
\end{minipage}

\subsubsection{Available Modifications for Freeway}
\begin{table}[h!]
    \centering
    \begin{tabular}{p{3cm} | p{10cm}} \toprule
        \textbf{Modification} & \textbf{Effect} \\ \midrule
        \_hall\_of\_fame\_start & Spawns the cars to make a "hall of fame" formation. \\
        bikes & Replaces all cars with uniquely colored bike sprites. \\
        black\_cars & Makes all cars black by overriding CAR\_COLORS constant. \\
        center\_cars\_on\_reset & Positions all cars in the center of the screen when the environment resets. \\
        change\_sprites & Bundles \textit{frog}, \textit{bikes}, \textit{new\_lane\_colors} and \textit{green\_score} \\
        frog & Replaces the player sprites with frog sprites. \\
        green\_score & Makes the score digits green. \\
        hall\_of\_fame &  Bundles \textit{\_hall\_of\_fame\_start} and \textit{static\_cars}.\\
        invert\_speed & Inverts the speed of all cars by overriding CAR\_UPDATES constant. \\
        new\_lane\_colors & Makes the lane separation black and the double lane separation red. \\
        slow\_cars & Halves the speed of all cars. \\
        static\_cars & Stops all cars after spawning. \\
        stop\_all\_cars & Stops all cars randomly with probability 0.4. \\
        too\_close\_penalty & Provides -1 reward if the chicken is too close to a car. \\
        \bottomrule
    \end{tabular}
\end{table}

\subsubsection{Selection of Available Visual Modifications}
\begin{figure}[H]
  \centering
  \setlength{\fboxsep}{0pt}%
  \setlength{\fboxrule}{0.4pt}%
  \fcolorbox{gray}{white}{%
    \includegraphics[width=\linewidth,keepaspectratio]{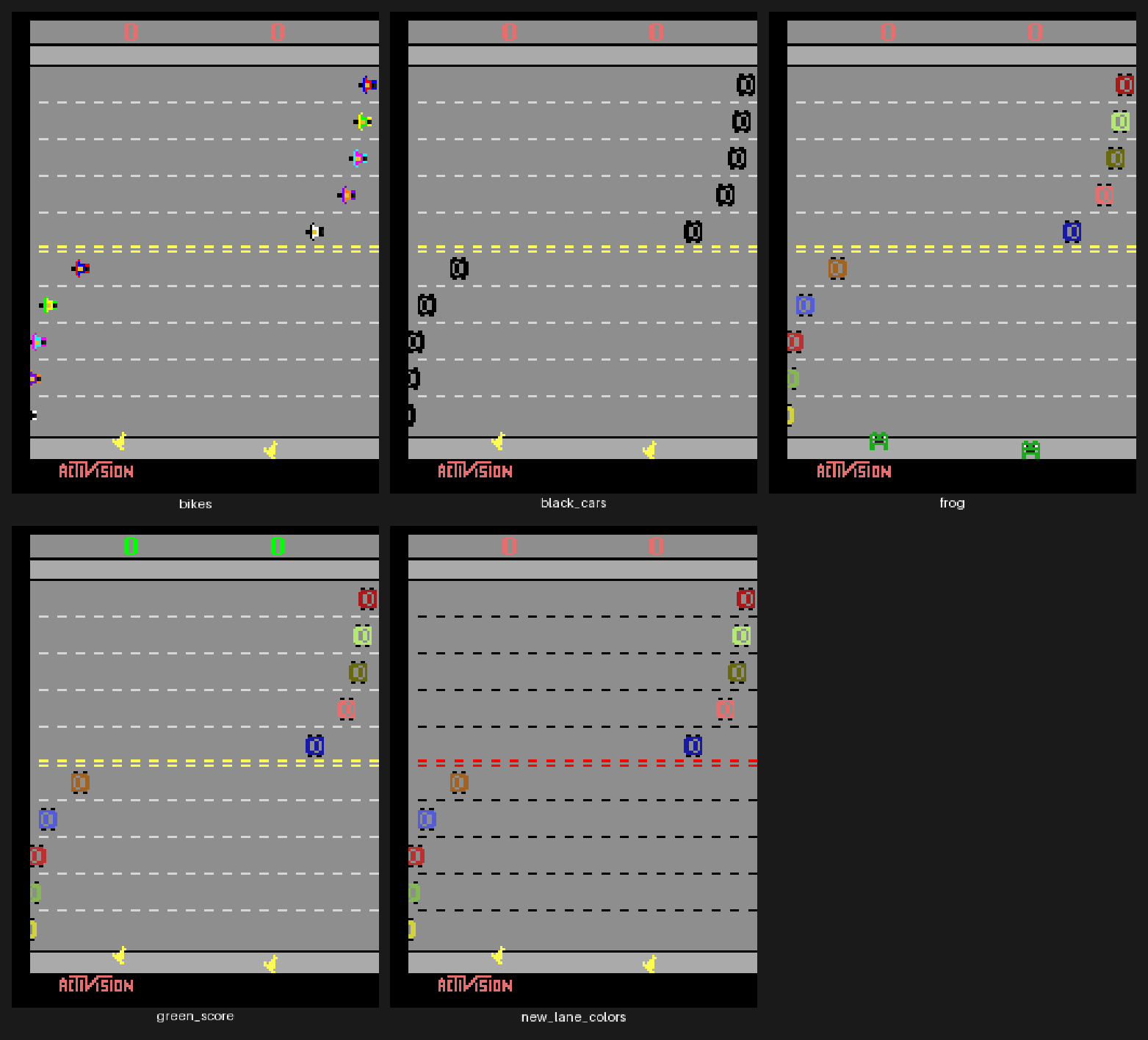}%
  }
  \caption{%
    A selection of visual modifications available for Freeway,
    with each frame showing a different mod applied to the base environment.%
  }
  \label{fig:mods-Freeway}
\end{figure}

\newpage

\subsection{Frostbite}
\subsubsection{Frostbite \ details}

\vspace{0.5cm}

\noindent
\begin{minipage}[c]{0.52\textwidth}
\raggedright
The player controls "Frostbite Bailey", hopping across an Arctic river and turning ice blocks from white to blue. Each converted block contributes to the construction of an igloo.
\end{minipage}%
\hfill%
\begin{minipage}[c]{0.38\textwidth}
\centering
\setlength{\fboxsep}{0pt}%
\setlength{\fboxrule}{0.4pt}%
\fcolorbox{gray}{white}{%
  \includegraphics[width=\linewidth,height=6.0cm,keepaspectratio]{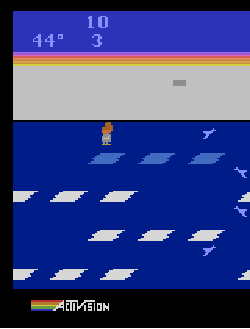}%
}
\end{minipage}

\subsubsection{Available Modifications for Frostbite}
\begin{table}[h]
    \centering
    \begin{tabular}{p{3cm} | p{10cm}} \toprule
        \textbf{Modification} & \textbf{Effect} \\ \midrule
        \_aligned\_ice & Makes the ice blocks perfectly centered. \\
        \_misaligned\_ice & Makes the ice blocks vertically aligned with each other but off-center. \\
        \_static\_ice & Sets ice block horizontal speeds to zero. \\
        bad\_fishes\_reward & Punishes the player for collecting fish. \\
        change\_sprites & Bundles \textit{tiger}, \textit{white\_igloo}, \textit{recolored\_obstacles} and \textit{lightblue\_ice}.\\
        dark\_night & Spawns constant night and makes the sky darker. \\
        early\_bear & Spawns the bear from the start (Level 1). \\
        left\_igloo & Places the igloo on the left side of the screen. \\
        lightblue\_ice & Changes the color of the ice blocks to light blue. \\
        no\_enemies & Disables all enemies and only keep fishes. \\
        recolored\_obstacles & Changes the colors of the floating obstacles. \\
        static\_aligned\_ice & Bundles \textit{\_static\_ice} and \textit{\_aligned\_ice}. \\
        static\_misaligned\_ice & Bundles \textit{\_static\_ice} and \textit{\_misaligned\_ice}. \\
        tiger & Switches the bear sprites to tiger sprites. \\
        white\_igloo & Makes the igloo white. \\
        \bottomrule
    \end{tabular}
\end{table}

\subsubsection{Selection of Available Visual Modifications}
\begin{figure}[H]
  \centering
  \setlength{\fboxsep}{0pt}%
  \setlength{\fboxrule}{0.4pt}%
  \fcolorbox{gray}{white}{%
    \includegraphics[height=6cm,keepaspectratio]{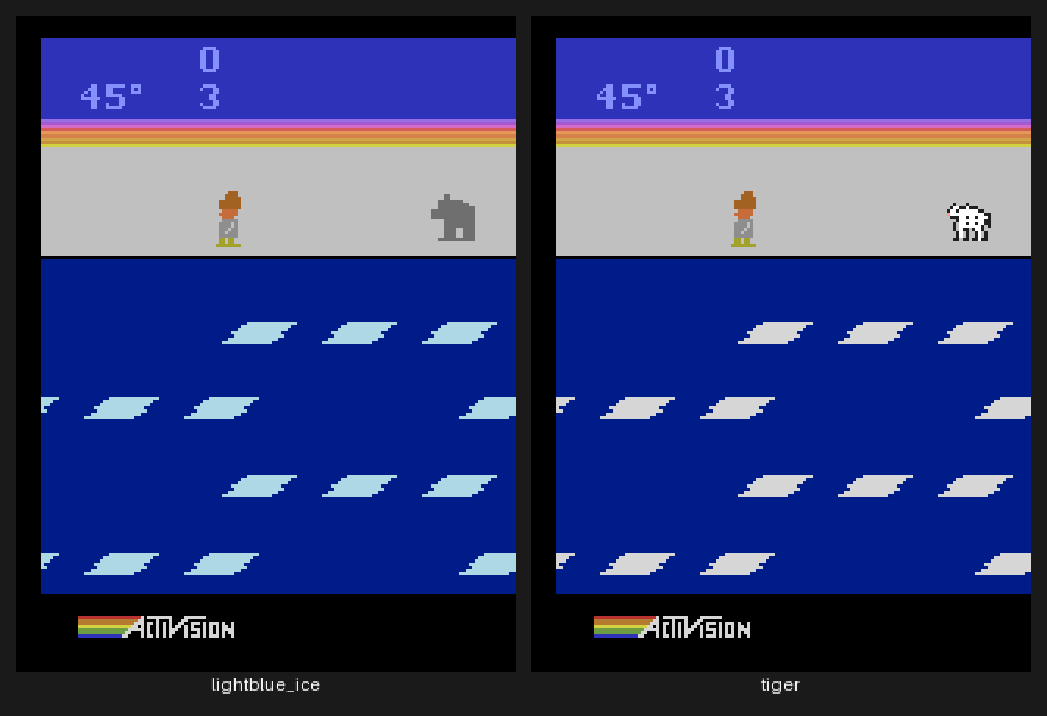}%
  }
  \caption{%
    A selection of visual modifications available for Frostbite,
    with each frame showing a different mod applied to the base environment.%
  }
  \label{fig:mods-Frostbite}
\end{figure}

\newpage

\subsection{Galaxian}
\subsubsection{Galaxian \ details}

\vspace{0.5cm}

\noindent
\begin{minipage}[c]{0.52\textwidth}
\raggedright
The player defends a laser base against successive waves of Galaxian invaders. Each wave must be cleared by shooting down all attackers before they reach the base.
\end{minipage}%
\hfill%
\begin{minipage}[c]{0.38\textwidth}
\centering
\setlength{\fboxsep}{0pt}%
\setlength{\fboxrule}{0.4pt}%
\fcolorbox{gray}{white}{%
  \includegraphics[width=\linewidth,height=6.0cm,keepaspectratio]{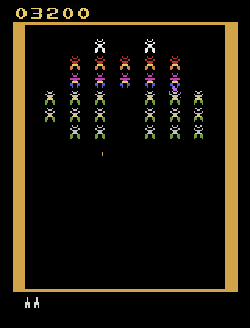}%
}
\end{minipage}

\subsubsection{Available Modifications for Galaxian}
\begin{table}[h]
    \centering
    \begin{tabular}{p{3cm} | p{10cm}} \toprule
        \textbf{Modification} & \textbf{Effect} \\ \midrule
       \multicolumn{2}{l}{\textit{No modifications implemented yet.}} \\
        \bottomrule
    \end{tabular}
\end{table}

\newpage

\subsection{Gravitar}
\subsubsection{Gravitar \ details}

\vspace{0.5cm}

\noindent
\begin{minipage}[c]{0.52\textwidth}
\raggedright
The player pilots a small spacecraft through a fictional solar system, free to explore its planets. Flying into a planet transitions the game to a side-view surface landscape.
\end{minipage}%
\hfill%
\begin{minipage}[c]{0.38\textwidth}
\centering
\setlength{\fboxsep}{0pt}%
\setlength{\fboxrule}{0.4pt}%
\fcolorbox{gray}{white}{%
  \includegraphics[width=\linewidth,height=6.0cm,keepaspectratio]{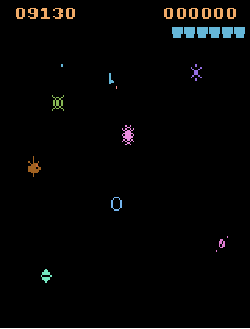}%
}
\end{minipage}

\subsubsection{Available Modifications for Gravitar}
\begin{table}[h]
    \centering
    \begin{tabular}{p{3cm} | p{10cm}} \toprule
        \textbf{Modification} & \textbf{Effect} \\ \midrule
        anti\_gravity & Reverse gravity from sun, planets, and reactors. \\
        fuel\_crisis & Increase fuel consumption rate by 5x. \\
        grayscale & Converts the visual palette to grayscale. \\
        harmless\_enemies & Make all enemies harmless by disabling their bullets. \\
        high\_speed & Make the ship faster by increasing thrust power and max speed. \\
        hyper\_gravity & Increase all gravity from sun, planets, and reactors. \\
        infinite\_fuel & Disable fuel consumption. \\
        inverted\_colors & Inverts the primary colors. \\
        long\_range\_tractor & Increase the range of the tractor beam. \\
        neon\_mode & Changes colors to bright neon variants. \\
        rapid\_fire & Increase active bullet caps for ship, saucer, and enemies to 4. \\
        red\_alert & Makes all terrain red/orange for a high-alert aesthetic. \\
        sharpshooter & Punishes shooting (-100) and rewards killing enemies (+500). \\
        slow\_enemies & Decrease the movement speed of saucers and bullets. \\
        valuable\_reactor & Populate reactor level with 3 enemies and 2 fuel tanks. \\
        zero\_gravity & Disable all gravity from sun, planets, and reactors. \\
        \bottomrule
    \end{tabular}
\end{table}

\subsubsection{Selection of Available Visual Modifications}
\begin{figure}[H]
  \centering
  \setlength{\fboxsep}{0pt}%
  \setlength{\fboxrule}{0.4pt}%
  \fcolorbox{gray}{white}{%
    \includegraphics[width=\linewidth,keepaspectratio]{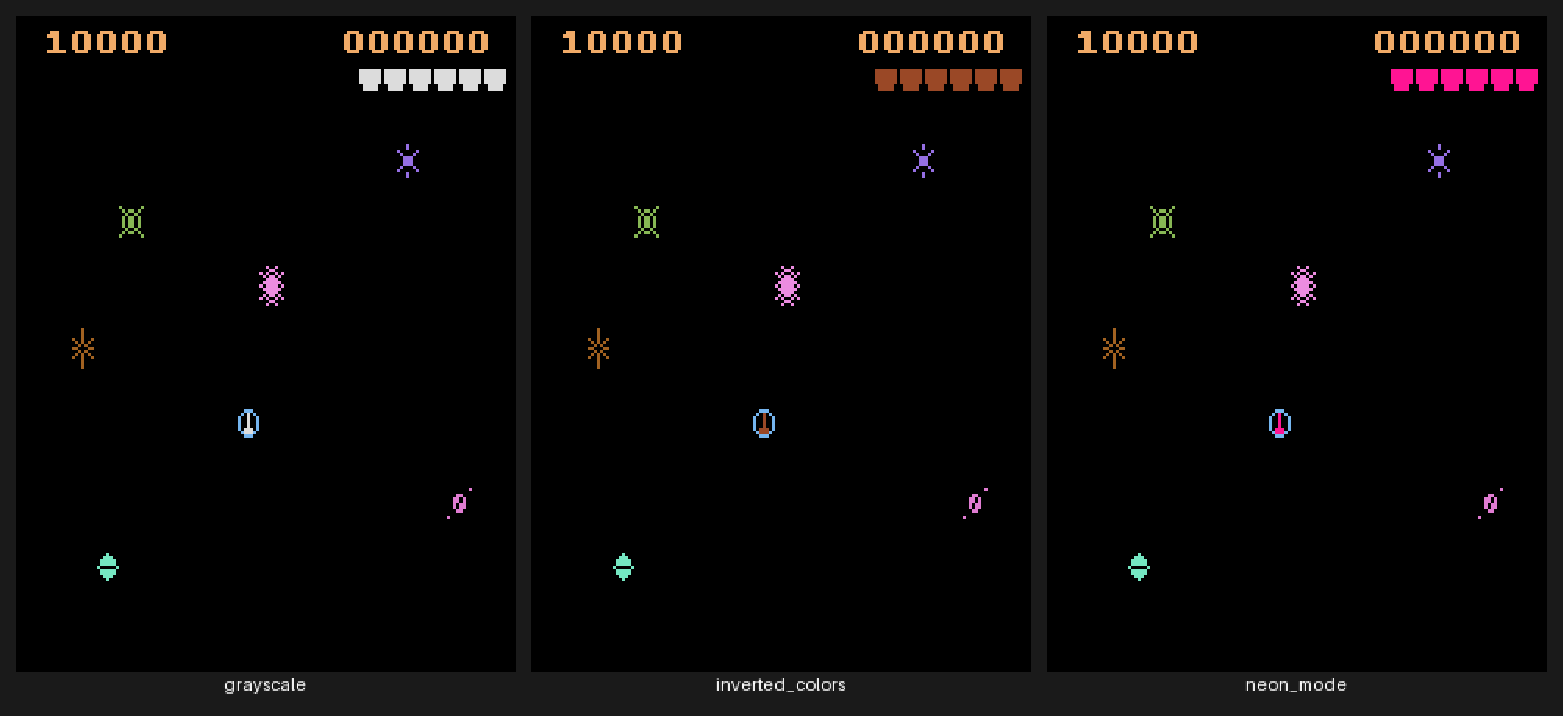}%
  }
  \caption{%
    A selection of visual modifications available for Gravitar,
    with each frame showing a different mod applied to the base environment.%
  }
  \label{fig:mods-Gravitar}
\end{figure}

\newpage

\subsection{HauntedHouse}
\subsubsection{HauntedHouse \ details}

\vspace{0.5cm}

\noindent
\begin{minipage}[c]{0.52\textwidth}
\raggedright
The player explores a mansion haunted by the ghost of the vengeful Samuel Graves, searching for three pieces of a magic urn. The mansion must be escaped with all three pieces before exhausting the nine available lives.
\end{minipage}%
\hfill%
\begin{minipage}[c]{0.38\textwidth}
\centering
\setlength{\fboxsep}{0pt}%
\setlength{\fboxrule}{0.4pt}%
\fcolorbox{gray}{white}{%
  \includegraphics[width=\linewidth,height=6.0cm,keepaspectratio]{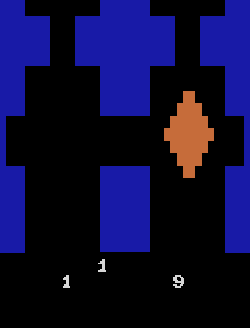}%
}
\end{minipage}

\subsubsection{Available Modifications for HauntedHouse}
\begin{table}[h]
    \centering
    \begin{tabular}{p{3cm} | p{10cm}} \toprule
        \textbf{Modification} & \textbf{Effect} \\ \midrule
       \multicolumn{2}{l}{\textit{No modifications implemented yet.}} \\
        \bottomrule
    \end{tabular}
\end{table}

\newpage

\subsection{HumanCannonball}
\subsubsection{HumanCannonball \ details}

\vspace{0.5cm}

\noindent
\begin{minipage}[c]{0.52\textwidth}
\raggedright
The player fires a human cannonball and must angle the shot to land the performer precisely into a water tower target.
\end{minipage}%
\hfill%
\begin{minipage}[c]{0.38\textwidth}
\centering
\setlength{\fboxsep}{0pt}%
\setlength{\fboxrule}{0.4pt}%
\fcolorbox{gray}{white}{%
  \includegraphics[width=\linewidth,height=6.0cm,keepaspectratio]{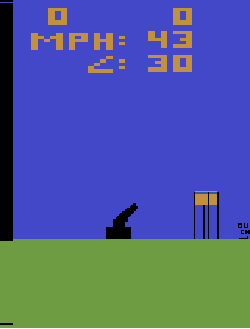}%
}
\end{minipage}

\subsubsection{Available Modifications for HumanCannonball}
\begin{table}[h]
    \centering
    \begin{tabular}{p{3cm} | p{10cm}} \toprule
        \textbf{Modification} & \textbf{Effect} \\ \midrule
       \multicolumn{2}{l}{\textit{No modifications implemented yet.}} \\
        \bottomrule
    \end{tabular}
\end{table}

\newpage

\subsection{Kangaroo}
\subsubsection{Kangaroo \ details}

\vspace{0.5cm}

\noindent
\begin{minipage}[c]{0.52\textwidth}
\raggedright
The player controls Mother Kangaroo on a mission to rescue her baby, starting with three lives. The rescue involves navigating a series of obstacles: climbing ladders, collecting bonus fruit, and throwing punches at interfering monkeys. The goal is to complete the rescue while accumulating as many points as possible.
\end{minipage}%
\hfill%
\begin{minipage}[c]{0.38\textwidth}
\centering
\setlength{\fboxsep}{0pt}%
\setlength{\fboxrule}{0.4pt}%
\fcolorbox{gray}{white}{%
  \includegraphics[width=\linewidth,height=6.0cm,keepaspectratio]{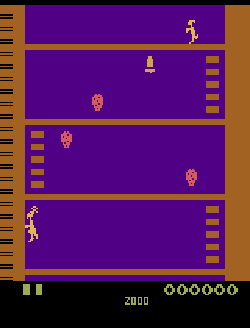}%
}
\end{minipage}

\subsubsection{Available Modifications for Kangaroo}
\begin{table}[h]
    \centering
    \begin{tabular}{p{5cm} | p{8cm}} \toprule
        \textbf{Modification} & \textbf{Effect} \\ \midrule
        \_chickens & Replaces the monkey sprite with a chicken sprite. \\
        \_lethal\_bell & Post-step mod that kills the player if they touch the bell. \\
        center\_ladders & Internal mod to center all ladder positions horizontally. \\
        chains & Replaces the ladder sprites with a chain. \\
        chickens & Bundles \textit{\_chickens} and \textit{no\_thrown\_coconut}. \\
        dont\_punch & Provides negative reward for punching monkeys. \\
        first\_level\_only & Internal mod to force the game to always stay on level 1. \\
        four\_ladders & Internal mod to add a fourth ladder to each level. \\
        high\_thrown\_coconuts & Forces coconuts to always spawn at the high position. \\
        invert\_ladders & Internal mod to invert all ladder positions (horizontally). \\
        no\_bell & Internal mod to disable the Bell. \\
        no\_danger & Bundles \textit{no\_monkey} and \textit{no\_falling\_coconut}. \\
        no\_falling\_coconut & Internal mod to disable the single falling coconut. \\
        no\_fruit & Internal mod to remove Fruits. \\
        no\_monkey & Internal mod to disable monkeys. \\
        no\_thrown\_coconut & Internal mod to disable thrown coconuts. \\
        pin\_child & Post-step mod to pin the child kangaroo in place. \\
        render\_debug\_info & Renders the player's X,Y coordinates. \\
        replace\_bell\_with\_danger\_sign & Replaces the bell sprite with a danger sign. \\
        replace\_child\_with\_monkey & Replaces the child sprite with a monkey sprite. \\
        replace\_coconut\_fireball & Replaces the coconut sprite with a fireball sprite. \\
        ropes & Replaces the ladder sprites with a rope. \\
        second\_level\_only & Internal mod to force the game to always stay on level 2. \\
        spawn\_on\_second\_floor & Spawns the player on the second level position. \\
        third\_level\_only & Internal mod to force the game to always stay on level 3. \\
        \bottomrule
    \end{tabular}
\end{table}

\subsubsection{Selection of Available Visual Modifications}
\begin{figure}[H]
  \centering
  \setlength{\fboxsep}{0pt}%
  \setlength{\fboxrule}{0.4pt}%
  \fcolorbox{gray}{white}{%
    \includegraphics[width=\linewidth,keepaspectratio]{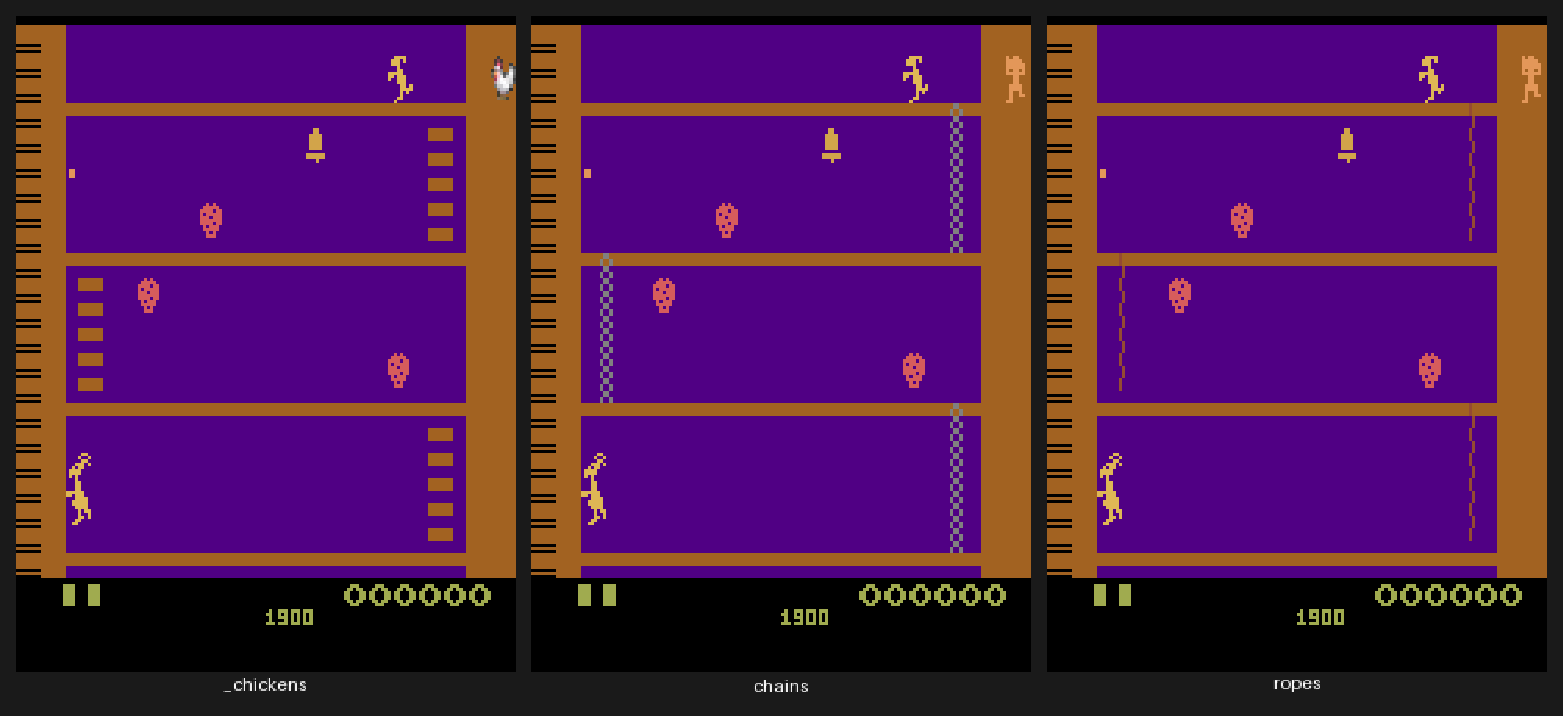}%
  }
  \caption{%
    A selection of visual modifications available for Kangaroo,
    with each frame showing a different mod applied to the base environment.%
  }
  \label{fig:mods-Kangaroo}
\end{figure}

\newpage

\subsection{KingKong}
\subsubsection{KingKong \ details}

\vspace{0.5cm}

\noindent
\begin{minipage}[c]{0.52\textwidth}
\raggedright
The player climbs the Empire State Building to rescue a person kidnapped by King Kong, who hurls bombs downward to impede the ascent. The bombs must be dodged throughout the climb to reach the top safely.
\end{minipage}%
\hfill%
\begin{minipage}[c]{0.38\textwidth}
\centering
\setlength{\fboxsep}{0pt}%
\setlength{\fboxrule}{0.4pt}%
\fcolorbox{gray}{white}{%
  \includegraphics[width=\linewidth,height=6.0cm,keepaspectratio]{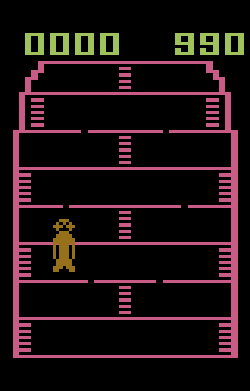}%
}
\end{minipage}

\subsubsection{Available Modifications for KingKong}
\begin{table}[h]
    \centering
    \begin{tabular}{p{3cm} | p{10cm}} \toprule
        \textbf{Modification} & \textbf{Effect} \\ \midrule
       \multicolumn{2}{l}{\textit{No modifications implemented yet.}} \\
        \bottomrule
    \end{tabular}
\end{table}

\newpage

\subsection{LaserGates}
\subsubsection{LaserGates \ details}

\vspace{0.5cm}

\noindent
\begin{minipage}[c]{0.52\textwidth}
\raggedright
The player navigates through a malfunctioning computer system using the Dante Dart, with the goal of locating and destroying four Failsafe Detonators scattered within.
\end{minipage}%
\hfill%
\begin{minipage}[c]{0.38\textwidth}
\centering
\setlength{\fboxsep}{0pt}%
\setlength{\fboxrule}{0.4pt}%
\fcolorbox{gray}{white}{%
  \includegraphics[width=\linewidth,height=6.0cm,keepaspectratio]{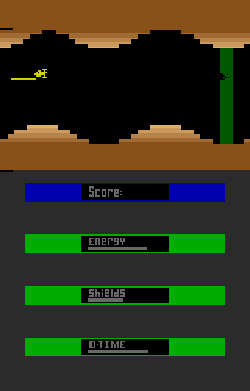}%
}
\end{minipage}

\subsubsection{Available Modifications for LaserGates}
\begin{table}[h]
    \centering
    \begin{tabular}{p{3cm} | p{10cm}} \toprule
        \textbf{Modification} & \textbf{Effect} \\ \midrule
       \multicolumn{2}{l}{\textit{No modifications implemented yet.}} \\
        \bottomrule
    \end{tabular}
\end{table}

\newpage

\subsection{MontezumaRevenge}
\subsubsection{MontezumaRevenge \ details}

\vspace{0.5cm}

\noindent
\begin{minipage}[c]{0.52\textwidth}
\raggedright
The player ventures through a sprawling maze of chambers inside the emperor's fortress in search of Montezuma's treasure. Along the way, deadly creatures must be avoided while collecting valuables and useful tools that aid in the escape with the treasure.
\end{minipage}%
\hfill%
\begin{minipage}[c]{0.38\textwidth}
\centering
\setlength{\fboxsep}{0pt}%
\setlength{\fboxrule}{0.4pt}%
\fcolorbox{gray}{white}{%
  \includegraphics[width=\linewidth,height=6.0cm,keepaspectratio]{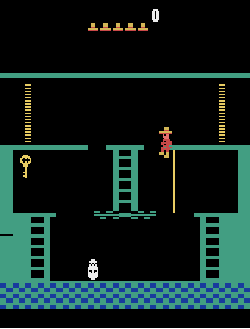}%
}
\end{minipage}

\subsubsection{Available Modifications for MontezumaRevenge}
\begin{table}[h]
    \centering
    \begin{tabular}{p{3cm} | p{10cm}} \toprule
        \textbf{Modification} & \textbf{Effect} \\ \midrule
        center\_bouncing\_skull & Make the rolling skull jump vertically at the center of the screen. \\
        debug\_hud & Internal mod to display debug info (Room ID, X, Y) on the HUD. \\
        fast\_player & Internal mod to increase player speed. \\
        infinite\_amulet & Post-step mod to keep the amulet active forever. \\
        jumping\_spiders & Post-step mod to make all spiders jump. \\
        moving\_snakes & Post-step mod to make snakes move horizontally in all rooms. \\
        no\_enemies & Post-step mod to immediately remove any enemies in the current room. \\
        no\_fall\_damage & Internal mod to disable fall damage. \\
        reveal\_map & Internal mod to make dark rooms visible, by granting the torch effect. \\
        rolling\_skulls & Make the two skulls that usually bump roll instead, \\
        super\_jump & Internal mod to increase jump height. \\
        sword\_kill\_bonus & Provide bonus points if the player kills an enemy with the sword. \\
        three\_swords & Post-step mod to start the player with 3 swords. \\
        \bottomrule
    \end{tabular}
\end{table}

\newpage

\subsection{MsPacman}
\subsubsection{MsPacman \ details}

\vspace{0.5cm}

\noindent
\begin{minipage}[c]{0.52\textwidth}
\raggedright
The player guides Ms. Pac-Man through a maze, collecting all pellets while evading four pursuing ghosts. Power pellets located at the corners of the maze temporarily turn the ghosts vulnerable, allowing Ms. Pac-Man to eat them for bonus points. Fruit items occasionally appear in the maze for additional score.
\end{minipage}%
\hfill%
\begin{minipage}[c]{0.38\textwidth}
\centering
\setlength{\fboxsep}{0pt}%
\setlength{\fboxrule}{0.4pt}%
\fcolorbox{gray}{white}{%
  \includegraphics[width=\linewidth,height=6.0cm,keepaspectratio]{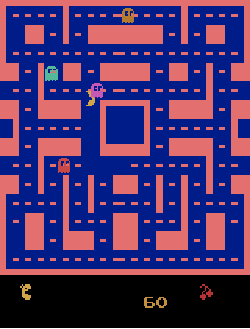}%
}
\end{minipage}

\subsubsection{Available Modifications for MsPacman}
\begin{table}[h]
    \centering
    \begin{tabular}{p{4cm} | p{9cm}} \toprule
        \textbf{Modification} & \textbf{Effect} \\ \midrule
        caged\_ghosts & Keeps the ghosts caged in the center. \\
        constant\_fruits & Always adds a fruit to the maze. \\
        fruit\_ghost\_bonus & Deactivates points for pellets, But multiplies eating ghost reward.\\
        matrix\_theme & A Matrix-themed mod: black background, green elements. \\
        only\_1\_ghost & Keeps only 1 ghost in the maze. \\
        only\_2\_ghost & Keeps only 2 ghost in the maze. \\
        only\_3\_ghost & Keeps only 3 ghost in the maze. \\
        random\_ghost\_navigation & Randomizes the navigation of all ghosts. \\
        set\_maze\_1 & Forces the level to always be maze 1. \\
        set\_maze\_2 & Forces the level to always be maze 2. \\
        set\_maze\_3 & Forces the level to always be maze 3. \\
        set\_maze\_4 & Forces the level to always be maze 4. \\
        \bottomrule
    \end{tabular}
\end{table}

\subsubsection{Selection of Available Visual Modifications}
\begin{figure}[H]
  \centering
  \setlength{\fboxsep}{0pt}%
  \setlength{\fboxrule}{0.4pt}%
  \fcolorbox{gray}{white}{%
    \includegraphics[height=6cm,keepaspectratio]{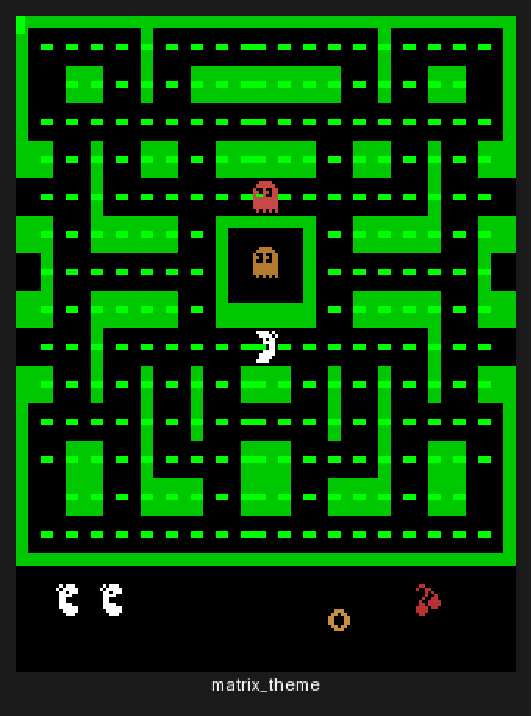}%
  }
  \caption{%
    A selection of visual modifications available for MsPacman,
    with each frame showing a different mod applied to the base environment.%
  }
  \label{fig:mods-MsPacman}
\end{figure}

\newpage

\subsection{NameThisGame}
\subsubsection{NameThisGame \ details}

\vspace{0.5cm}

\noindent
\begin{minipage}[c]{0.52\textwidth}
\raggedright
The player dives to protect a discovered underwater treasure, fending off attacks from both a shark and an octopus. Oxygen supplies must be carefully monitored throughout, adding an additional layer of urgency to the defense.
\end{minipage}%
\hfill%
\begin{minipage}[c]{0.38\textwidth}
\centering
\setlength{\fboxsep}{0pt}%
\setlength{\fboxrule}{0.4pt}%
\fcolorbox{gray}{white}{%
  \includegraphics[width=\linewidth,height=6.0cm,keepaspectratio]{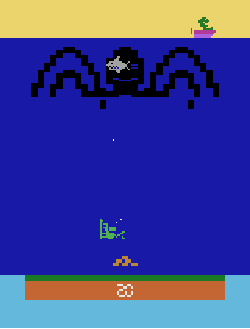}%
}
\end{minipage}

\subsubsection{Available Modifications for NameThisGame}
\begin{table}[h]
    \centering
    \begin{tabular}{p{3cm} | p{10cm}} \toprule
        \textbf{Modification} & \textbf{Effect} \\ \midrule
       \multicolumn{2}{l}{\textit{No modifications implemented yet.}} \\
        \bottomrule
    \end{tabular}
\end{table}

\newpage

\subsection{Phoenix}
\subsubsection{Phoenix \ details}

\vspace{0.5cm}

\noindent
\begin{minipage}[c]{0.52\textwidth}
\raggedright
The player fights through successive waves of enemy war birds, shooting them down while dodging their bombs. The ultimate objective is to reach and defeat the alien pilot commanding the assault.
\end{minipage}%
\hfill%
\begin{minipage}[c]{0.38\textwidth}
\centering
\setlength{\fboxsep}{0pt}%
\setlength{\fboxrule}{0.4pt}%
\fcolorbox{gray}{white}{%
  \includegraphics[width=\linewidth,height=6.0cm,keepaspectratio]{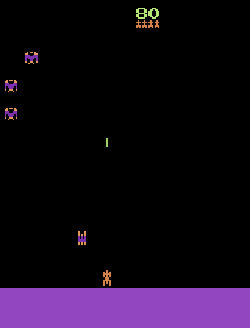}%
}
\end{minipage}

\subsubsection{Available Modifications for Phoenix}
\begin{table}[h]
    \centering
    \begin{tabular}{p{4cm} | p{9cm}} \toprule
        \textbf{Modification} & \textbf{Effect} \\ \midrule
        blood\_moon & A dark red themed mod. \\
        boss\_late\_missiles & Make boss missiles appear later. \\
        fast\_enemy\_bullets & Increases speed of enemy projectiles. \\
        fast\_player & Increases player movement speed. \\
        grayscale & Turns the entire game into grayscale. \\
        infinite\_lives & Set player lives to 99. \\
        inverted\_colors & Inverts all colors in the game. \\
        invincible\_player & Player is always invincible. \\
        matrix\_theme & A Matrix-themed mod: black background, green elements. \\
        night\_mode & Dims the entire screen by 50\% for a night mode experience. \\
        no\_ability\_cooldown & Removes cooldown for the special ability (shield). \\
        penalize\_shoot\_cooldown & Penalize for each time the player shoots (-50). \\
        \bottomrule
    \end{tabular}
\end{table}

\subsubsection{Selection of Available Visual Modifications}
\begin{figure}[H]
  \centering
  \setlength{\fboxsep}{0pt}%
  \setlength{\fboxrule}{0.4pt}%
  \fcolorbox{gray}{white}{%
    \includegraphics[width=\linewidth,keepaspectratio]{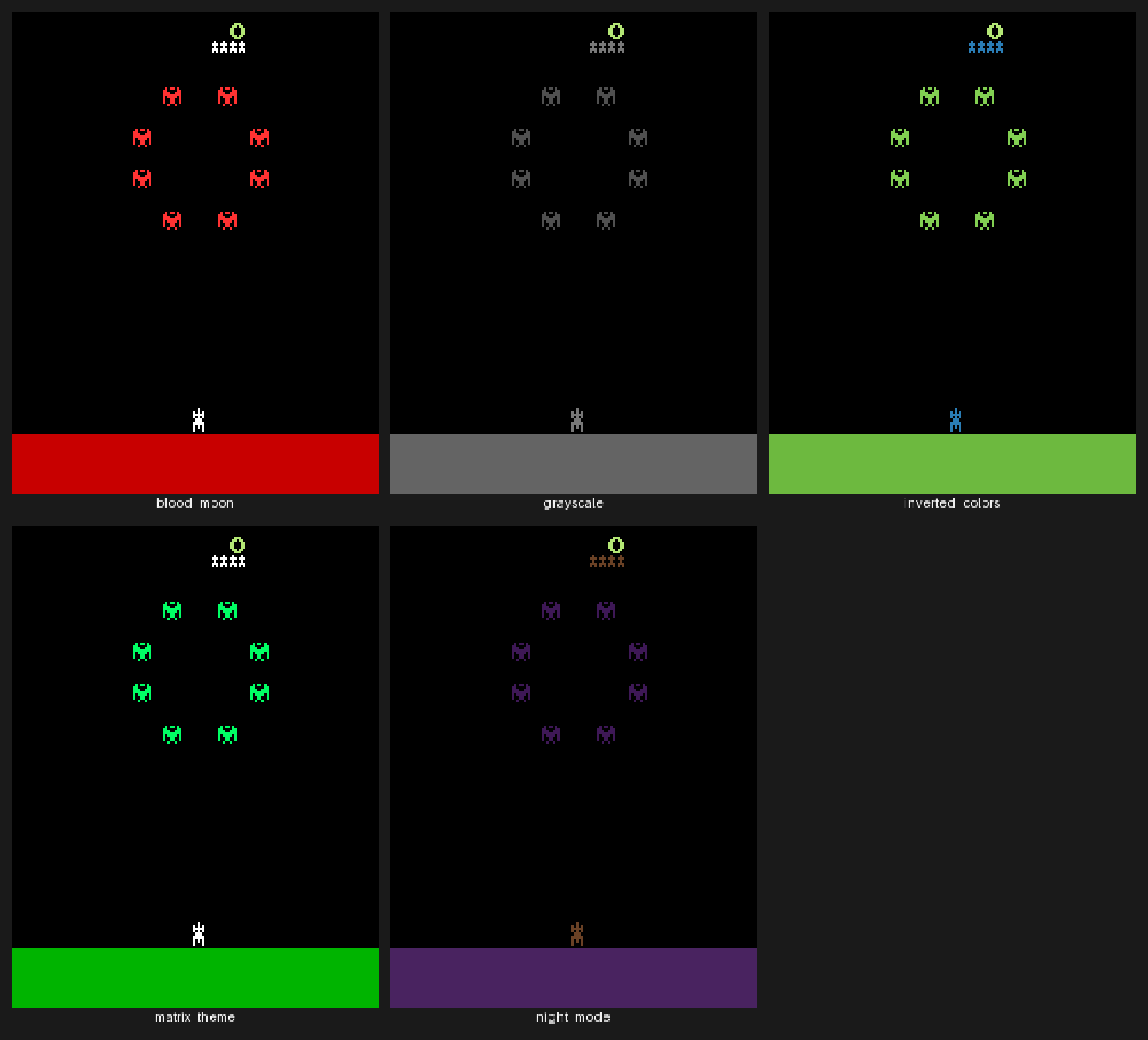}%
  }
  \caption{%
    A selection of visual modifications available for Phoenix,
    with each frame showing a different mod applied to the base environment.%
  }
  \label{fig:mods-Phoenix}
\end{figure}

\newpage

\subsection{Pong}
\subsubsection{Pong \ details}

\vspace{0.5cm}

\noindent
\begin{minipage}[c]{0.52\textwidth}
\raggedright
The player controls the right paddle, competing against a computer-controlled left paddle. Both sides attempt to deflect the ball past the opponent's paddle and into their goal while keeping it out of their own.
\end{minipage}%
\hfill%
\begin{minipage}[c]{0.38\textwidth}
\centering
\setlength{\fboxsep}{0pt}%
\setlength{\fboxrule}{0.4pt}%
\fcolorbox{gray}{white}{%
  \includegraphics[width=\linewidth,height=6.0cm,keepaspectratio]{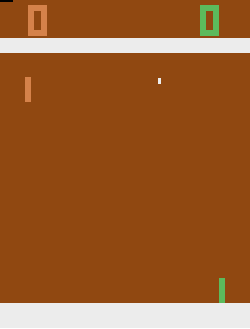}%
}
\end{minipage}

\subsubsection{Available Modifications for Pong}
\begin{table}[h]
    \centering
    \begin{tabular}{p{4cm} | p{9cm}} \toprule
        \textbf{Modification} & \textbf{Effect} \\ \midrule
        change\_background\_color & Changes the background color to navy blue. \\
        change\_player\_color & Changes the player color to red. \\
        lazy\_enemy & Enemy does not move after returning a shot until player hits ball. \\
        linear\_movement & Changes the movement to a linear one (no acceleration). \\
        no\_fire & Disables the FIRE action. \\
        random\_enemy & Randomizes the enemy movement. \\
        reward\_acceleration & Rewards player when ball is accelerated by player (+2). \\
        shift\_enemy & Shifts the enemy on the x-axis. \\
        shift\_player & Shifts the player on the x-axis. \\
        zero\_score & Forces the score to always be zero. \\
        \bottomrule
    \end{tabular}
\end{table}

\subsubsection{Selection of Available Visual Modifications}
\begin{figure}[H]
  \centering
  \setlength{\fboxsep}{0pt}%
  \setlength{\fboxrule}{0.4pt}%
  \fcolorbox{gray}{white}{%
    \includegraphics[height=6cm,keepaspectratio]{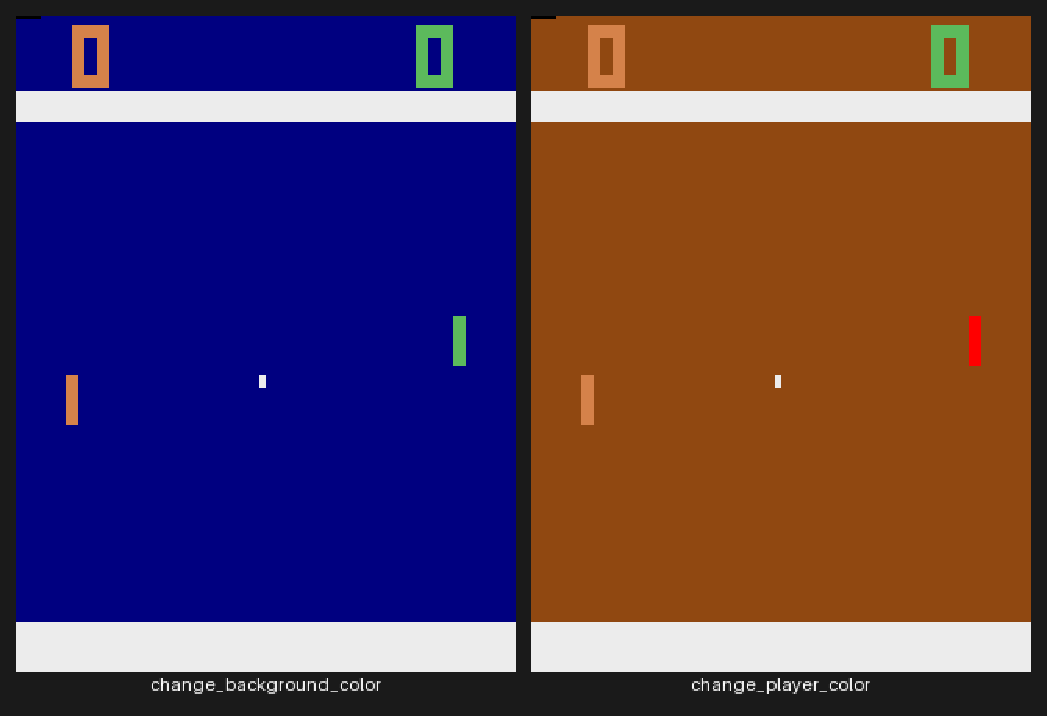}%
  }
  \caption{%
    A selection of visual modifications available for Pong,
    with each frame showing a different mod applied to the base environment.%
  }
  \label{fig:mods-Pong}
\end{figure}

\newpage

\subsection{Qbert}
\subsubsection{Qbert \ details}

\vspace{0.5cm}

\noindent
\begin{minipage}[c]{0.52\textwidth}
\raggedright
The player controls Q*bert, whose goal is to change every cube on a pyramid to a target color by hopping on each one individually. Various hostile creatures roam the pyramid and must be avoided throughout.
\end{minipage}%
\hfill%
\begin{minipage}[c]{0.38\textwidth}
\centering
\setlength{\fboxsep}{0pt}%
\setlength{\fboxrule}{0.4pt}%
\fcolorbox{gray}{white}{%
  \includegraphics[width=\linewidth,height=6.0cm,keepaspectratio]{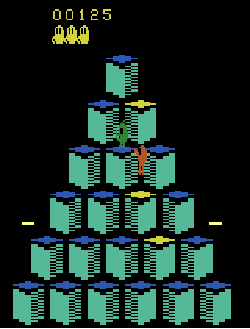}%
}
\end{minipage}

\subsubsection{Available Modifications for Qbert}
\begin{table}[h]
    \centering
    \begin{tabular}{p{4cm} | p{9cm}} \toprule
        \textbf{Modification} & \textbf{Effect} \\ \midrule
        collecting\_bonus\_only & Overrides the rewards to only get reward from collectibles. \\
        dark\_pyramid & Changes the pyramid to a dark aesthetic. \\
        diagonal\_control & Enables diagonal controls. \\
        grayscale & Makes the entire game grayscale. \\
        ice\_pyramid & Changes the pyramid to an icy aesthetic. \\
        inverted\_colors & Swaps Qbert's and Coily's colors. \\
        night\_mode & Dims the background and the characters for a night-time aesthetic. \\
        no\_coily & Removes Coily from the screen. \\
        no\_enemies & Removes all enemies from the screen. \\
        no\_green\_ball & Removes green balls from the screen. \\
        no\_purple\_ball & Removes purple balls from the screen. \\
        no\_red\_balls & Removes red balls from the screen. \\
        no\_sam & Removes Sam from the screen. \\
        penalize\_all\_collectables & Penalizes collecting items. \\
        red\_coily & Changes the color of Coily to red. \\
        swap\_collectibles\_enemies &  Swaps the colors of the collectibles and Enemies. \\
        swap\_colors & Swaps the colors of the tiles. \\
        \bottomrule
    \end{tabular}
\end{table}

\subsubsection{Selection of Available Visual Modifications}
\begin{figure}[H]
  \centering
  \setlength{\fboxsep}{0pt}%
  \setlength{\fboxrule}{0.4pt}%
  \fcolorbox{gray}{white}{%
    \includegraphics[width=\linewidth,keepaspectratio]{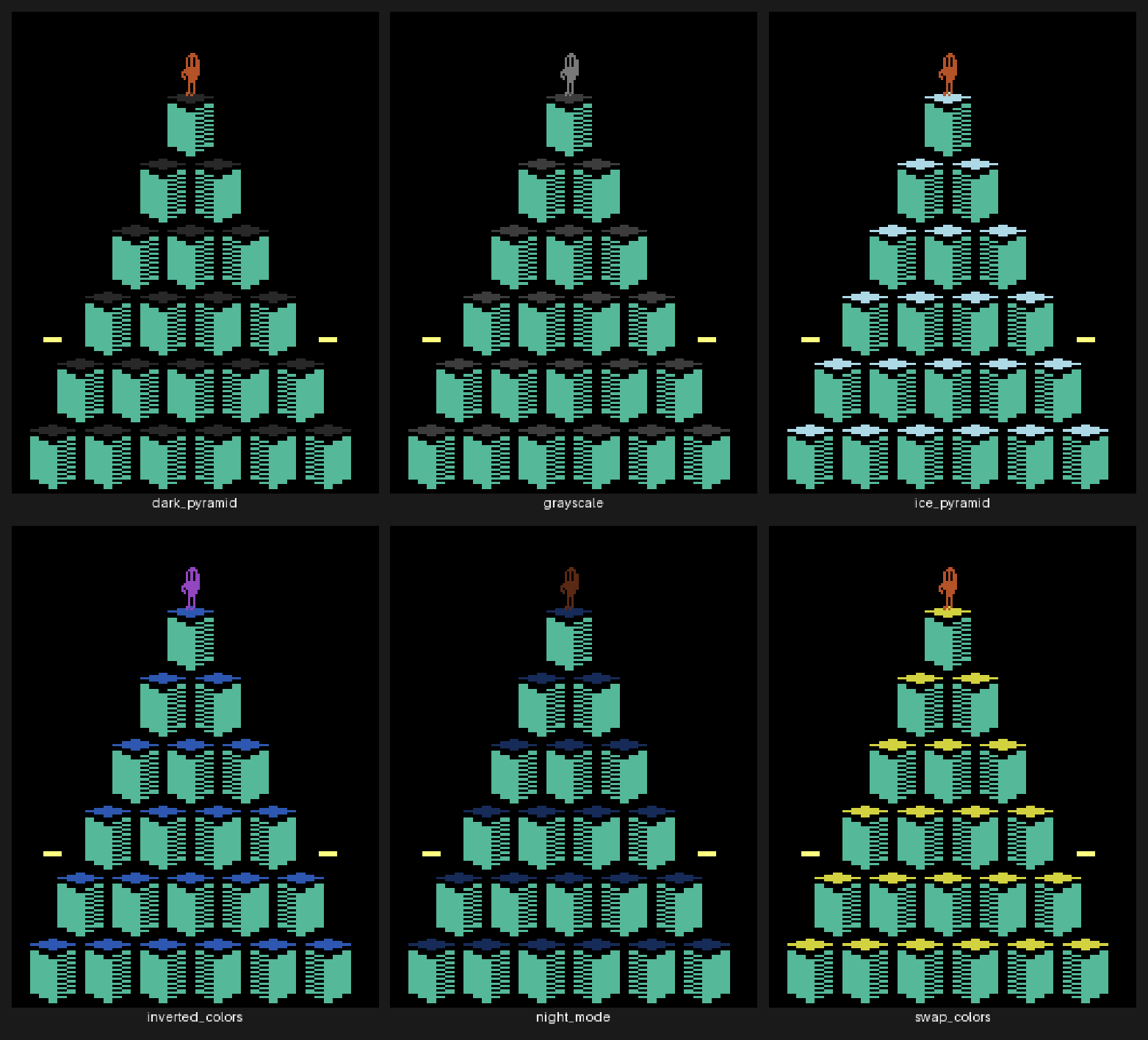}%
  }
  \caption{%
    A selection of visual modifications available for Qbert,
    with each frame showing a different mod applied to the base environment.%
  }
  \label{fig:mods-Qbert}
\end{figure}

\newpage

\subsection{Riverraid}
\subsubsection{Riverraid \ details}

\vspace{0.5cm}

\noindent
\begin{minipage}[c]{0.52\textwidth}
\raggedright
The player pilots a jet flying along a river, steering laterally and firing missiles to destroy enemy targets for points. Fuel depletes continuously and must be replenished by flying over fuel depots before running dry, as an empty tank costs a jet. Collisions with the riverbank or any enemy object, except fuel depots, are equally fatal. The game starts with three jets in reserve, with an additional jet awarded for every 10,000 points scored, up to a maximum of nine.
\end{minipage}%
\hfill%
\begin{minipage}[c]{0.38\textwidth}
\centering
\setlength{\fboxsep}{0pt}%
\setlength{\fboxrule}{0.4pt}%
\fcolorbox{gray}{white}{%
  \includegraphics[width=\linewidth,height=6.0cm,keepaspectratio]{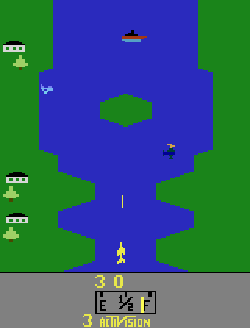}%
}
\end{minipage}

\subsubsection{Available Modifications for Riverraid}
\begin{table}[h]
    \centering
    \begin{tabular}{p{3cm} | p{10cm}} \toprule
        \textbf{Modification} & \textbf{Effect} \\ \midrule
       \multicolumn{2}{l}{\textit{No modifications implemented yet.}} \\
        \bottomrule
    \end{tabular}
\end{table}

\newpage

\subsection{Seaquest}
\subsubsection{Seaquest \ details}

\vspace{0.5cm}

\noindent
\begin{minipage}[c]{0.52\textwidth}
\raggedright
The player commands a submarine capable of moving in all directions and firing torpedoes. The objective is to rescue as many divers as possible while destroying enemy submarines and sharks. The game begins with one active sub and three in reserve, with an additional sub awarded for every 10,000 points scored, up to a maximum of six reserves. The sub is destroyed upon colliding with anything other than friendly divers. Oxygen depletes continuously and must be replenished by surfacing in time. Failing to do so costs a sub. Surfacing with fewer than six divers on board also results in losing one diver.
\end{minipage}%
\hfill%
\begin{minipage}[c]{0.38\textwidth}
\centering
\setlength{\fboxsep}{0pt}%
\setlength{\fboxrule}{0.4pt}%
\fcolorbox{gray}{white}{%
  \includegraphics[width=\linewidth,height=6.0cm,keepaspectratio]{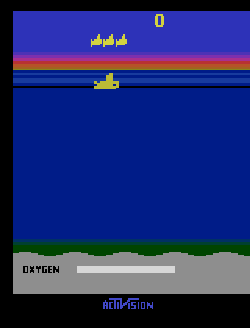}%
}
\end{minipage}

\subsubsection{Available Modifications for Seaquest}
\begin{table}[h]
    \centering
    \begin{tabular}{p{4cm} | p{9cm}} \toprule
        \textbf{Modification} & \textbf{Effect} \\ \midrule
        disable\_enemies & Disable enemies in the environment. \\
        dont\_kill & Internal mod that punishes killing and shooting. \\
        gravity &  Adds gravity to that pulls the player downwards. \\
        mines & Replaces both sharks and enemy submarines with mine sprites. \\
        no\_divers & Internal mod to remove divers from the game. \\
        unlimited\_oxygen & Player always has full oxygen. \\
        \bottomrule
    \end{tabular}
\end{table}

\subsubsection{Selection of Available Visual Modifications}
\begin{figure}[H]
  \centering
  \setlength{\fboxsep}{0pt}%
  \setlength{\fboxrule}{0.4pt}%
  \fcolorbox{gray}{white}{%
    \includegraphics[height=6cm,keepaspectratio]{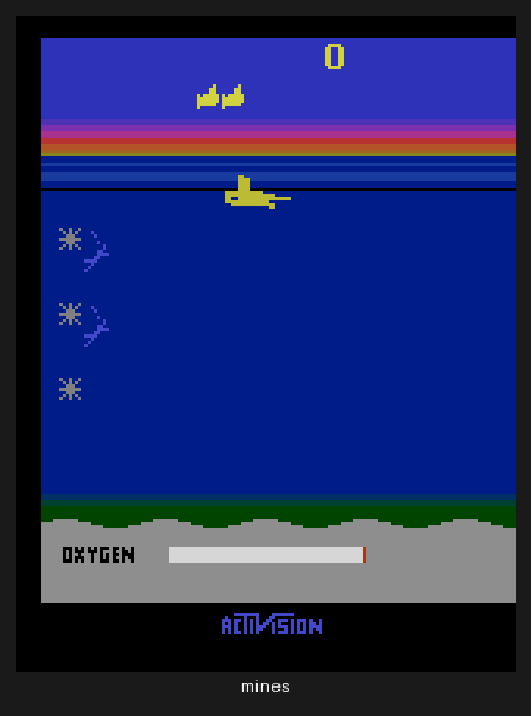}%
  }
  \caption{%
    A selection of visual modifications available for Seaquest,
    with each frame showing a different mod applied to the base environment.%
  }
  \label{fig:mods-Seaquest}
\end{figure}

\newpage

\subsection{SirLancelot}
\subsubsection{SirLancelot \ details}

\vspace{0.5cm}

\noindent
\begin{minipage}[c]{0.52\textwidth}
\raggedright
The player controls Sir Lancelot, mounted on Pegasus, on a mission to rescue a prisoner held within a castle and guarded by a fire-breathing dragon.
\end{minipage}%
\hfill%
\begin{minipage}[c]{0.38\textwidth}
\centering
\setlength{\fboxsep}{0pt}%
\setlength{\fboxrule}{0.4pt}%
\fcolorbox{gray}{white}{%
  \includegraphics[width=\linewidth,height=6.0cm,keepaspectratio]{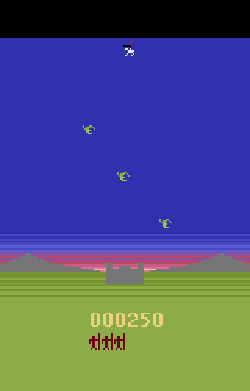}%
}
\end{minipage}

\subsubsection{Available Modifications for SirLancelot}
\begin{table}[h]
    \centering
    \begin{tabular}{p{3cm} | p{10cm}} \toprule
        \textbf{Modification} & \textbf{Effect} \\ \midrule
       \multicolumn{2}{l}{\textit{No modifications implemented yet.}} \\
        \bottomrule
    \end{tabular}
\end{table}

\newpage

\subsection{Skiing}
\subsubsection{Skiing \ details}

\vspace{0.5cm}

\noindent
\begin{minipage}[c]{0.52\textwidth}
\raggedright
The player controls a skier moving laterally down a slope, aiming to pass through all gates, marked by pairs of poles, in the shortest time possible. Missing a gate incurs a five-second penalty. Collisions with gates or trees briefly knock the skier down before the run continues.
\end{minipage}%
\hfill%
\begin{minipage}[c]{0.38\textwidth}
\centering
\setlength{\fboxsep}{0pt}%
\setlength{\fboxrule}{0.4pt}%
\fcolorbox{gray}{white}{%
  \includegraphics[width=\linewidth,height=6.0cm,keepaspectratio]{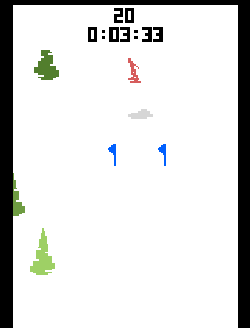}%
}
\end{minipage}

\subsubsection{Available Modifications for Skiing}
\begin{table}[h]
    \centering
    \begin{tabular}{p{3cm} | p{10cm}} \toprule
        \textbf{Modification} & \textbf{Effect} \\ \midrule
        \_dangerous\_moguls & Makes colliding with moguls cause the skier to fall. \\
        \_more\_moguls & Spawns more moguls during the race. \\
        \_more\_trees & Spawns more trees during the race. \\
        \_trees\_everywhere & Allows trees to spawn anywhere across the entire horizontal axis. \\
        blue\_skiier & Replaces the skier sprites with blue versions. \\
        classic\_trees & Replaces the default tree sprites with classic versions. \\
        flag\_flurry & Dramatically increases the number of flags on the mountain. \\
        green\_flags & Recolors all flags to green. \\
        hall\_of\_fame & Places the gates dead center and creates a corridor of trees. \\
        invert\_flag\_colors & Inverts the colors of the slalom flags. \\
        invert\_flags & Flips the orientation or state of the slalom flags on the course. \\
        jump\_to\_break & Allows the skier to jump over moguls using the FIRE action. \\
        moguls\_to\_trees & Transforms all moguls into solid trees. \\
        moving\_flags & Causes the flags to slide horizontally across the screen while you ski. \\
        random\_flags & Randomizes the horizontal placement of every flag. \\
        reward\_at\_gate & Changes the reward function to give +1 reward for passing a gate, \\
        speed\_burst & Allows the skier to accelerate beyond max. speed using DOWN.\\
        thin\_moguls & Replaces the standard moguls with thinner ones. \\
        \bottomrule
    \end{tabular}
\end{table}

\subsubsection{Selection of Available Visual Modifications}
\begin{figure}[H]
  \centering
  \setlength{\fboxsep}{0pt}%
  \setlength{\fboxrule}{0.4pt}%
  \fcolorbox{gray}{white}{%
    \includegraphics[width=\linewidth,keepaspectratio]{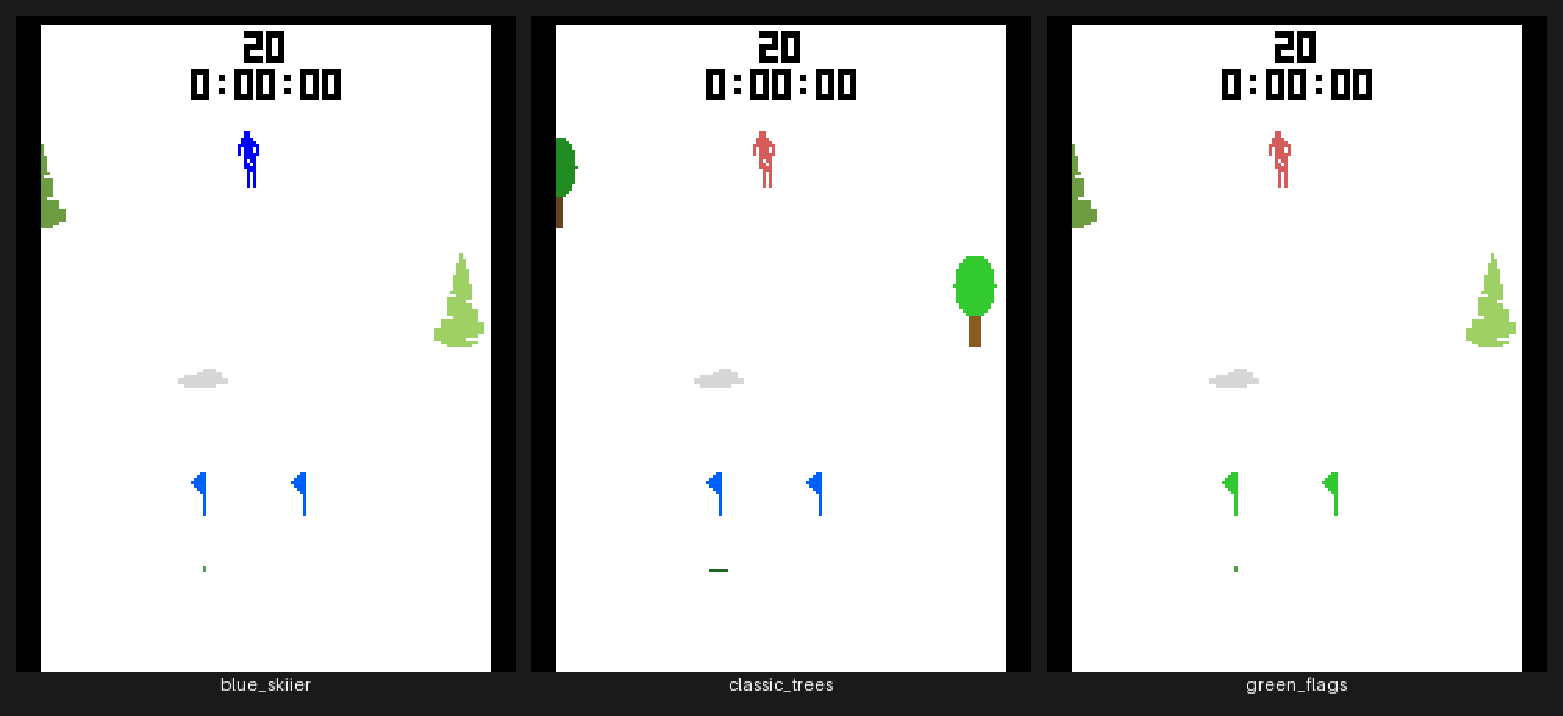}%
  }
  \caption{%
    A selection of visual modifications available for Skiing,
    with each frame showing a different mod applied to the base environment.%
  }
  \label{fig:mods-Skiing}
\end{figure}

\newpage

\subsection{SpaceInvaders}
\subsubsection{SpaceInvaders \ details}

\vspace{0.5cm}

\noindent
\begin{minipage}[c]{0.52\textwidth}
\raggedright
The player operates a laser cannon tasked with shooting down waves of descending space invaders before they reach Earth. The game ends either when all lives are lost to enemy fire or when the invaders successfully reach the ground.
\end{minipage}%
\hfill%
\begin{minipage}[c]{0.38\textwidth}
\centering
\setlength{\fboxsep}{0pt}%
\setlength{\fboxrule}{0.4pt}%
\fcolorbox{gray}{white}{%
  \includegraphics[width=\linewidth,height=6.0cm,keepaspectratio]{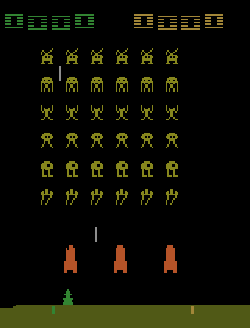}%
}
\end{minipage}

\subsubsection{Available Modifications for SpaceInvaders}
\begin{table}[h]
    \centering
    \begin{tabular}{p{3cm} | p{10cm}} \toprule
        \textbf{Modification} & \textbf{Effect} \\ \midrule
        controllable\_missile & Forces the fired missile's horizontal position to match the player's tank. \\
        disable\_shield\_left & Erases the left bunker from the screen. \\
        disable\_shield\_middle & Erases the middle bunker from the screen. \\
        disable\_shield\_right & Erases the right bunker from the screen. \\
        no\_danger & Removes all player shields and neutralizes incoming enemy projectiles. \\
        shift\_shields & Teleports all bunkers to new horizontal positions. \\
        \bottomrule
    \end{tabular}
\end{table}

\newpage

\subsection{Spacewar}
\subsubsection{Spacewar \ details}

\vspace{0.5cm}

\noindent
\begin{minipage}[c]{0.52\textwidth}
\raggedright
The player pilots a Star Ship through a series of dueling games, aiming to land as many shots on the opponent as possible while avoiding being hit in return.
\end{minipage}%
\hfill%
\begin{minipage}[c]{0.38\textwidth}
\centering
\setlength{\fboxsep}{0pt}%
\setlength{\fboxrule}{0.4pt}%
\fcolorbox{gray}{white}{%
  \includegraphics[width=\linewidth,height=6.0cm,keepaspectratio]{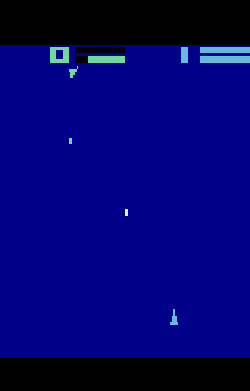}%
}
\end{minipage}

\subsubsection{Available Modifications for Spacewar}
\begin{table}[h]
    \centering
    \begin{tabular}{p{3cm} | p{10cm}} \toprule
        \textbf{Modification} & \textbf{Effect} \\ \midrule
       \multicolumn{2}{l}{\textit{No modifications implemented yet.}} \\
        \bottomrule
    \end{tabular}
\end{table}

\newpage

\subsection{Tennis}
\subsubsection{Tennis \ details}

\vspace{0.5cm}

\noindent
\begin{minipage}[c]{0.52\textwidth}
\raggedright
The player controls the orange player competing against a computer-controlled blue opponent under standard tennis rules. A set is won by reaching at least six games with a margin of two. If the score reaches 6-6, the first player to pull two games ahead takes the match.
\end{minipage}%
\hfill%
\begin{minipage}[c]{0.38\textwidth}
\centering
\setlength{\fboxsep}{0pt}%
\setlength{\fboxrule}{0.4pt}%
\fcolorbox{gray}{white}{%
  \includegraphics[width=\linewidth,height=6.0cm,keepaspectratio]{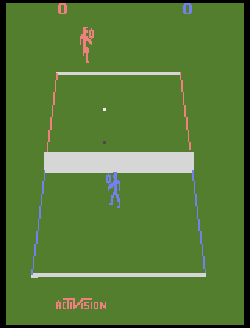}%
}
\end{minipage}

\subsubsection{Available Modifications for Tennis}
\begin{table}[h]
    \centering
    \begin{tabular}{p{3cm} | p{10cm}} \toprule
        \textbf{Modification} & \textbf{Effect} \\ \midrule
        clay\_court & Changes the court colors to a clay court aesthetic. \\
        fast\_enemy & Enemy moves faster. \\
        fast\_player & Increases player walk speed. \\
        grass\_court & Changes the court colors to a grass court aesthetic. \\
        grayscale & Makes the court grayscale. \\
        hard\_court & Changes the court colors to a hard court aesthetic. \\
        high\_bounce & Increases ball bounce velocity. \\
        inverted\_colors & Swaps the player and opponent colors. \\
        lazy\_enemy & Enemy moves slower. \\
        night\_mode & Darkens the court for a night-time aesthetic. \\
        random\_ball\_speed & Ball has random speed after every hit. \\
        random\_walk\_speed & Player has random walk speed everytime they start moving. \\
        super\_gravity & Increases ball gravity. \\
        \bottomrule
    \end{tabular}
\end{table}

\subsubsection{Selection of Available Visual Modifications}
\begin{figure}[H]
  \centering
  \setlength{\fboxsep}{0pt}%
  \setlength{\fboxrule}{0.4pt}%
  \fcolorbox{gray}{white}{%
    \includegraphics[width=\linewidth,keepaspectratio]{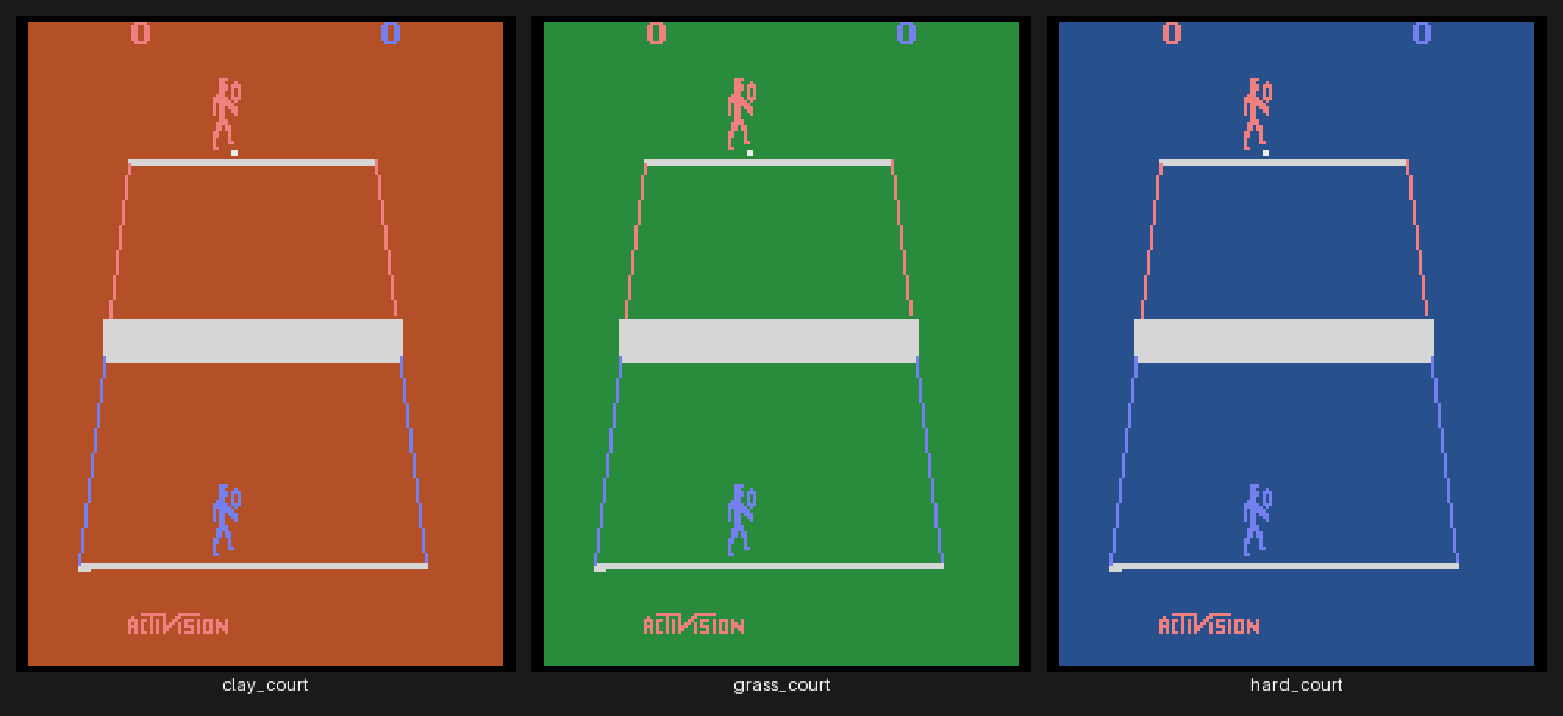}%
  }
  \caption{%
    A selection of visual modifications available for Tennis,
    with each frame showing a different mod applied to the base environment.%
  }
  \label{fig:mods-Tennis}
\end{figure}

\newpage

\subsection{Tetris}
\subsubsection{Tetris \ details}

\vspace{0.5cm}

\noindent
\begin{minipage}[c]{0.52\textwidth}
\raggedright
The player directs a sequence of falling tile-based shapes, maneuvering and rotating them to fill horizontal rows on the board without leaving any gaps.
\end{minipage}%
\hfill%
\begin{minipage}[c]{0.38\textwidth}
\centering
\setlength{\fboxsep}{0pt}%
\setlength{\fboxrule}{0.4pt}%
\fcolorbox{gray}{white}{%
  \includegraphics[width=\linewidth,height=6.0cm,keepaspectratio]{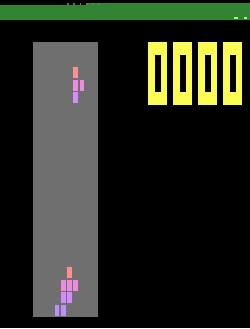}%
}
\end{minipage}

\subsubsection{Available Modifications for Tetris}
\begin{table}[h]
    \centering
    \begin{tabular}{p{3cm} | p{10cm}} \toprule
        \textbf{Modification} & \textbf{Effect} \\ \midrule
       \multicolumn{2}{l}{\textit{No modifications implemented yet.}} \\
        \bottomrule
    \end{tabular}
\end{table}

\newpage

\subsection{Timepilot}
\subsubsection{Timepilot \ details}

\vspace{0.5cm}

\noindent
\begin{minipage}[c]{0.52\textwidth}
\raggedright
The player pilots an aircraft through a series of increasingly challenging encounters, facing enemies whose technology grows progressively more advanced and futuristic as the game progresses.
\end{minipage}%
\hfill%
\begin{minipage}[c]{0.38\textwidth}
\centering
\setlength{\fboxsep}{0pt}%
\setlength{\fboxrule}{0.4pt}%
\fcolorbox{gray}{white}{%
  \includegraphics[width=\linewidth,height=6.0cm,keepaspectratio]{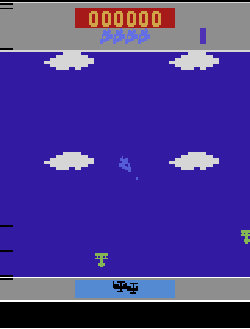}%
}
\end{minipage}

\subsubsection{Available Modifications for Timepilot}
\begin{table}[h]
    \centering
    \begin{tabular}{p{3cm} | p{10cm}} \toprule
        \textbf{Modification} & \textbf{Effect} \\ \midrule
       \multicolumn{2}{l}{\textit{No modifications implemented yet.}} \\
        \bottomrule
    \end{tabular}
\end{table}

\newpage

\subsection{Tron}
\subsubsection{Tron \ details}

\vspace{0.5cm}

\noindent
\begin{minipage}[c]{0.52\textwidth}
\raggedright
The player wields a flying saucer as a weapon, using it to knock out opponents that close in from all directions before they can reach the player.
\end{minipage}%
\hfill%
\begin{minipage}[c]{0.38\textwidth}
\centering
\setlength{\fboxsep}{0pt}%
\setlength{\fboxrule}{0.4pt}%
\fcolorbox{gray}{white}{%
  \includegraphics[width=\linewidth,height=6.0cm,keepaspectratio]{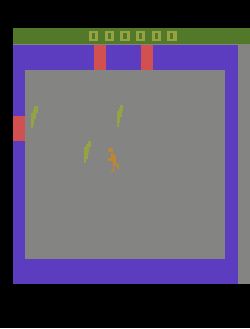}%
}
\end{minipage}

\subsubsection{Available Modifications for Tron}
\begin{table}[h]
    \centering
    \begin{tabular}{p{3cm} | p{10cm}} \toprule
        \textbf{Modification} & \textbf{Effect} \\ \midrule
       \multicolumn{2}{l}{\textit{No modifications implemented yet.}} \\
        \bottomrule
    \end{tabular}
\end{table}

\newpage

\subsection{Turmoil}
\subsubsection{Turmoil \ details}

\vspace{0.5cm}

\noindent
\begin{minipage}[c]{0.52\textwidth}
\raggedright
The player battles waves of descending aliens, shooting them down while weaving through their projectiles and avoiding direct collisions. Enemy formations grow more aggressive and faster as waves are cleared, demanding increasingly precise movement and quick reflexes to survive.
\end{minipage}%
\hfill%
\begin{minipage}[c]{0.38\textwidth}
\centering
\setlength{\fboxsep}{0pt}%
\setlength{\fboxrule}{0.4pt}%
\fcolorbox{gray}{white}{%
  \includegraphics[width=\linewidth,height=6.0cm,keepaspectratio]{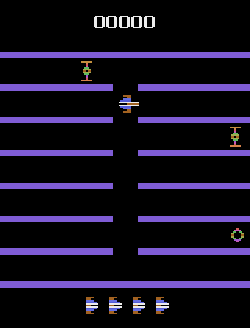}%
}
\end{minipage}

\subsubsection{Available Modifications for Turmoil}
\begin{table}[h]
    \centering
    \begin{tabular}{p{3cm} | p{10cm}} \toprule
        \textbf{Modification} & \textbf{Effect} \\ \midrule
       \multicolumn{2}{l}{\textit{No modifications implemented yet.}} \\
        \bottomrule
    \end{tabular}
\end{table}

\newpage

\subsection{Venture}
\subsubsection{Venture \ details}

\vspace{0.5cm}

\noindent
\begin{minipage}[c]{0.52\textwidth}
\raggedright
The player navigates a dungeon made up of multiple chambers, each containing treasure to collect and monsters to defeat. Armed with a bow, the player must clear each room of its monsters and claim the treasure within, while also avoiding the Hallway Lurkers that roam the corridors between chambers. Lingering too long in a room causes the monsters to become increasingly aggressive.
\end{minipage}%
\hfill%
\begin{minipage}[c]{0.38\textwidth}
\centering
\setlength{\fboxsep}{0pt}%
\setlength{\fboxrule}{0.4pt}%
\fcolorbox{gray}{white}{%
  \includegraphics[width=\linewidth,height=6.0cm,keepaspectratio]{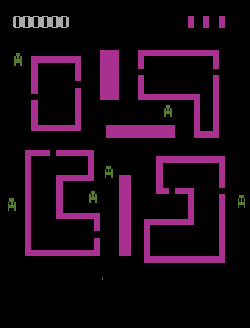}%
}
\end{minipage}

\subsubsection{Available Modifications for Venture}
\begin{table}[h]
    \centering
    \begin{tabular}{p{4cm} | p{9cm}} \toprule
        \textbf{Modification} & \textbf{Effect} \\ \midrule
        blood\_moon & A dark red themed mod. \\
        fast\_enemies & Increases the speed of all enemies. \\
        grayscale & Turns the entire game into grayscale. \\
        inverted\_colors & Inverts all colors in the game. \\
        matrix\_theme & A Matrix-themed mod: black background, green walls/monsters. \\
        night\_mode & Dims the entire screen by 50\% for a night mode experience. \\
        remove\_overworld\_mobs & Removes all monsters from the main map. \\
        reward\_for\_kill & Adds a reward bonus of 100 per kill in the rooms. \\
        slow\_enemies & Reduces the speed of all enemies. \\
        \bottomrule
    \end{tabular}
\end{table}

\subsubsection{Selection of Available Visual Modifications}
\begin{figure}[H]
  \centering
  \setlength{\fboxsep}{0pt}%
  \setlength{\fboxrule}{0.4pt}%
  \fcolorbox{gray}{white}{%
    \includegraphics[width=\linewidth,keepaspectratio]{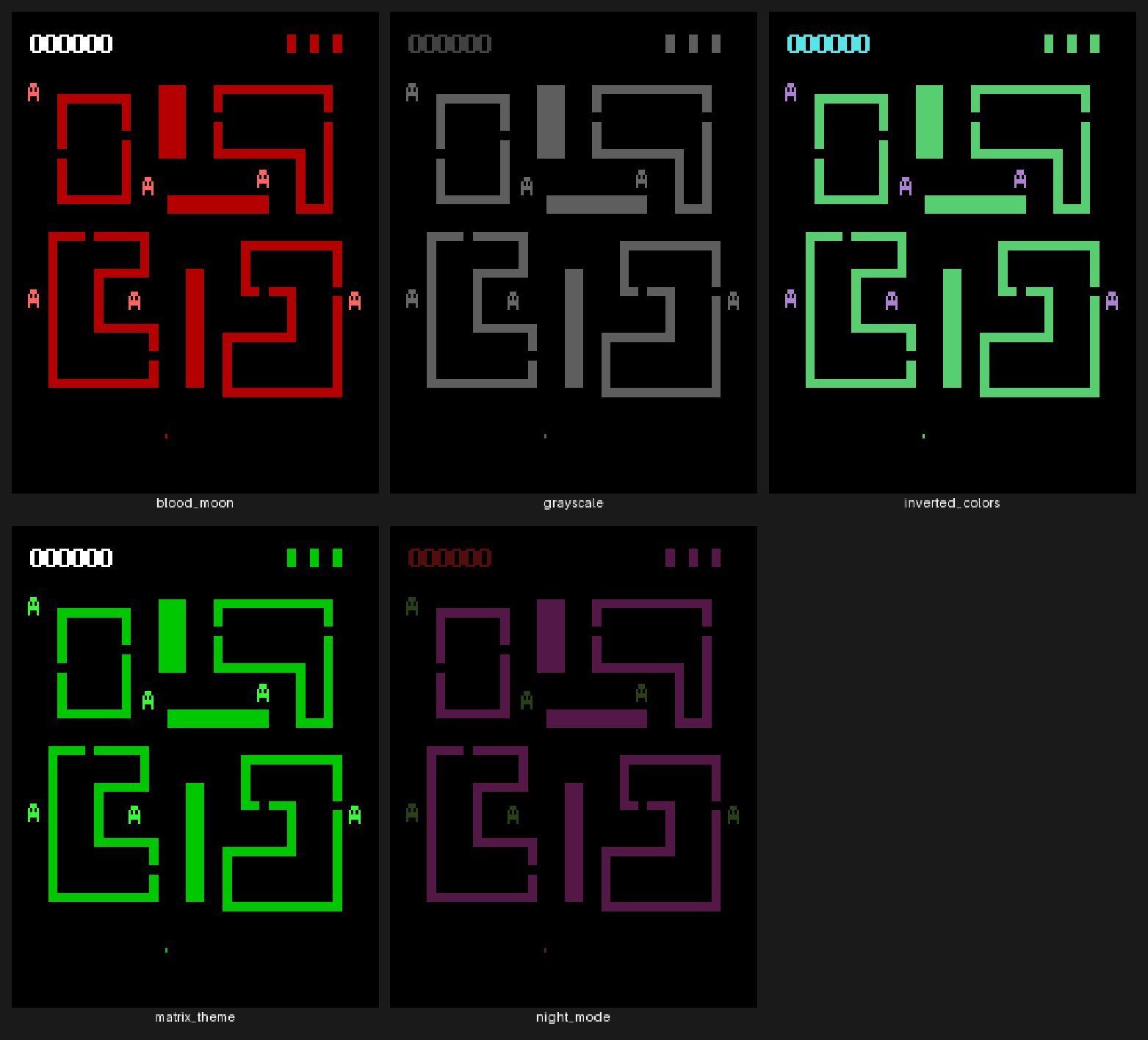}%
  }
  \caption{%
    A selection of visual modifications available for Venture,
    with each frame showing a different mod applied to the base environment.%
  }
  \label{fig:mods-Venture}
\end{figure}

\newpage

\subsection{VideoCube}
\subsubsection{VideoCube \ details}

\vspace{0.5cm}

\noindent
\begin{minipage}[c]{0.52\textwidth}
\raggedright
The player guides Hubie around the faces of a Rubik's cube, swapping adjacent tiles with one another rather than rotating entire rows or columns. The goal is to arrange each face of the cube so that it consists entirely of a single color, offering a distinctly unconventional take on the classic puzzle.
\end{minipage}%
\hfill%
\begin{minipage}[c]{0.38\textwidth}
\centering
\setlength{\fboxsep}{0pt}%
\setlength{\fboxrule}{0.4pt}%
\fcolorbox{gray}{white}{%
  \includegraphics[width=\linewidth,height=6.0cm,keepaspectratio]{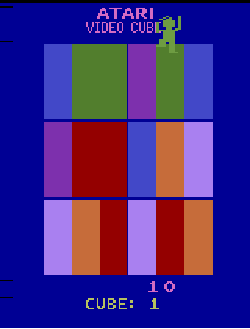}%
}
\end{minipage}

\subsubsection{Available Modifications for VideoCube}
\begin{table}[h]
    \centering
    \begin{tabular}{p{3cm} | p{10cm}} \toprule
        \textbf{Modification} & \textbf{Effect} \\ \midrule
       \multicolumn{2}{l}{\textit{No modifications implemented yet.}} \\
        \bottomrule
    \end{tabular}
\end{table}

\newpage

\subsection{VideoPinball}
\subsubsection{VideoPinball \ details}

\vspace{0.5cm}

\noindent
\begin{minipage}[c]{0.52\textwidth}
\raggedright
The player controls a digital pinball table, using flippers to keep the ball in play as long as possible while racking up points. Targets, bumpers, and other classic pinball elements reward precise flipper control and strategic ball placement.
\end{minipage}%
\hfill%
\begin{minipage}[c]{0.38\textwidth}
\centering
\setlength{\fboxsep}{0pt}%
\setlength{\fboxrule}{0.4pt}%
\fcolorbox{gray}{white}{%
  \includegraphics[width=\linewidth,height=6.0cm,keepaspectratio]{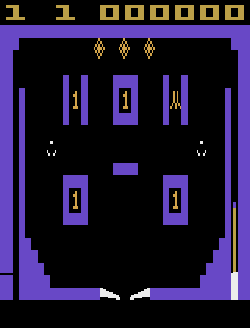}%
}
\end{minipage}

\subsubsection{Available Modifications for VideoPinball}
\begin{table}[h]
    \centering
    \begin{tabular}{p{4cm} | p{9cm}} \toprule
        \textbf{Modification} & \textbf{Effect} \\ \midrule
        constant\_ball\_dynamics & Removes most of the ability to change the trajectory. \\
        low\_tilt\_effect & Decreases the tilt effect when nudging. \\
        never\_activate\_tilt\_mode & Prevent entering tilt mode by resetting the tilt counter. \\
        no\_scoring\_cooldown & Remove the cooldown period after hitting a target. \\
        \bottomrule
    \end{tabular}
\end{table}

\newpage

\subsection{WordZapper}
\subsubsection{WordZapper \ details}

\vspace{0.5cm}

\noindent
\begin{minipage}[c]{0.52\textwidth}
\raggedright
The player shoots letters scrolling across the screen, racing against the clock to hit them in a specific prescribed order. Speed and accuracy are both essential as the pace increases over time.
\end{minipage}%
\hfill%
\begin{minipage}[c]{0.38\textwidth}
\centering
\setlength{\fboxsep}{0pt}%
\setlength{\fboxrule}{0.4pt}%
\fcolorbox{gray}{white}{%
  \includegraphics[width=\linewidth,height=6.0cm,keepaspectratio]{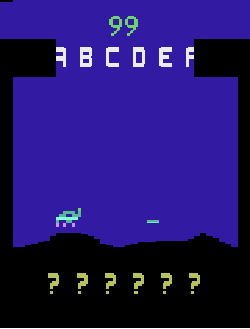}%
}
\end{minipage}

\subsubsection{Available Modifications for WordZapper}
\begin{table}[h]
    \centering
    \begin{tabular}{p{3cm} | p{10cm}} \toprule
        \textbf{Modification} & \textbf{Effect} \\ \midrule
       \multicolumn{2}{l}{\textit{No modifications implemented yet.}} \\
        \bottomrule
    \end{tabular}
\end{table}

\newpage

\newpage


\begin{thebibliography}{43}
\providecommand{\natexlab}[1]{#1}
\providecommand{\url}[1]{\texttt{#1}}
\expandafter\ifx\csname urlstyle\endcsname\relax
  \providecommand{\doi}[1]{doi: #1}\else
  \providecommand{\doi}{doi: \begingroup \urlstyle{rm}\Url}\fi

\bibitem[Agarwal et~al.(2021)Agarwal, Schwarzer, Castro, Courville, and
  Bellemare]{agarwal2021deep}
Agarwal, R., Schwarzer, M., Castro, P.~S., Courville, A.~C., and Bellemare, M.
\newblock Deep reinforcement learning at the edge of the statistical precipice.
\newblock \emph{Advances in Neural Information Processing Systems}, 2021.

\bibitem[Anand et~al.(2019)Anand, Racah, Ozair, Bengio, C{\^o}t{\'e}, and
  Hjelm]{anand2019unsupervised}
Anand, A., Racah, E., Ozair, S., Bengio, Y., C{\^o}t{\'e}, M.-A., and Hjelm,
  R.~D.
\newblock Unsupervised state representation learning in atari.
\newblock In \emph{Advances in Neural Information Processing Systems}, 2019.

\bibitem[Badia et~al.(2020)Badia, Piot, Kapturowski, Sprechmann, Vitvitskyi,
  Guo, and Blundell]{Badia2020agent57}
Badia, A.~P., Piot, B., Kapturowski, S., Sprechmann, P., Vitvitskyi, A., Guo,
  Z.~D., and Blundell, C.
\newblock Agent57: Outperforming the atari human benchmark.
\newblock In \emph{Proceedings of the 37th International Conference on Machine
  Learning}, 2020.

\bibitem[Bellemare et~al.(2013)Bellemare, Naddaf, Veness, and
  Bowling]{bellemare2013arcade}
Bellemare, M.~G., Naddaf, Y., Veness, J., and Bowling, M.
\newblock The arcade learning environment: An evaluation platform for general
  agents.
\newblock \emph{Journal of Artificial Intelligence Research}, 2013.

\bibitem[Beukman et~al.(2026)Beukman, Khetarpal, Zheng, Dabney, Foerster,
  Dennis, and Lyle]{beukman2026preventing}
Beukman, M., Khetarpal, K., Zheng, Z., Dabney, W., Foerster, J., Dennis, M.,
  and Lyle, C.
\newblock Preventing {Learning} {Stagnation} in {PPO} by {Scaling} to 1
  {Million} {Parallel} {Environments}.
\newblock \emph{arXiv}, 2026.

\bibitem[Bl{\"u}ml et~al.(2026)Bl{\"u}ml, Derstroff, Gregori, Dillies,
  Delfosse, and Kersting]{bluml2025deep}
Bl{\"u}ml, J., Derstroff, C., Gregori, B., Dillies, E., Delfosse, Q., and
  Kersting, K.
\newblock Do object channels improve robustness in deep reinforcement
  learning?
\newblock \emph{Trans. Mach. Learn. Res.}, 2026, 2026.
\newblock URL \url{https://openreview.net/forum?id=7BFbso4B3R}.

\bibitem[Bonnet et~al.(2024)Bonnet, Luo, Byrne, Surana, Abramowitz, Duckworth,
  Coyette, Midgley, Tegegn, Kalloniatis, Mahjoub, Macfarlane, Smit, Grinsztajn,
  Boige, Waters, Mimouni, Sob, de~Kock, Singh, Furelos-Blanco, Le, Pretorius,
  and Laterre]{bonnet2024jumanji}
Bonnet, C., Luo, D., Byrne, D., Surana, S., Abramowitz, S., Duckworth, P.,
  Coyette, V., Midgley, L.~I., Tegegn, E., Kalloniatis, T., Mahjoub, O.,
  Macfarlane, M., Smit, A.~P., Grinsztajn, N., Boige, R., Waters, C.~N.,
  Mimouni, M.~A., Sob, U. A.~M., de~Kock, R., Singh, S., Furelos-Blanco, D.,
  Le, V., Pretorius, A., and Laterre, A.
\newblock Jumanji: a diverse suite of scalable reinforcement learning
  environments in jax, 2024.

\bibitem[Cao et~al.(2026)Cao, Deng, Shindo, Delfosse, Wen, Wang, Bl{\"u}ml,
  Tauchmann, and Kersting]{cao2026kintsugi}
Cao, T., Deng, Y., Shindo, H., Delfosse, Q., Wen, L., Wang, S., Bl{\"u}ml, J.,
  Tauchmann, C., and Kersting, K.
\newblock Kintsugi: Learning policies by repairing executable knowledge bases.
\newblock \emph{arXiv preprint arXiv:2605.09487}, 2026.

\bibitem[Cobbe et~al.(2019)Cobbe, Klimov, Hesse, Kim, and
  Schulman]{cobbe2019quantifying}
Cobbe, K., Klimov, O., Hesse, C., Kim, T., and Schulman, J.
\newblock Quantifying generalization in reinforcement learning.
\newblock In \emph{Proceedings of the 36th International Conference on Machine
  Learning}, 2019.

\bibitem[Cobbe et~al.(2020)Cobbe, Hesse, Hilton, and
  Schulman]{cobbe2020procgen}
Cobbe, K., Hesse, C., Hilton, J., and Schulman, J.
\newblock Leveraging procedural generation to benchmark reinforcement learning.
\newblock In \emph{International conference on machine learning}, 2020.

\bibitem[Dalton \& Frosio(2020)Dalton and Frosio]{dalton2020accelerating}
Dalton, S. and Frosio, I.
\newblock Accelerating reinforcement learning through {GPU} atari emulation.
\newblock In Larochelle, H., Ranzato, M., Hadsell, R., Balcan, M., and Lin, H.
  (eds.), \emph{Advances in Neural Information Processing Systems 33: Annual
  Conference on Neural Information Processing Systems 2020, NeurIPS 2020,
  December 6-12, 2020, virtual}, 2020.

\bibitem[Delfosse et~al.(2023)Delfosse, Shindo, Dhami, and
  Kersting]{Delfosse2023InterpretableAE}
Delfosse, Q., Shindo, H., Dhami, D.~S., and Kersting, K.
\newblock Interpretable and explainable logical policies via neurally guided
  symbolic abstraction.
\newblock \emph{Advances in Neural Information Processing Systems (NeurIPS)}, 2023.

\bibitem[Delfosse et~al.(2024{\natexlab{a}})Delfosse, Bl{\"u}ml, Gregori, and
  Kersting]{delfosse2024hackatari}
Delfosse, Q., Bl{\"u}ml, J., Gregori, B., and Kersting, K.
\newblock Hackatari: Atari learning environments for robust and continual
  reinforcement learning.
\newblock \emph{arXiv preprint}, 2024{\natexlab{a}}.

\bibitem[Delfosse et~al.(2024{\natexlab{b}})Delfosse, Bl{\"{u}}ml, Gregori,
  Sztwiertnia, and Kersting]{delfosse2024ocatari}
Delfosse, Q., Bl{\"{u}}ml, J., Gregori, B., Sztwiertnia, S., and Kersting, K.
\newblock {OCAtari}: {O}bject-centric {Atari} 2600 reinforcement learning
  environments.
\newblock \emph{Reinforcement Learning Journal}, 2024{\natexlab{b}}.

\bibitem[Delfosse et~al.(2025)Delfosse, Bl{\"u}ml, Tatai, Vincent, Gregori,
  Dillies, Peters, Rothkopf, and Kersting]{delfosse2025deep}
Delfosse, Q., Bl{\"u}ml, J., Tatai, F., Vincent, T., Gregori, B., Dillies, E.,
  Peters, J., Rothkopf, C., and Kersting, K.
\newblock Deep reinforcement learning agents are not even close to human
  intelligence.
\newblock \emph{arXiv}, 2025.

\bibitem[Diuk et~al.(2008)Diuk, Cohen, and Littman]{diuk2008object}
Diuk, C., Cohen, A., and Littman, M.~L.
\newblock An object-oriented representation for efficient reinforcement
  learning.
\newblock In \emph{Proceedings of the 25th international conference on Machine
  learning}, 2008.

\bibitem[Farebrother et~al.(2018)Farebrother, Machado, and
  Bowling]{farebrother2018generalization}
Farebrother, J., Machado, M.~C., and Bowling, M.
\newblock Generalization and regularization in dqn.
\newblock In \emph{NeurIPS Deep Reinforcement Learning Workshop}, 2018.

\bibitem[Freeman et~al.(2021)Freeman, Frey, Raichuk, Girgin, Mordatch, and
  Bachem]{freeman2021brax}
Freeman, C.~D., Frey, E., Raichuk, A., Girgin, S., Mordatch, I., and Bachem, O.
\newblock Brax--a differentiable physics engine for large scale rigid body
  simulation.
\newblock In \emph{Thirty-fifth Conference on Neural Information Processing
  Systems Datasets and Benchmarks Track}, 2021.

\bibitem[Fu et~al.(2020)Fu, Kumar, Nachum, Tucker, and Levine]{fu2020d4rl}
Fu, J., Kumar, A., Nachum, O., Tucker, G., and Levine, S.
\newblock D4rl: Datasets for deep data-driven reinforcement learning.
\newblock \emph{arXiv preprint}, 2020.

\bibitem[Garnelo et~al.(2016)Garnelo, Arulkumaran, and
  Shanahan]{garnelo2016towards}
Garnelo, M., Arulkumaran, K., and Shanahan, M.
\newblock Towards deep symbolic reinforcement learning.
\newblock \emph{arXiv preprint}, 2016.

\bibitem[Gu et~al.(2025)Gu, Shi, Wen, Jin, Mazumdar, Chi, Wierman, and
  Spanos]{robustrl2024}
Gu, S., Shi, L., Wen, M., Jin, M., Mazumdar, E., Chi, Y., Wierman, A., and
  Spanos, C.
\newblock Robust gymnasium: A unified modular benchmark for robust
  reinforcement learning.
\newblock \emph{ICLR}, 2025.

\bibitem[Huang et~al.(2022)Huang, Dossa, Ye, Braga, Chakraborty, Mehta, and
  Araújo]{huang2022cleanrl}
Huang, S., Dossa, R. F.~J., Ye, C., Braga, J., Chakraborty, D., Mehta, K., and
  Araújo, J.~G.
\newblock Cleanrl: High-quality single-file implementations of deep
  reinforcement learning algorithms.
\newblock \emph{Journal of Machine Learning Research}, 2022.

\bibitem[Koch et~al.(2021)Koch, Langosco, Pfau, Le, and
  Sharkey]{koch2021objective}
Koch, J., Langosco, L., Pfau, J., Le, J., and Sharkey, L.
\newblock Objective robustness in deep reinforcement learning.
\newblock \emph{arXiv}, 2021.

\bibitem[Koyamada et~al.(2023)Koyamada, Okano, Nishimori, Murata, Habara, Kita,
  and Ishii]{koyamada2023pgx}
Koyamada, S., Okano, S., Nishimori, S., Murata, Y., Habara, K., Kita, H., and
  Ishii, S.
\newblock Pgx: Hardware-accelerated parallel game simulators for reinforcement
  learning.
\newblock In \emph{Advances in Neural Information Processing Systems},
  volume~36, pp.\  45716--45743, 2023.

\bibitem[Lange(2026)]{gymnax2022github}
Lange, R.~T.
\newblock {gymnax}: A {JAX}-based reinforcement learning environment library,
  2026.
\newblock URL \url{http://github.com/RobertTLange/gymnax}.

\bibitem[Laskin et~al.(2020)Laskin, Lee, Stooke, Pinto, Abbeel, and
  Srinivas]{laskin2020rad}
Laskin, M., Lee, K., Stooke, A., Pinto, L., Abbeel, P., and Srinivas, A.
\newblock Reinforcement learning with augmented data.
\newblock In \emph{Advances in Neural Information Processing Systems}, pp.\
  19884--19895, 2020.

\bibitem[Liu et~al.(2024)Liu, Yuan, Hu, Li, Chen, Liu, Lu, and
  Jia]{liu2024rlgpt}
Liu, S., Yuan, H., Hu, M., Li, Y., Chen, Y., Liu, S., Lu, Z., and Jia, J.
\newblock {RL-GPT:} integrating reinforcement learning and code-as-policy.
\newblock In Globersons, A., Mackey, L., Belgrave, D., Fan, A., Paquet, U.,
  Tomczak, J.~M., and Zhang, C. (eds.), \emph{Advances in Neural Information
  Processing Systems 37: Annual Conference on Neural Information Processing
  Systems 2024, NeurIPS 2024, Vancouver, BC, Canada, December 10 - 15, 2024},
  2024.
\newblock URL
  \url{http://papers.nips.cc/paper_files/paper/2024/hash/31f119089f702e48ecfd138c1bc82c4a-Abstract-Conference.html}.

\bibitem[Machado et~al.(2018)Machado, Bellemare, Talvitie, Veness, Hausknecht,
  and Bowling]{machado2018revisiting}
Machado, M.~C., Bellemare, M.~G., Talvitie, E., Veness, J., Hausknecht, M., and
  Bowling, M.
\newblock Revisiting the arcade learning environment: Evaluation protocols and
  open problems for general agents.
\newblock \emph{Journal of Artificial Intelligence Research}, 2018.

\bibitem[Matthews et~al.(2024)Matthews, Beukman, Ellis, Samvelyan, Jackson,
  Coward, and Foerster]{matthews2024craftax}
Matthews, M., Beukman, M., Ellis, B., Samvelyan, M., Jackson, M.~T., Coward,
  S., and Foerster, J.~N.
\newblock Craftax: A lightning-fast benchmark for open-ended reinforcement
  learning.
\newblock In \emph{Forty-first International Conference on Machine Learning},
  2024.

\bibitem[Mnih et~al.(2015)Mnih, Kavukcuoglu, Silver, Rusu, Veness, Bellemare,
  Graves, Riedmiller, Fidjeland, Ostrovski, Petersen, Beattie, Sadik,
  Antonoglou, King, Kumaran, Wierstra, Legg, and Hassabis]{mnih2015human}
Mnih, V., Kavukcuoglu, K., Silver, D., Rusu, A.~A., Veness, J., Bellemare,
  M.~G., Graves, A., Riedmiller, M., Fidjeland, A.~K., Ostrovski, G., Petersen,
  S., Beattie, C., Sadik, A., Antonoglou, I., King, H., Kumaran, D., Wierstra,
  D., Legg, S., and Hassabis, D.
\newblock Human-level control through deep reinforcement learning.
\newblock \emph{Nature}, 518\penalty0 (7540):\penalty0 529--533, 2015.

\bibitem[Moos et~al.(2022)Moos, Hansel, Abdulsamad, Stark, Clever, and
  Peters]{moos2022robust}
Moos, J., Hansel, K., Abdulsamad, H., Stark, S., Clever, D., and Peters, J.
\newblock Robust reinforcement learning: A review of foundations and recent
  advances.
\newblock \emph{Machine Learning and Knowledge Extraction}, 2022.

\bibitem[Nikulin et~al.(2023)Nikulin, Kurenkov, Zisman, Sinii, Agarkov, and
  Kolesnikov]{nikulin2023xlandminigrid}
Nikulin, A., Kurenkov, V., Zisman, I., Sinii, V., Agarkov, A., and Kolesnikov,
  S.
\newblock {XL}and-minigrid: Scalable meta-reinforcement learning environments
  in {JAX}.
\newblock In \emph{Intrinsically-Motivated and Open-Ended Learning Workshop,
  NeurIPS2023}, 2023.

\bibitem[Radji et~al.(2026)Radji, Michel, and Piteau]{radji2026octax}
Radji, W., Michel, T., and Piteau, H.
\newblock Octax: Accelerated {CHIP}-8 arcade environments for reinforcement
  learning in {JAX}.
\newblock In \emph{The Fourteenth International Conference on Learning
  Representations}, 2026.

\bibitem[Rutherford et~al.(2024)Rutherford, Ellis, Gallici, Cook, Lupu,
  Ingvarsson, Willi, Hammond, Khan, de~Witt, Souly, Bandyopadhyay, Samvelyan,
  Jiang, Lange, Whiteson, Lacerda, Hawes, Rockt{\"a}schel, Lu, and
  Foerster]{flair2024jaxmarl}
Rutherford, A., Ellis, B., Gallici, M., Cook, J., Lupu, A., Ingvarsson, G.,
  Willi, T., Hammond, R., Khan, A., de~Witt, C.~S., Souly, A., Bandyopadhyay,
  S., Samvelyan, M., Jiang, M., Lange, R.~T., Whiteson, S., Lacerda, B., Hawes,
  N., Rockt{\"a}schel, T., Lu, C., and Foerster, J.~N.
\newblock Jaxmarl: Multi-agent rl environments and algorithms in jax.
\newblock In \emph{The Thirty-eighth Conference on Neural Information Processing
  Systems Datasets and Benchmarks Track}, 2024.

\bibitem[Schulman et~al.(2017)Schulman, Wolski, Dhariwal, Radford, and
  Klimov]{Schulman2017ProximalPO}
Schulman, J., Wolski, F., Dhariwal, P., Radford, A., and Klimov, O.
\newblock Proximal policy optimization algorithms.
\newblock \emph{arXiv preprint}, 2017.

\bibitem[Shindo et~al.(2025)Shindo, Delfosse, Dhami, and
  Kersting]{shindo2024blendrl}
Shindo, H., Delfosse, Q., Dhami, D.~S., and Kersting, K.
\newblock Blendrl: A framework for merging symbolic and neural policy learning.
\newblock In \emph{International Conference on Learning Representations}, 2025.

\bibitem[Taiga et~al.(2023)Taiga, Agarwal, Farebrother, Courville, and
  Bellemare]{taiga2023investigating}
Taiga, A.~A., Agarwal, R., Farebrother, J., Courville, A., and Bellemare, M.~G.
\newblock Investigating multi-task pretraining and generalization in
  reinforcement learning.
\newblock In \emph{The eleventh international conference on learning
  representations}, 2023.

\bibitem[Todorov et~al.(2012)Todorov, Erez, and Tassa]{todorov2012mujoco}
Todorov, E., Erez, T., and Tassa, Y.
\newblock Mujoco: A physics engine for model-based control.
\newblock In \emph{International conference on intelligent robots and systems},
  2012.

\bibitem[Tunyasuvunakool et~al.(2020)Tunyasuvunakool, Muldal, Doron, Liu,
  Bohez, Merel, Erez, Lillicrap, Heess, and
  Tassa]{tunyasuvunakool2020dm_control}
Tunyasuvunakool, S., Muldal, A., Doron, Y., Liu, S., Bohez, S., Merel, J.,
  Erez, T., Lillicrap, T., Heess, N., and Tassa, Y.
\newblock dm\_control: Software and tasks for continuous control.
\newblock \emph{Software Impacts}, 2020.

\bibitem[Waytowich et~al.(2024)Waytowich, White, Sunbeam, and
  Goecks]{waytowich2024atarigpt}
Waytowich, N.~R., White, D., Sunbeam, M.~D., and Goecks, V.~G.
\newblock Atari-{GPT}: Benchmarking multimodal large language models as
  low-level policies in {Atari} games.
\newblock \emph{arXiv preprint arXiv:2408.15950}, 2024.

\bibitem[Weng et~al.(2022)Weng, Lin, Huang, Liu, Makoviichuk, Makoviychuk, Liu,
  Song, Luo, Jiang, Xu, and YAN]{weng2022envpool}
Weng, J., Lin, M., Huang, S., Liu, B., Makoviichuk, D., Makoviychuk, V., Liu,
  Z., Song, Y., Luo, T., Jiang, Y., Xu, Z., and YAN, S.
\newblock Envpool: A highly parallel reinforcement learning environment
  execution engine.
\newblock In \emph{Thirty-sixth Conference on Neural Information Processing
  Systems Datasets and Benchmarks Track}, 2022.

\bibitem[Yarats et~al.(2021)Yarats, Kostrikov, and Fergus]{yarats2021image}
Yarats, D., Kostrikov, I., and Fergus, R.
\newblock Image augmentation is all you need: Regularizing deep reinforcement
  learning from pixels.
\newblock In \emph{International Conference on Learning Representations}, 2021.

\bibitem[Young \& Tian(2019)Young and Tian]{young2019minatar}
Young, K. and Tian, T.
\newblock Minatar: An atari-inspired testbed for thorough and reproducible
  reinforcement learning experiments.
\newblock \emph{arXiv}, 2019.

\end{thebibliography}
\end{document}